\documentclass[11pt]{article}

\usepackage[final]{acl}

\usepackage{times}
\usepackage{latexsym}

\usepackage[T1]{fontenc}

\usepackage[utf8]{inputenc}

\usepackage{microtype}

\usepackage{inconsolata}

\usepackage{graphicx}

\usepackage{hyperref}
\usepackage{url}
\usepackage{graphicx}
\usepackage{float}
\usepackage{enumitem}
\usepackage{xspace}
\usepackage{amsmath}
\usepackage{amssymb}
\usepackage[dvipsnames]{xcolor}
\usepackage{cleveref}
\usepackage{listings}
\usepackage{multirow}
\usepackage{rotating}
\usepackage{subcaption}
\usepackage{tabularx}
\usepackage{booktabs}
\usepackage{enumitem}
\usepackage{fontawesome5}

\newcommand\blfootnote[1]{%
  \begingroup
  \renewcommand\thefootnote{}\footnote{#1}%
  \addtocounter{footnote}{-1}%
  \endgroup
}

\definecolor{codekw}{HTML}{2563EB}    
\definecolor{codestr}{HTML}{16A34A}   
\definecolor{codecom}{HTML}{8A94A6}   
\definecolor{codeself}{HTML}{7C3AED}  
\definecolor{codenum}{HTML}{B45309}   
\definecolor{codebg}{HTML}{F8FAFC}    
\definecolor{codeframe}{HTML}{E2E8F0} 

\lstdefinestyle{torch}{
    language=Python,
    basicstyle=\ttfamily\footnotesize,
    keywordstyle=\color{codekw}\bfseries,
    stringstyle=\color{codestr},
    commentstyle=\color{codecom}\itshape,
    emph={self},
    emphstyle=\color{codeself}\itshape,
    showstringspaces=false,
    keepspaces=true,
    columns=fullflexible,
    breaklines=true,
    breakatwhitespace=true,
    tabsize=4,
}

\usepackage[most]{tcolorbox}
\usepackage{xcolor}

\definecolor{rqblue}{HTML}{EAF2FF}
\definecolor{rqblueframe}{HTML}{2F6FDB}

\usepackage[most]{tcolorbox}
\usepackage{xcolor}

\usepackage[most]{tcolorbox}
\usepackage{xcolor}

\definecolor{rqbluebg}{HTML}{EAF2FF}
\definecolor{rq1color}{HTML}{f2cc8f}
\definecolor{rqgreenbg}{HTML}{EAF7EF}
\definecolor{rqpurplebg}{HTML}{F3EAFE}

\definecolor{rqblue}{HTML}{2563EB}
\definecolor{rqgreen}{HTML}{16A34A}
\definecolor{rqpurple}{HTML}{7C3AED}

\newtcolorbox{researchquestion}[3]{
    enhanced,
    breakable,
    colback=#2,
    colframe=#3,
    colbacktitle=#3,
    coltitle=white,
    fonttitle=\bfseries,
    title=#1,
    boxrule=0.8pt,
    arc=2mm,
    left=2mm,
    right=2mm,
    top=1mm,
    bottom=1mm,
    boxed title style={
        arc=2mm,
        boxrule=0pt,
    }
}

\usepackage{amssymb} 

\newtcolorbox{takeawaybox}[3]{
    enhanced,
    breakable,
    colback=#2,
    colframe=#3,
    coltitle=#3,
    fonttitle=\bfseries,
    title={\checkmark\ #1},
    boxrule=0.9pt,
    arc=2mm,
    left=2mm,
    right=2mm,
    top=2mm,
    bottom=2mm,
    attach boxed title to top left={xshift=6mm, yshift=-\tcboxedtitleheight/2},
    boxed title style={
        boxrule=0pt,
        frame hidden,
        interior code={
            \path[fill=white]
                (interior.north west) rectangle (interior.east);
            \path[fill=#2]
                (interior.west) rectangle (interior.south east);
        },
    },
    before skip=15pt,
    after skip=15pt,
}

\definecolor{rqbluebg}{HTML}{EAF2FF}
\definecolor{rqblue}{HTML}{2563EB}

 \newtcolorbox{emptyroundbox}[3]{
      enhanced,
      breakable,
      colback=#2,
      colframe=#3,
      coltitle=#3,
      fonttitle=\bfseries,
      title=#1,
      boxrule=0.8pt,
      arc=2mm,
      left=2mm,
      right=2mm,
      top=2mm,
      bottom=2mm,
      attach boxed title to top left={xshift=6mm, yshift=-\tcboxedtitleheight/2},
      boxed title style={
          boxrule=0pt,
          frame hidden,
          interior code={
              \path[fill=white]
                  (interior.north west) rectangle (interior.east);
              \path[fill=#2]
                  (interior.west) rectangle (interior.south east);
          },      
      },  
      before skip=15pt,
      after skip=15pt,
  }

\definecolor{takeawaybluebg}{HTML}{EFF6FF}
\definecolor{takeawaygreenbg}{HTML}{ECFDF5}
\definecolor{takeawaypurplebg}{HTML}{F5F3FF}

\definecolor{takeawayblue}{HTML}{1D4ED8}
\definecolor{takeawaygreen}{HTML}{059669}
\definecolor{takeawaypurple}{HTML}{7C3AED}

\newcommand{\rsq}{$\rm{R}^2$\xspace}
\newcommand{\llada}{LLaDA\xspace}
\newcommand{\dream}{Dream\xspace}
\newcommand{\mask}{[\textsc{mask}]\xspace}

\title{
\vspace{0.0em}
\fontsize{19}{23}\selectfont
\textbf{Subliminal Clocks: \\
Latent Time Modelling in Diffusion Language Models}}
\author{
\normalsize\mdseries
Maximo Rulli\textsuperscript{1} \quad
Thomas Fontanari\textsuperscript{1,*} \quad
Simone Petruzzi\textsuperscript{1,*} \quad
Federico Alvetreti\textsuperscript{1}
\\[0.35em]
\normalsize
Giorgio Strano\textsuperscript{1} \quad
Donato Crisostomi\textsuperscript{1} \quad
Giorgos Nikolaou\textsuperscript{2} \quad
Tommaso Mencattini\textsuperscript{2}
\\[0.35em]
\normalsize
Andrea Santilli\textsuperscript{3,$\dagger$} \quad
Emanuele Rodol\`a\textsuperscript{1} \quad
Simone Scardapane\textsuperscript{1} \quad
Alessio Devoto\textsuperscript{3,$\dagger$}
}

\begin{document}
\maketitle
\blfootnote{%
  \textsuperscript{*}Equal contribution.\\
  \textsuperscript{1}Sapienza University of Rome. \quad
  \textsuperscript{2}EPFL. \quad
  \textsuperscript{3}NVIDIA.\\
  \textsuperscript{$\dagger$}Work performed while at Sapienza University of Rome.\\
  Correspondence: \texttt{rulli.2154435@studenti.uniroma1.it}.%
}
\begin{abstract}
Diffusion Language Models (DLMs) have recently emerged as a promising alternative to autoregressive models. Unlike standard diffusion-based approaches, DLMs are not explicitly conditioned on a timestep, raising a natural question: do these models internally represent denoising progress, and how is such information used downstream?
In this work, we show that DLMs do in fact encode a latent representation related to the diffusion timestep within their residual streams. 
We find that this signal can be reliably extracted using probes across layers, indicating that denoising progress is decodable from internal activations. 
We further demonstrate that steering the model along a low-dimensional subspace associated with the inferred timestep allows us to systematically modulate its notion of denoising progress, leading to predictable changes in model confidence and entropy. 
Finally, we analyse the geometry of the identified representation, showing that it exhibits structured and interpretable properties in activation space, and shedding light on how such a signal is processed by these models.
\end{abstract}

\section{Introduction}
Large Language Models (LLMs) \citep{claude, llama4, gpt4, deepseekv3, gemma_2025} are driving a paradigm shift across various scientific and social domains. Understanding the capabilities of these models remains an active area of research. Many works aim to characterise the manifolds and subspaces spanned by the features of these models' activations \citep{saglam2026largelanguagemodelsencode, tiblias2026hypothesisdrivenfeaturemanifoldanalysis, joshi2025geometrydecisionmakinglanguage, shai2025transformersrepresentbeliefstate, bhalla2026sparseautoencoderscaptureconcept}, helping explain how they represent and transform concepts across their computations.
\begin{figure}
     \begin{subfigure}{0.45\textwidth}
        \includegraphics[width=\columnwidth]{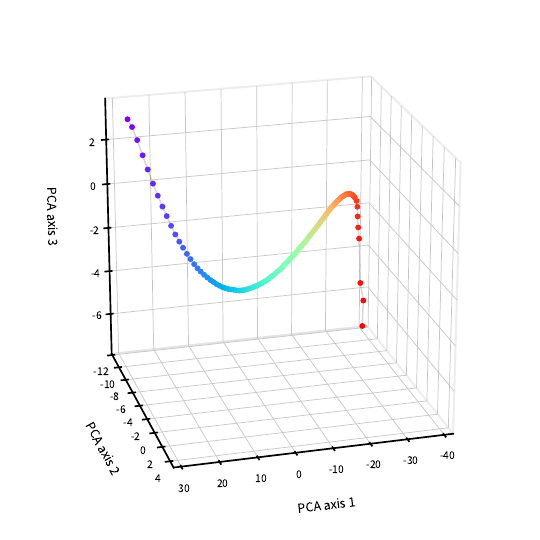}
    \end{subfigure}
     \caption{\textbf{3D projection of latent denoising step modelling for \llada}. \llada represents its $\tau$ subspace as a low-dimensional manifold-like curve, progressively modelling the denoising progress from all \mask (red) to no remaining \mask (purple).}
     \label{fig:example_3d_llada}
\end{figure}
Recently, large-scale Masked Diffusion Language Models \citep{llada, llada-1.5, dream} have emerged as an alternative paradigm for text generation. 
Rather than generating strictly left-to-right, they progressively denoise \mask tokens in a BERT-style manner \citep{bert}; throughout this work, we call this masked-token family DLMs.
Most existing research has primarily focused on improving their efficiency and generation quality~\citep{fastdlm2, arriola2025block, du2026r, zekri2026generalized,yu2026introspective}. In contrast, their internal mechanisms and representations remain largely unexplored. 
More recently, a small but growing body of work has begun investigating the interpretability of these models. 
For instance, \citet{saellada} train Sparse Autoencoders~\citep{sae} on \llada and \dream to enable concept steering at inference time, \citet{rulli2025attentionsinks} analyse how attention sinks move across denoising steps and how robust DLMs are to pruning them, \citet{maskdevil} demonstrate how \mask tokens can be manipulated to induce harmful or deceptive generations, and \citet{maskdistract} show that DLM performance is strongly influenced by the number of \mask tokens present at each denoising step.
While these works provide valuable insights into the behaviour of DLMs, the internal dynamics underlying the denoising process itself remain poorly understood. 
In particular, it remains unclear whether DLMs internally represent denoising progress during generation. 
Previous theoretical work suggests that the strong performance of absorbing-state DLMs may be largely attributed to the special role of the \mask token: the reverse process only needs to denoise masked positions, and the corresponding score can be expressed as a clean-data conditional distribution up to an analytic time-dependent factor \citep{ou2024your}. 
Complementarily, masked diffusion objectives can be reformulated in terms of the number of masked tokens, making continuous denoising time closely related to a relaxation of the mask ratio and connecting these models to time-agnostic masked models \citep{zheng2025maskeddiffusionmodelssecretly}. 
This makes the fraction of masked tokens a natural observable proxy for denoising progress, but leaves open whether and how such a signal is internally represented by the model.
%

In this work, we investigate this through the following three research questions:
\begin{emptyroundbox}{RQ1}{rqbluebg}{rqblue} \label{rq:1}
Do Diffusion Language Models internally represent a signal related to the denoising step?
\end{emptyroundbox}
\begin{emptyroundbox}{RQ2}{rqgreenbg}{rqgreen} \label{rq:2}
Is the identified signal important to the generation process? 
How does modifying it affect the models’ downstream computations?
\end{emptyroundbox}
\begin{emptyroundbox}{RQ3}{rqpurplebg}{rqpurple} \label{rq:3}
What characteristics does the signal exhibit? 
Is there a unified model-level representation?
\end{emptyroundbox}
We address these questions in three stages: first by probing for denoising-time information, then by causally steering the recovered signal, and finally by characterising its geometry across layers. 
We conduct our analysis on LLaDA-1.5 \citep{llada-1.5} (which we refer to as \llada) and \dream \citep{dream}, two representative large-scale masked diffusion language models.
\section{Background}
%



DLMs are trained to recover a clean sequence $x_0$ from a corrupted version of it~\citep{li2022diffusion, sahoo, ou2024your, shi2025simplifiedgeneralizedmaskeddiffusion}. In absorbing-state models, the corrupted sequence $x_s$ is obtained from $x_0$ by replacing a subset of its positions with a special $\mask$ token. We define: 
\[ \mathcal{M}(x_s) = \{j: x_s^j = \mask \,\} \]
to be the set of masked positions. 
The scalar $s \in [0, 1]$ controls the corruption level: $s=0$ recovers the clean sequence, while $s = 1$ masks every position. 
During training, a noise level $s \sim \mathcal{U}(0, 1]$ is sampled and each position is corrupted independently according to the forward process
\[ x_s^j \sim \mathrm{Cat}\big(\cdot; p=(1 - s)\, x_0^j + s\, \mask\big) \]
so that position $j$ keeps its original token with probability $1 - s$ and is replaced by $\mask$ with probability $s$. 
The DLM $p_\theta$ takes the corrupted sequence $x_s$ as input and predicts, at every position $j$, a distribution $p_\theta(x_0^j \mid x_s)$ over the original vocabulary $\mathcal{V}$. 
The training minimises a cross-entropy loss applied only at the masked positions, reweighted by $1/s$:
\begin{equation} \label{eq:loss}
    -\mathbb{E}_{x_0,\,s,\,x_s}\left[\frac{1}{s} \sum_{j \in \mathcal{M}(x_s)} \log p_\theta(x^j_0 \mid x_s) \right].
\end{equation}
Inference proceeds through a sequence of $T$ denoising steps. Generation is initialised at step $t = 0$ with $L$ \mask tokens appended to the prompt, and ends at $t=T$ with all tokens unmasked. At each step $t$, the model operates on the current partially-masked sequence $x_t$ to obtain $p_\theta(\cdot \mid x_t)$ at every masked position, and a subset of $\mathcal{M}(x_t)$ is selected to be unmasked by sampling according to an unmasking policy. This policy determines how many tokens are revealed per step.
At inference time we measure the fraction of unmasked tokens as 
\begin{equation}
\label{eq:mask_ratio}
\tau_t
:=
1 - \frac{|\mathcal{M}(x_{t})|}{L}.
\end{equation}
In \Cref{sec:timetauAppendix}, we provide more details of how $\tau_t$ is equivalent, in expectation, to the complement of the continuous diffusion-time variable $s$.

\begin{figure*}[t] 
    \centering
    \hfill 
    \includegraphics[width=1\textwidth]{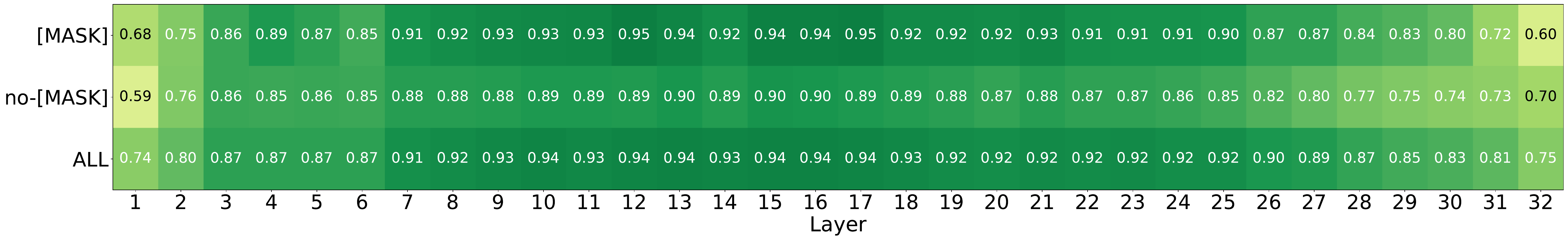}
    \caption{\textbf{MLP probes recover the $\tau$ signal}. The \rsq coefficient (higher is better) degrades as we probe deeper into the model, to the point where the MSE almost reduces to half the variance of $\tau$. Moreover, both \mask and non-\mask tokens seem to carry this information, attaining similarly high coefficients.}
    \label{fig:llada_mlp_sweep}
\end{figure*}

We therefore interpret $\tau_t$ as an empirical measure of denoising progress and use it as a proxy for the notion of denoising time in masked diffusion language models.
\section{Recovering \texorpdfstring{$\tau$}{tau}} 
\label{sec:probe_training}
In this section we address RQ\ref{rq:1} by attempting to recover the $\tau$ signal from the model's hidden states.
To test whether $\tau$ can be recovered from the residual stream of the model, we train MLP probes to predict the current $\tau_t$ of a given sequence by using a residual-stream hidden state as input.
Specifically, we fit an MLP ($\phi_l(\cdot) : \mathbb{R}^d \rightarrow (0, 1)$) for each layer $l$ in the model, where $d$ is the dimensionality of the hidden state (see \Cref{sec:probetrainAppendix} for architectural and training details). 
Each MLP is trained to minimise the regression loss
\[
\mathcal{L}_\text{MLP} = \mathbb{E}_{t,n,j}\left[\left(\tau_t - \phi_l(h^j_{t, l, n}) \right)^2\right]
\]
where $n$ is the example index, $j$ is the token index in the example sequence, and $h^j_{t,l,n}$ denotes the hidden state. Throughout this work, we use $\mathbb{E}_{r}$ to denote the empirical expectation over the corresponding quantity $r$.
We fit separate probes for every layer in both \llada and \dream. 
Additionally, we vary the subset of tokens used for training, considering only \mask tokens, non-\mask tokens, or all tokens.
This allows us to investigate whether information related to $\tau$ is preferentially encoded in specific types of tokens.
\Cref{fig:llada_mlp_sweep} shows the \rsq{} obtained for each layer of \llada.
The probes maintain an $\mathrm{R}^2 > 0.5$ across all layers of the model, indicating that information about $\tau$ is consistently represented throughout the network depth. 
Performance slightly degrades in the earliest and latest layers, suggesting that the representation becomes less accessible at the boundaries of the computation. 
We observe similar trends for \dream\ (see \Cref{sec:probeSteeringAppendix}).

We also find that activations corresponding to \mask\ tokens yield marginally better predictions than non-\mask\ tokens, though both token types support accurate recovery.
This suggests that $\tau$ information is not exclusively localised to masked positions, but is instead distributed across the residual stream.

Importantly, each probe receives a single contextualised hidden state, without direct access to the full sequence or neighbouring activations; hinting that: individual contextualised token representations themselves encode substantial information about $\tau$.
Thus, the high performance of the probes indicates that individual token representations \textbf{carry a latent representation} of the sequence-level statistic $\tau$.

%
As a result, we answer affirmatively to RQ\ref{rq:1} and conclude that DLMs do internally model a signal related to the current denoising step.
%
%
In the next section, we show a natural construction of a signal related to $\tau$ and use it to steer the model at inference time.
\begin{takeawaybox}{Takeaway for RQ1}{takeawaybluebg}{takeawayblue}
\emph{Diffusion Language Models do internally encode a denoising-step-related signal, and we can accurately recover it.}
\end{takeawaybox}
\section{Assessing the Signal's Importance}
\label{sec:signalimportance}
We now investigate whether the found signal can be explicitly extracted and used to steer the model.
In particular, we aim to identify directions in activation space associated with the denoising-time statistic $\tau$, and study whether steering along these directions changes the model's behaviour during inference.
%
\subsection{Approximating \texorpdfstring{$\tau$}{tau}}
To characterise and approximate the internal representation of $\tau$, we turn to mean activation vectors~\citep{gurnee2025counting, wang2025semmeansteering, panickssery2024meansteering}. 
The key intuition behind this approach is that averaging activations over many samples suppresses instance-specific information while preserving systematic signals shared across examples, such as the one associated with $\tau$.

Following \citet{wang2025semmeansteering}, we compute mean activation vectors over all hidden states in the response window, grouping them according to their denoising-progress bin. 
More formally, for each denoising-progress bin \(t\) and layer $l$, we define:
\begin{equation}
\label{mean_activation}
    \mu_{t,l}
    :=
    \mathbb{E}_{n,j}
    \left[
        h^j_{t,l,n}
    \right],
\end{equation}

We discretise the denoising process into $100$ bins, yielding $100$ mean vectors per layer. These bins discretize normalised denoising progress and do not imply that generation uses 100 inference steps; each inference state is assigned to the bin closest to its corresponding \(\tau_t\) value. 
This results in $3200$ mean vectors for \llada and $2800$ for \dream.

\begin{figure}[t]
    \centering
    \begin{subfigure}{0.23\textwidth}
        \includegraphics[width=\textwidth]{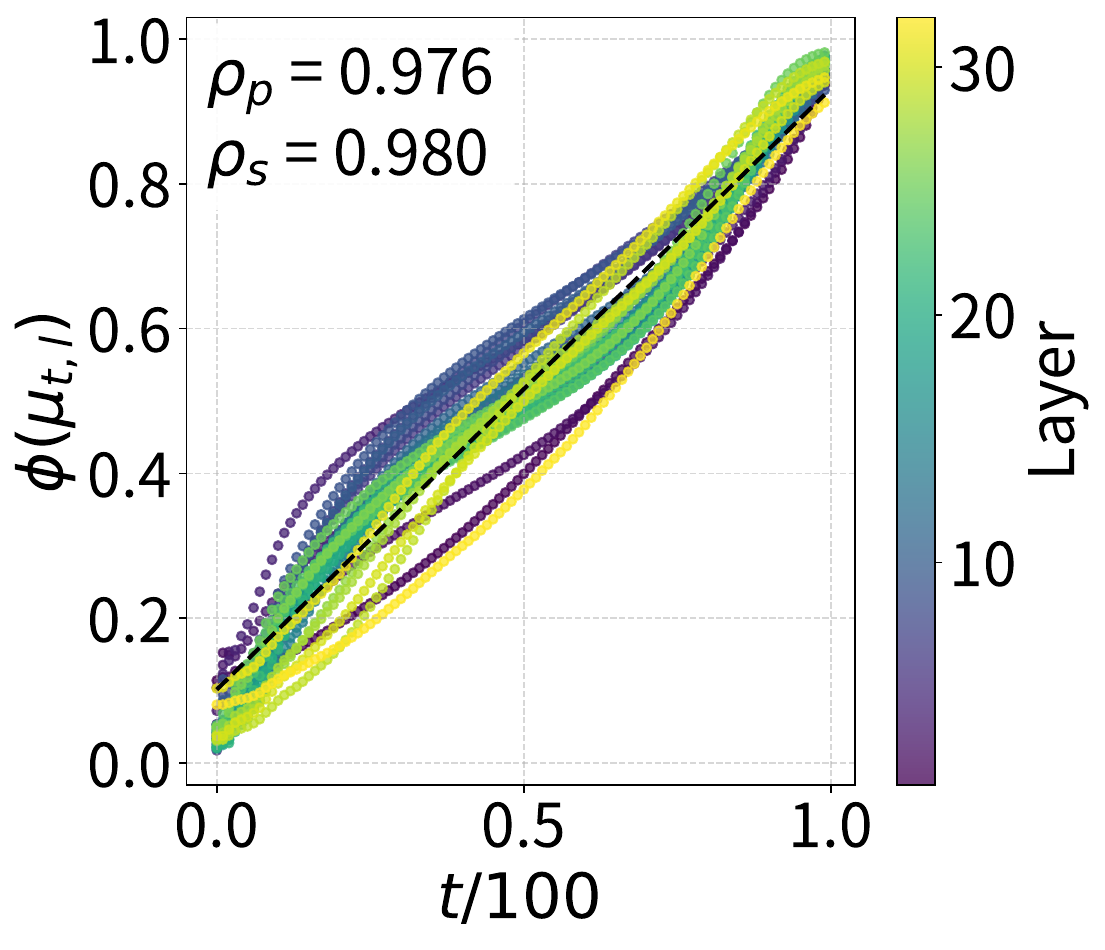}
    \end{subfigure}
    \medskip
    \begin{subfigure}{0.23\textwidth}
        \includegraphics[width=\textwidth]{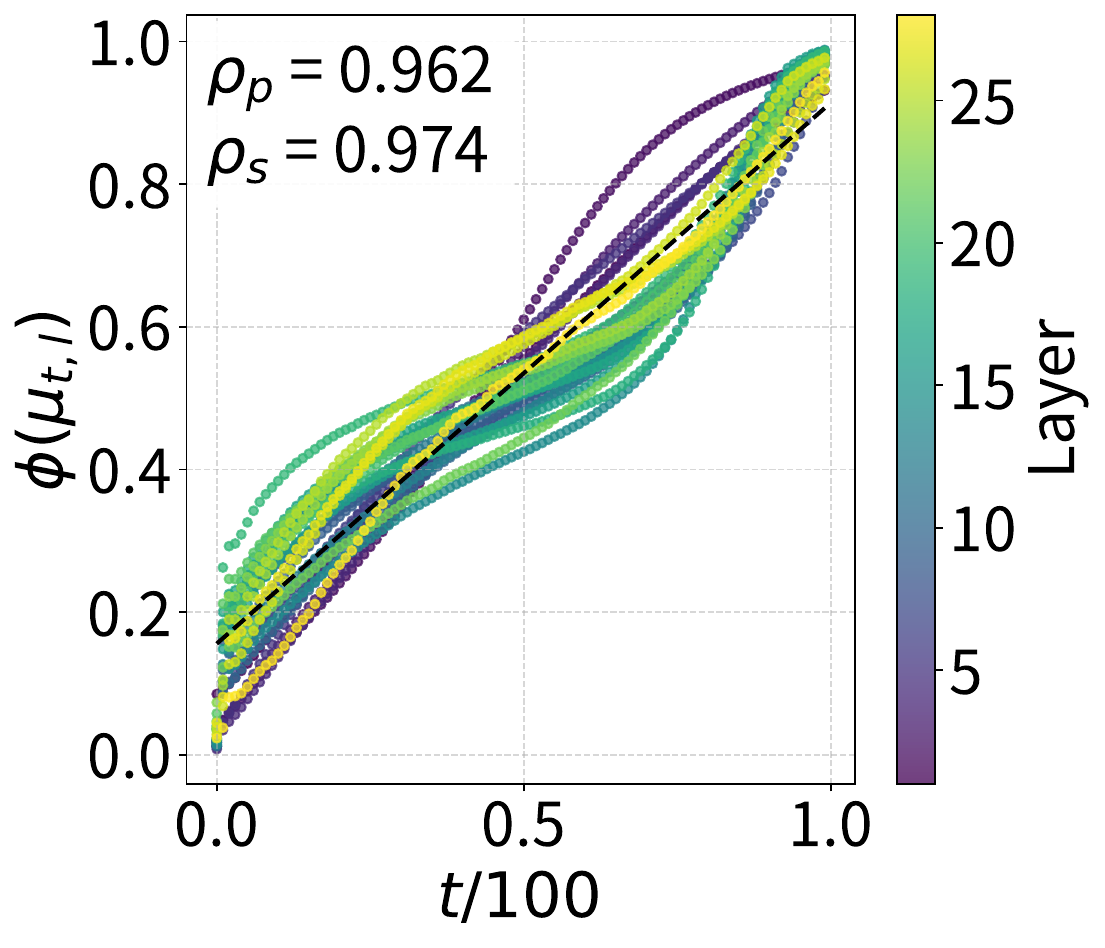}
    \end{subfigure}
    \caption{\textbf{$\phi_l(\mu_{t,l})$ and $\mu_{t,l}$ are highly correlated}. For each of the found $\mu_{t,l}$ we compute $\phi_l(\mu_{t,l})$ and plot it against the corresponding $\tau=t/100$. We observe high correlations for both \llada (left) and \dream (right).}
    \label{fig:mean_mlps_correlation}
\end{figure}

To verify whether the computed mean vectors capture the same signal identified in \Cref{sec:probe_training}, we apply the same trained probes, without any additional tuning, to \(\mu_{t,l}\) and compute the correlation between \(\phi_l(\mu_{t,l})\) and the normalised bin index \(t/100\).
If both methods capture a similar latent direction, then the probe should behave monotonically as the mean vector's index increases.
For \llada, the correlations reach $0.976$ Pearson and $0.980$ Spearman, while for \dream\ they reach $0.962$ and $0.974$, respectively. 
Moreover, as shown in \Cref{fig:mean_mlps_correlation}, the probe predictions align almost perfectly with the denoising bins encoded by each mean vector.
These results suggest that the $\tau$ information identified by the probes is not merely recoverable at the token level, but also emerges as a robust global structure in the residual stream after averaging across examples. 
In other words, the $\tau$ signal appears to correspond to a coherent latent direction that is consistently represented across tokens and sequences, making it a natural target for the steering interventions we introduce next.
Having verified that the mean vectors track the same signal recovered by the probes, we next assess its causal relevance by intervening on it during inference.
%

\subsection{Steering}
\label{sec:ratio_perturbation}
We use the computed activations to steer the model by swapping the mean vectors for the current step with those of the target $\hat{t}$ (where $\hat{t}$ is the discretised target bin whose denoising progress we steer the model towards):
\begin{equation} \label{eq:steer_mean_switch}
    \tilde{h}^j_{t,l} := h^j_{t,l} - \mu_{t,l} + \mu_{\hat{t},l} .
\end{equation}
We note that, since the perturbation is defined as a difference $\Delta_{l}^{t \to \hat{t}} := \mu_{\hat{t},l}-\mu_{t,l}$, the shared component $\bar{\mu}_l := \mathbb{E}_t[\mu_{t,l}]$ cancels out. 
This removes the part of the representation that is independent of denoising progress and intrinsic to the hidden states of layer $l$. 
Based on the results shown in \Cref{fig:llada_mlp_sweep}, we apply steering across all tokens to ensure that no $\tau$-related information is inadvertently preserved when perturbing the hidden states.
As a control condition, we construct the empirical covariance matrix $C_{t,l}$ (formed from the activations at step $t$ and layer $l$), and we sample a random vector that is applied across all hidden states at that step:
\begin{equation} \label{eq:random_steer}
    \tilde{h}^j_{t,l} := h^j_{t,l} + a^{t \to \hat{t}}, \quad a \sim \mathcal{N}(\mathbf{0},C_{t,l})
\end{equation}
We denote by $a^{t \to \hat{t}}$ the normalised version of the sampled vector $a$, rescaled such that $\left\|a^{t \to \hat{t}}\right\| = \left\| \mu_{\hat{t},l} - \mu_{t,l} \right\|$.
%
%
%

%
%
%
%
\subsection{Experimental results}
We now investigate whether steering the internal representation of $\tau$ produces consistent downstream changes in the model predictions. 
Intuitively, if the characterised signal truly relates to a model’s mechanism for representing $\tau$, then shifting this representation towards larger or smaller $\tau$ values should alter the model’s confidence, entropy, and token distributions accordingly.
To quantify these effects, we compare the clean token distributions
\[
    p^j_{t,n} := p_\theta(x^j_{0,n} \mid x_{t,n})
\]
against the steered ones
\[
    \tilde p^j_{t,n} := p_\theta(x^j_{0,n} \mid \tilde x_{t,n}),
\]
where $n \in [N]$ denotes the sequence index.

We evaluate three complementary quantities: the variation in entropy, the variation in confidence, and the KL-divergence between clean and steered distributions:
\begin{equation}
\begin{aligned}
\Delta \bar{S}_{t}
&:= \mathbb{E}_{n,j}
\left[
\mathrm{H}(\tilde p^j_{t,n})
-
\mathrm{H}(p^j_{t,n})
\right], \\
\Delta \bar{c}_{t}
&:= \mathbb{E}_{n,j}
\left[
\max(\tilde p^j_{t,n})
-
\max(p^j_{t,n})
\right], \\
\overline{\mathrm{KL}}_{t}
&:= \mathbb{E}_{n,j}
\left[
D_{\mathrm{KL}}
\left(
p^j_{t,n}
\,\|\, 
\tilde p^j_{t,n}
\right)
\right].
\end{aligned}
\end{equation}

\begin{figure}[t]
    \begin{subfigure}{\columnwidth}
        \includegraphics[width=\textwidth]{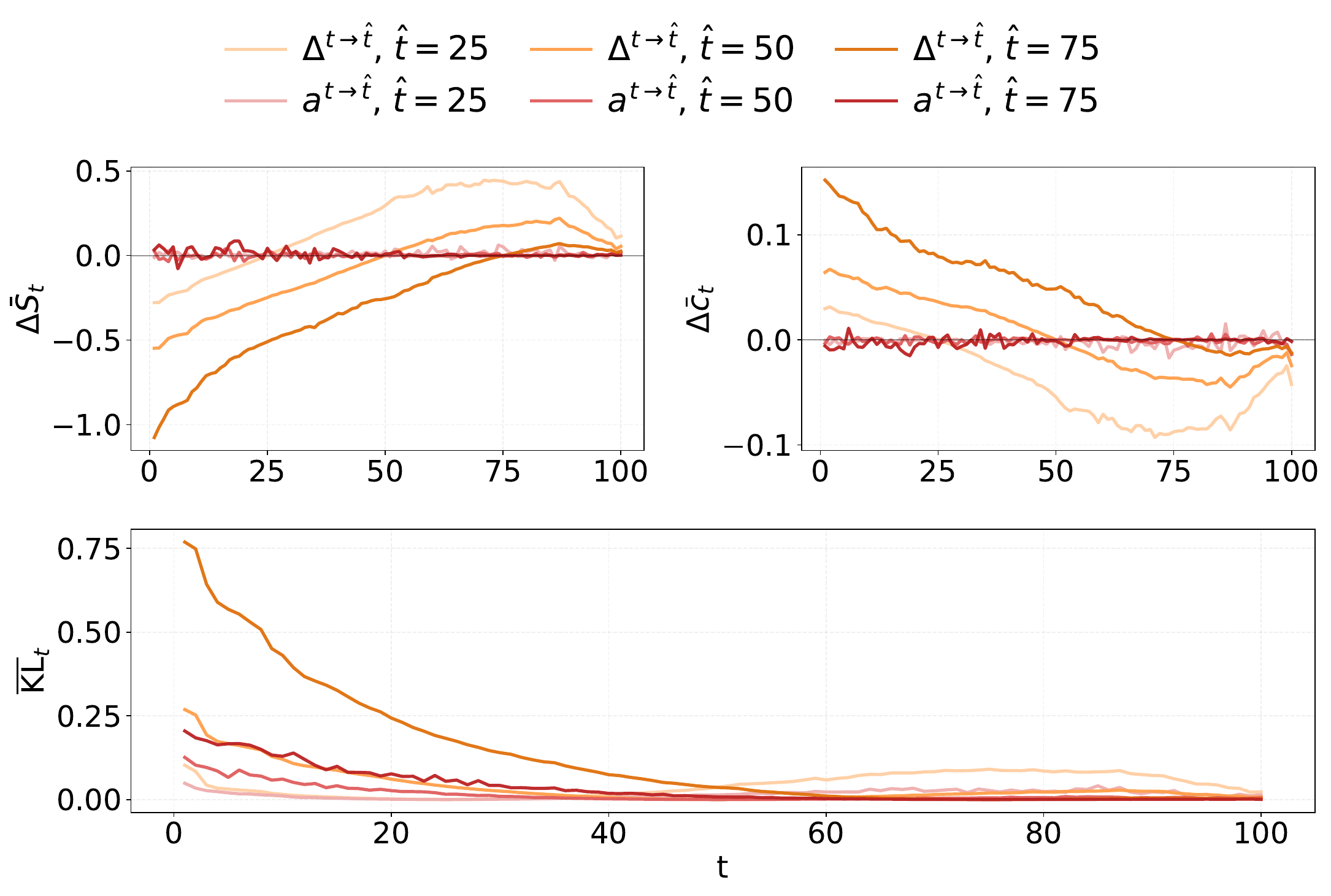}
    \end{subfigure}
    \caption{\textbf{Steering on layer 29 downstream effects}. We steer the activations of layer 29 using \Cref{eq:steer_mean_switch} targeting different $\hat{t}$ values (blue), and compare the resulting entropy, confidence, and KL-divergence variations against norm-matched \Cref{eq:random_steer} perturbations (red). Steering along the found $\tau$ directions produces systematic and interpretable effects.}
    \label{fig:mean_steering_llada_l29}
\end{figure}

We apply the steering vectors at different transformer layers and observe that the strongest downstream effects consistently emerge in the final layers of both \llada and \dream.
\Cref{fig:mean_steering_llada_l29} presents the results obtained when steering layer 29 of \llada. Similar trends are observed in \dream when steering layer 25. 
In \Cref{sec:probeSteeringAppendix}, we show results across various layers for both \llada and \dream.

Remarkably, steering with \Cref{eq:steer_mean_switch} produces a behaviour that closely matches what would be expected from directly modifying the model's internal notion of $\tau$. 
When steering towards larger values of $\hat{t}$ relative to the current step $t$, i.e.\ $\hat{t} - t > 0$, the model becomes more confident and its entropy decreases. Conversely, when $\hat{t} - t < 0$, confidence decreases and entropy increases. 
At the same time, the KL divergence grows approximately proportionally to the distance $|\hat{t} - t|$, indicating progressively larger deviations from the original distribution as the steering target moves further away from the current denoising step.

To verify that these effects are specific to the identified signal and are not a phenomenon of the denoising mechanics of these models, we compare them against \Cref{eq:random_steer} with matched norm.
Unlike the $\tau$-based perturbations, random perturbations do not induce coherent trends in entropy or confidence across denoising steps. 
Moreover, their downstream impact is substantially weaker. 
For a comparable perturbation norm, the KL divergence induced by random perturbations is approximately half of that produced by $\tau$ steering, suggesting that the learned directions correspond to a particularly sensitive and semantically meaningful subspace in the residual stream.
%
%
%
\begin{figure}
    \centering
    \includegraphics[width=0.7\linewidth]{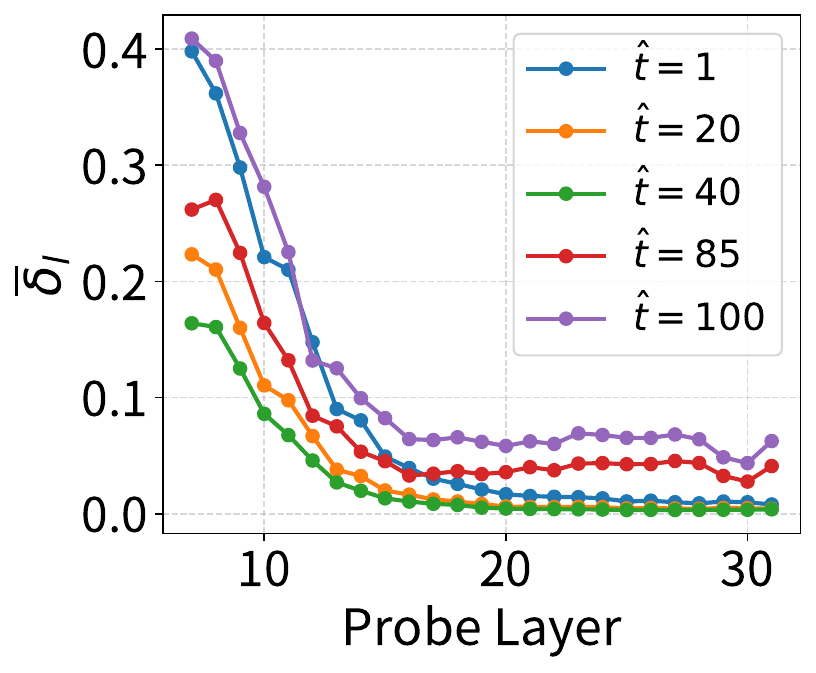}
    \caption{\textbf{Average probe drift across layers and steps after steering in \llada.} The model progressively compensates for the perturbation, reducing the discrepancy between clean and steered representations of $\tau$.}
    \label{fig:depth_correction}
\end{figure}
Contrary to the behaviour observed in the final layers, steering in shallow layers produces effects that are largely indistinguishable from the random baseline. Representative results for shallow- and mid-layer steering experiments, are reported in \Cref{sec:probeSteeringAppendix}.
Combined with the probe analysis in \Cref{fig:llada_mlp_sweep}, this suggests that $\tau$ information is not merely propagated unchanged across layers: early perturbations can be attenuated by later computation as the model progressively compensates for them.
We test this by injecting a steering perturbation at layer 6 and tracking the probe-predicted $\tau$ across the remaining layers.
%
%
\begin{figure*}[t]
    \centering

    \begin{subfigure}{0.49\textwidth}
        \centering
        \includegraphics[width=\linewidth]{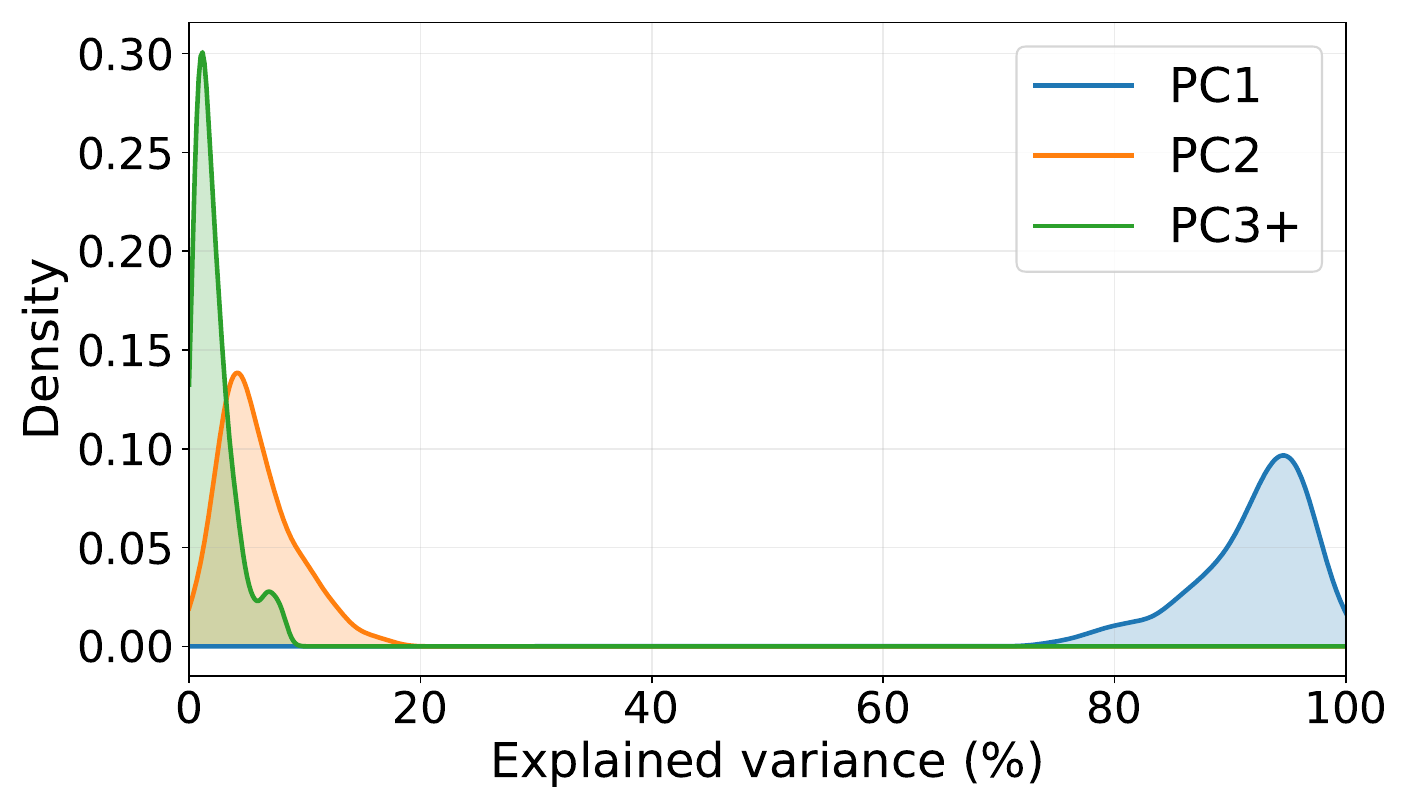}
        \caption{The PCA distribution of the mean vectors.}
        \label{fig:pca_distribution}
    \end{subfigure}
    \hfill
    \begin{subfigure}{0.37\textwidth}
        \centering
        \includegraphics[width=\linewidth]{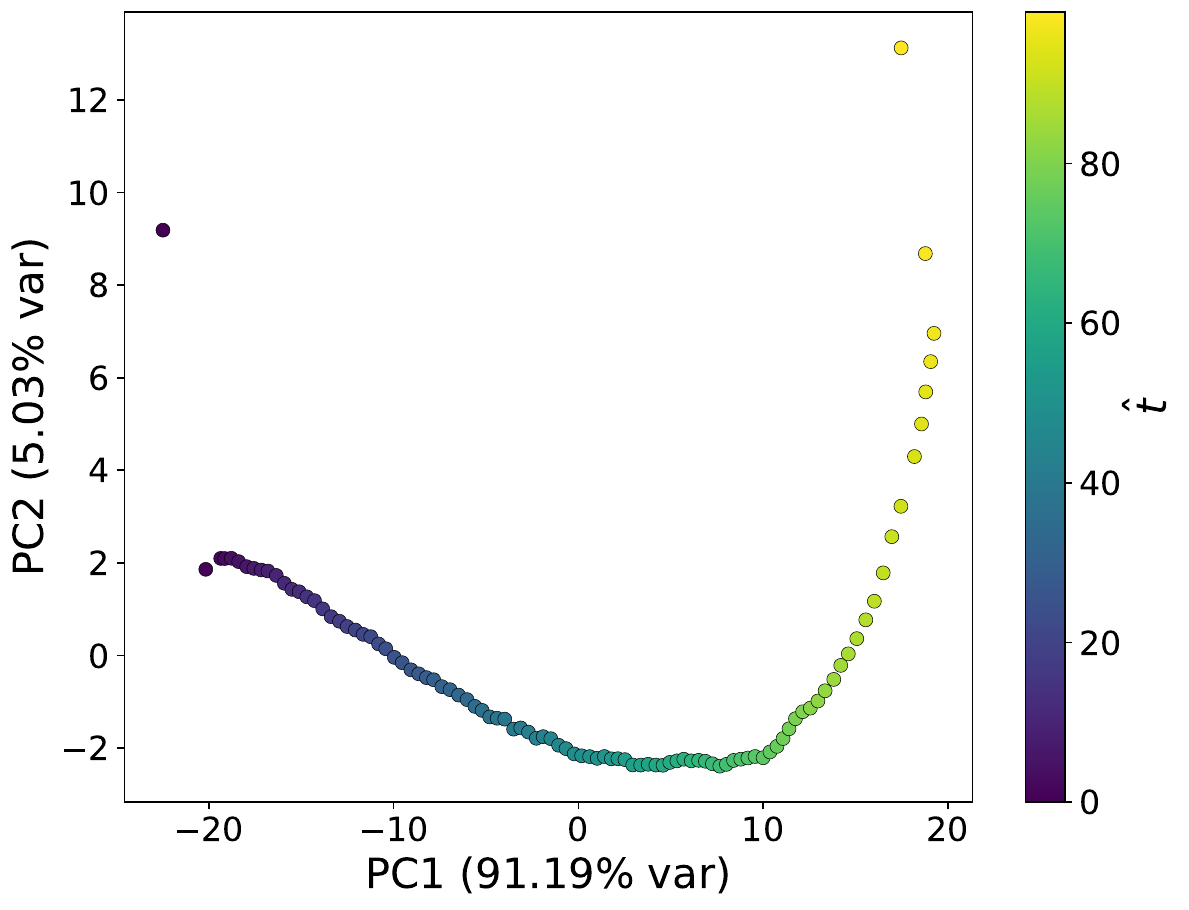}
        \caption{\dream's layer 25 2D projection}
        \label{fig:sample_2d_dream}
    \end{subfigure}

    \caption{\textbf{PCA distribution and sampled projection on the top-2 principal components of the mean vectors.}
    Left: PCA distribution of the mean vectors for \llada; notably, across layers, most of the variance of the identified set of mean vectors can be explained by fewer than three dimensions. Right: sampled projections for \dream. We use the mean vectors obtained from layer 25 of \dream. We observe a parabola-like geometry.}
    \label{fig:pca_full_width}
\end{figure*}

\subsection*{What happens to the signal at early layers?} \label{sec:reconstruction}
To explain the null result obtained when steering early layers, we investigate this phenomenon more deeply and measure how the probe predictions evolve after steering.
If the model does indeed correct an early perturbation as computation proceeds, then the $\tau$ value read by the probes should drift back towards its clean value at deeper layers; conversely, a perturbation that survives unchanged would keep the steered and clean predictions far apart throughout.
We therefore track the average deviation between the probe outputs of clean and steered activations across layers
\[
\bar\delta_l := \mathbb{E}_{t}\left[\left| \mathbb{E}_j\left[\phi_l(\tilde{h}^j_{t,l})-\phi_l(h^j_{t,l}) \right] \right| \right].
\]
Starting from a perturbation injected at layer 6, we evaluate how different steering targets $\hat{t}$ propagate through the networks' depth.

As shown in \Cref{fig:depth_correction}, the model is able to correct most of the perturbation introduced in early layers, such that the probe predictions under the perturbation become nearly indistinguishable from the clean ones in the final blocks.
However, steering toward extreme values such as $\hat{t} = 100$ remains harder to compensate for, leaving a persistent discrepancy even in deeper layers. 
This observation also helps explain the sharp peaks previously observed for shallow-layer steering (see \Cref{sec:probeSteeringAppendix}): at those denoising steps, the model is temporarily unable to fully recover from the injected perturbation before producing its prediction (additional analysis is provided in \Cref{sec:appendixdepthCorrection}).
Having established that this signal affects the models’ output distributions in a predictable way, we now move on to characterising its properties and look for the emergence of model-level patterns in how the signal is organised.

\begin{takeawaybox}{Takeaway for RQ2}{takeawaygreenbg}{takeawaygreen}
\emph{The found signal has direct causal implications for modelling dynamics, predictably affecting downstream entropy, confidence and KL divergence. The models also internally recompute it several times across their depth, allowing for correction.}
\end{takeawaybox}

\section{Characterising the Signal}
\label{sec:characterising_signal}
Having established how the models react to changes in the signal, we now examine how this signal is represented internally and whether consistent structural patterns emerge.

\subsection{\texorpdfstring{$\mu$}{mu} vectors' low-dimensional structure}
We begin by analysing the structural properties of the discovered mean vectors, focusing on the subspace spanned by these representations.
Specifically, we examine its effective dimensionality.
The resulting distribution of explained variance across principal components is shown in \Cref{fig:pca_distribution}. 
Surprisingly, a large fraction of the intra-layer variance can be captured by a single principal component, with the distribution heavily concentrated above $90\%$ explained variance.
Motivated by this observation, we project the mean vectors from selected layers into two- and three-dimensional PCA spaces to better visualise their structure. 
We show a sample 2D and 3D projection in \Cref{fig:sample_2d_dream} and \Cref{fig:example_3d_llada}, respectively. 
Interestingly, we find the 3D shape to be closely related to structures that models have been shown to exhibit when operating on counting or time-dependent and sequentially ordered tasks \citep{gurnee2025counting, engels2025not, karkada2026symmetry, modell2025origins}, characterised by a continuous and ordered representation of the time-axis being analysed (in our case $\tau$).

\begin{figure}[t]
    \centering
    \includegraphics[width=\columnwidth]{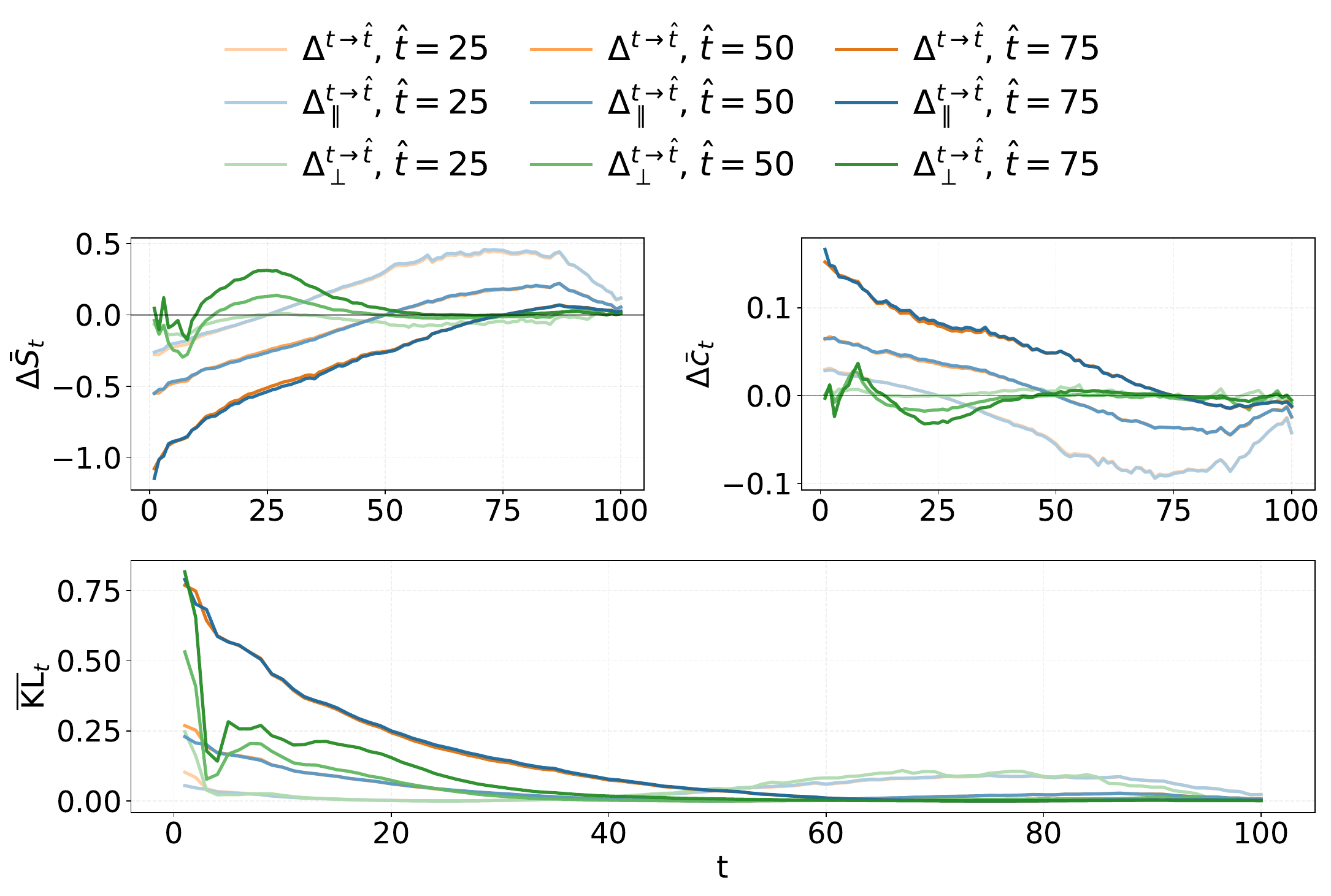}
    \caption{\textbf{Low-dimensional subspace steering in \llada.} Using \Cref{eq:subspace_mean_distortion}, we steer the model within the two-dimensional subspace ($k=2$) spanned by the top principal components of the layer-29 mean vectors. Steering within the subspace closely resembles the unrestricted one, while the orthogonal perturbation produces incoherent effects.}
    \label{fig:example_llada_subspace_steering}
\end{figure}

To test whether the functional effect of $\tau$ steering is concentrated in this low-dimensional subspace, we repeat the intervention of \Cref{eq:steer_mean_switch}, but restrict the steering vector $\Delta_l^{t \to \hat{t}} := \mu_{\hat{t},l} - \mu_{t,l}$ to a chosen set of principal directions. 
Concretely, we isolate the part of $\Delta_l^{t \to \hat{t}}$ lying in the subspace spanned by the first $k$ principal components and the part orthogonal to it, rescaling each to match
the norm of the original steering vector:
\begin{equation} \label{eq:subspace_mean_distortion}
    \begin{aligned}
        \Delta_{\parallel,l}^{t \to \hat{t}} &:= P_{\parallel,l}\, \Delta_l^{t \to \hat{t}}\,
        \frac{\| \Delta_l^{t \to \hat{t}} \|}
        {\| P_{\parallel,l}\, \Delta_l^{t \to \hat{t}} \|}
        \\
        \Delta_{\perp,l}^{t \to \hat{t}} &:= P_{\perp,l}\, \Delta_l^{t \to \hat{t}}\,
        \frac{\| \Delta_l^{t \to \hat{t}} \|}
        {\| P_{\perp,l}\, \Delta_l^{t \to \hat{t}} \|}
    \end{aligned}
\end{equation}
Here $P_{\parallel,l}$ denotes the orthogonal projection onto the subspace
spanned by the first $k$ principal components of layer $l$, and
$P_{\perp,l} = I - P_{\parallel,l}$ the projection onto its complement.
Because the top components capture most of the variance
(\Cref{fig:pca_distribution}), we have
$\| P_{\parallel,l}\, \Delta_l^{t \to \hat{t}} \| \gg
 \| P_{\perp,l}\, \Delta_l^{t \to \hat{t}} \|$; rescaling both perturbations to
$\| \Delta_l^{t \to \hat{t}} \|$ therefore isolates the effect of the
steering \emph{direction} from that of its \emph{magnitude}.
As shown in \Cref{fig:example_llada_subspace_steering}, projecting onto just the first two principal components recovers the same behaviour as the full, unprojected perturbation.
Interestingly, the orthogonal perturbation $\Delta_{\perp,l}^{t \to \hat{t}}$ produces incoherent effects at the same norm, implying that those directions carry little information of the $\tau$ representation.
We present more results on \Cref{sec:appendixsubspacesteering}.
These results motivate a closer examination of whether this geometry can be consistently described across layers and aggregated into a unified, model-level representation.

\subsection{A shared representation across layers}
From the observation that many layers exhibit a similar parabolic shape, we investigate whether a general model-level 2D representation of the mean vector components exists. 
In particular, we aim to model a 2D trajectory $f(t): [100] \to \mathbb{R}^2$ that describes the geometry of the top two principal components across the entire model.

\begin{figure}[t]
    \centering
    \includegraphics[width=\columnwidth]{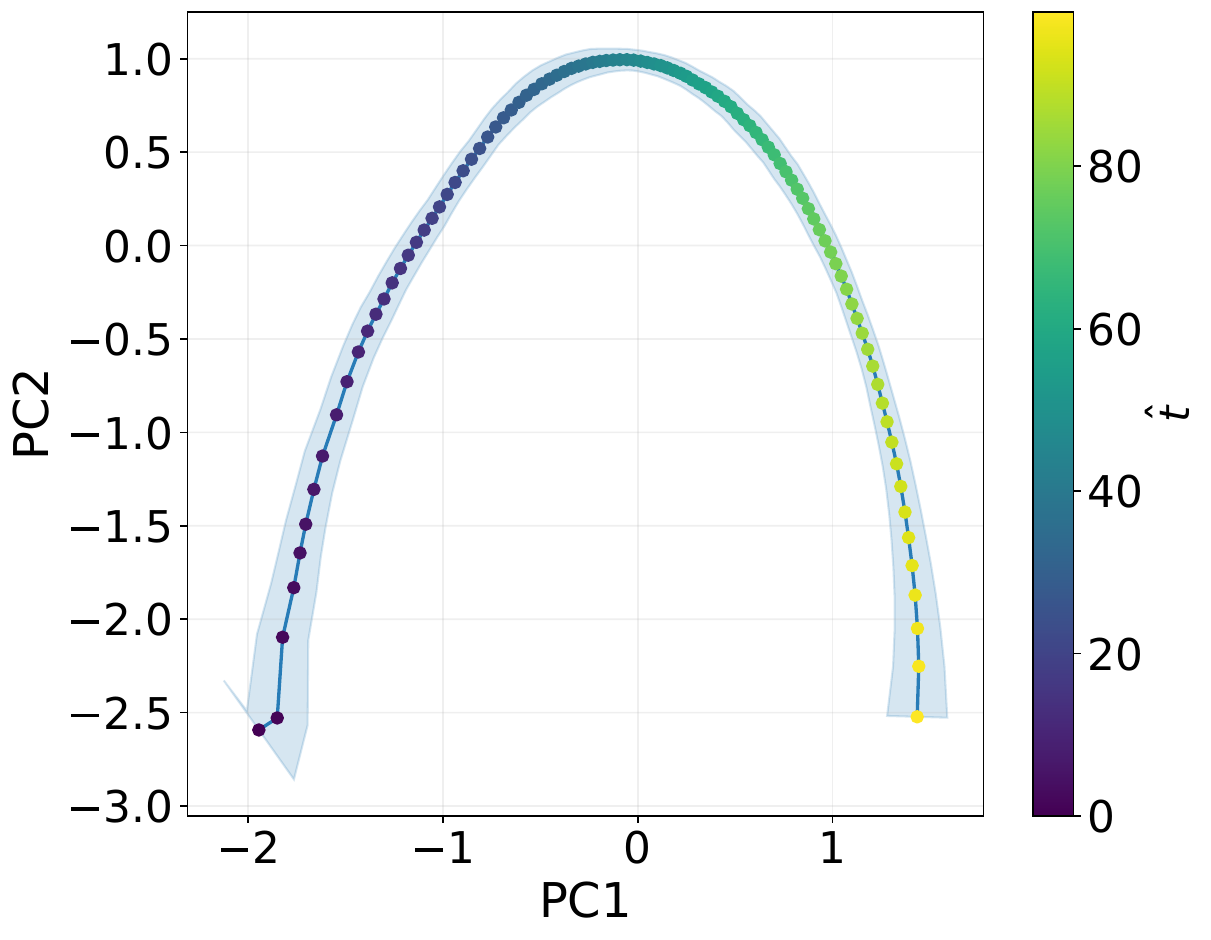}
    \caption{\textbf{The shared 2-dimensional geometry of $\hat{t}$ in \llada.}
    We compute an average trajectory by taking the standardised 2D PC projections of the mean vectors across the model. The parabolic geometry is shared across layers. PC1 and PC2 are standardised to unit variance.}
    \label{fig:canonical_2d}
\end{figure}

To this end, we collect the 2D projections from all layers and standardise them to have zero mean and unit variance. We then compute, for each $t \in [100]$ (corresponding to the 100 mean vectors projected per layer), the empirical mean of the 2D points associated with the mean vectors binned by $t$. This yields a mean-standardised projection across layers.
From this procedure, we obtain a non-parametric 2D trajectory that represents the standardised expectation of the projections for each $\tau$ index. 
We report the resulting plot in \Cref{fig:canonical_2d} for \llada (see \Cref{sec:dreamAnalysisAppendix} for similar results on \dream). Remarkably, all points closely follow the proposed shared trajectory, with error bars remaining within approximately $0.1$, indicating that the proposed approximation faithfully captures the general model-level geometry of the subspace. This further implies that the principal components of the mean vector distributions follow the same parabolic shape up to scaling and offset factors.

While this construction reveals a stable and coherent 2D structure shared across layers, it remains unclear whether the underlying representation is consistently encoded throughout the depth of the model or whether it emerges from layer-specific implementations. We next address this question by studying the cross-layer alignment of the discovered mean vectors.
\subsection{Cross-layer \texorpdfstring{$\tau$}{tau} representations}
Based on these results, we further investigate how the discovered mean vectors are connected across layers. In particular, we ask whether the model maintains a consistent direction-wise representation of a given $\mu_t$ vector across depth, or instead develops layer-local representations that do not generalise.
\begin{figure}
     \begin{subfigure}{0.45\textwidth}
        \includegraphics[width=\textwidth]{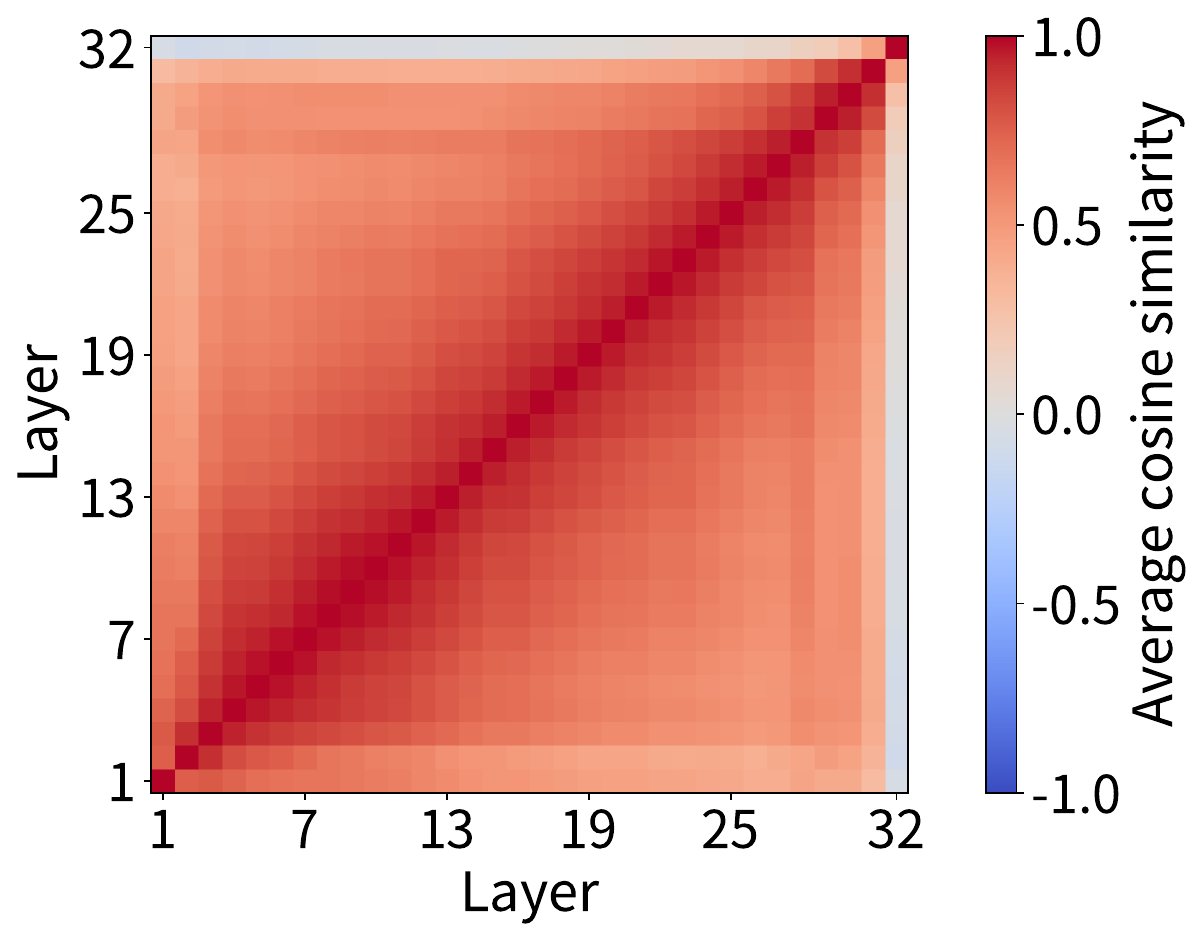}
    \end{subfigure}
     \caption{\textbf{Average cosine similarity of same-indexed vectors across layers in \llada.} We observe that most layers maintain a highly correlated representation of the same index $t$, and that this relationship degrades with distance. Layer 32 instead maintains a representation that is largely independent of the other layers.}
     \label{fig:llada_mean_similarity}
\end{figure}
To address this question, we compute the average pairwise cosine similarity between all centred vectors encoding a given index $t$ across different layers. We report the results in \Cref{fig:llada_mean_similarity} for \llada and in \Cref{sec:dreamAnalysisAppendix} for \dream. 

Notably, we observe a clear structural difference between the two models in how they preserve representations of the same mean vector across depth. In \llada, most layers exhibit strong alignment in their representations, with the exception of layer 32, which appears nearly orthogonal to the others. In contrast, \dream shows a more heterogeneous organisation: representations are only consistently aligned within specific blocks of layers, while layers outside these blocks are largely uncorrelated with those inside them.
This suggests that there may be a model-level basis for encoding $\tau$ across layers, as having high correlation implies that the downstream effects would be similar if interchanging the mean vectors from those layers.
%
%
%
%
Although the previous sections establish the existence of structured and consistent $\tau$ representations across layers, they do not reveal how these representations are formed within the computations of each layer, which we investigate in the next section.
\subsection*{How is $\tau$ represented within a layer?} \label{sec:internal_representation}
\begin{figure}
    \centering
    \includegraphics[width=1\linewidth]{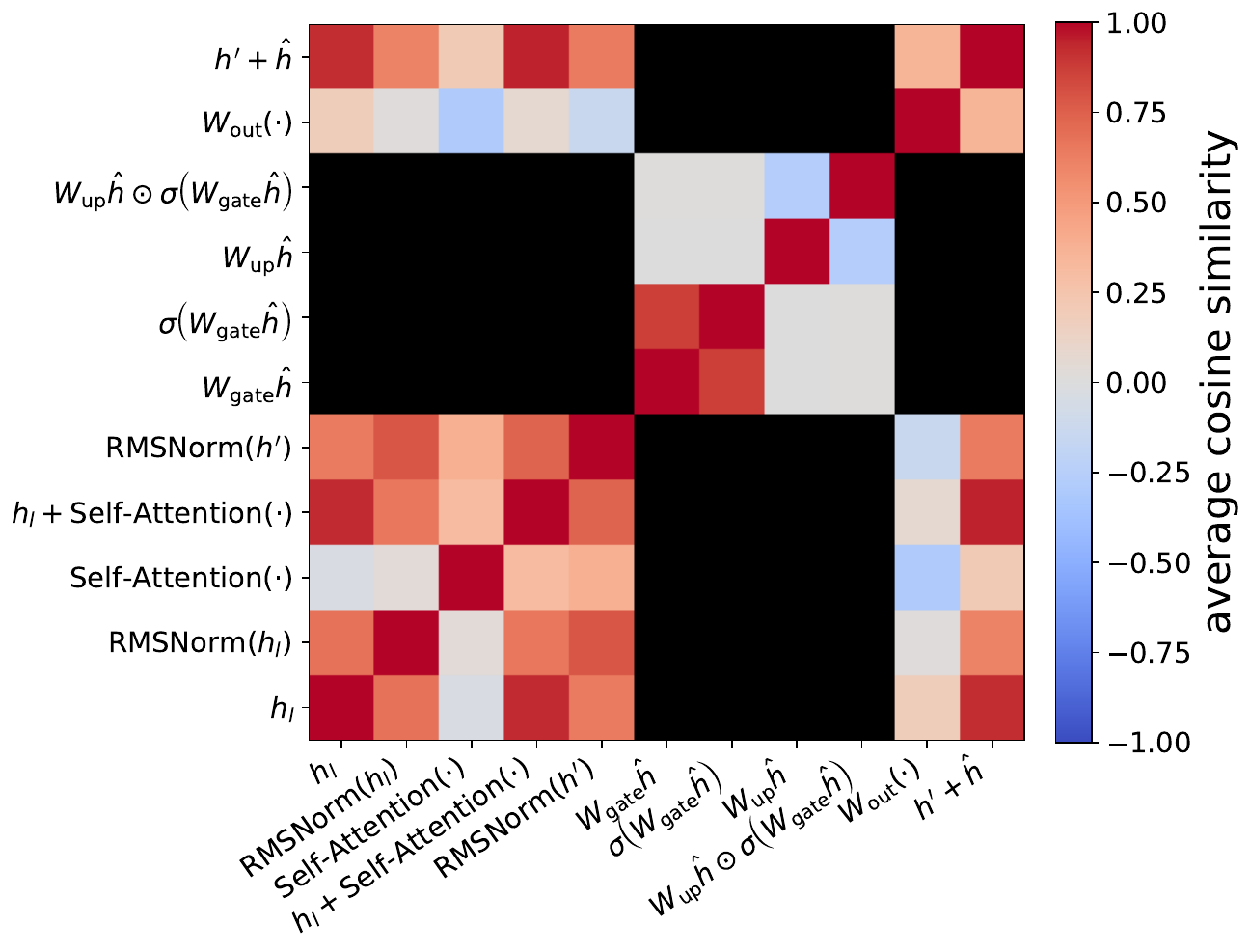}
    \caption{\textbf{Average cosine similarity across layer components in \llada.} Where $h' := h_l + \text{Self-Attention}(\cdot)$, and $\hat{h} := \text{RMSNorm}(h')$. We clearly see how across most of the internal representations there is high-correlation or almost complete orthogonality. We note how self-attention and the outputs of the out matrix seem to be anticorrelated.}
    \label{fig:llada_intermediates_alignment}
\end{figure}

\begin{figure}
     \begin{subfigure}{0.45\textwidth}
        \includegraphics[width=\textwidth]{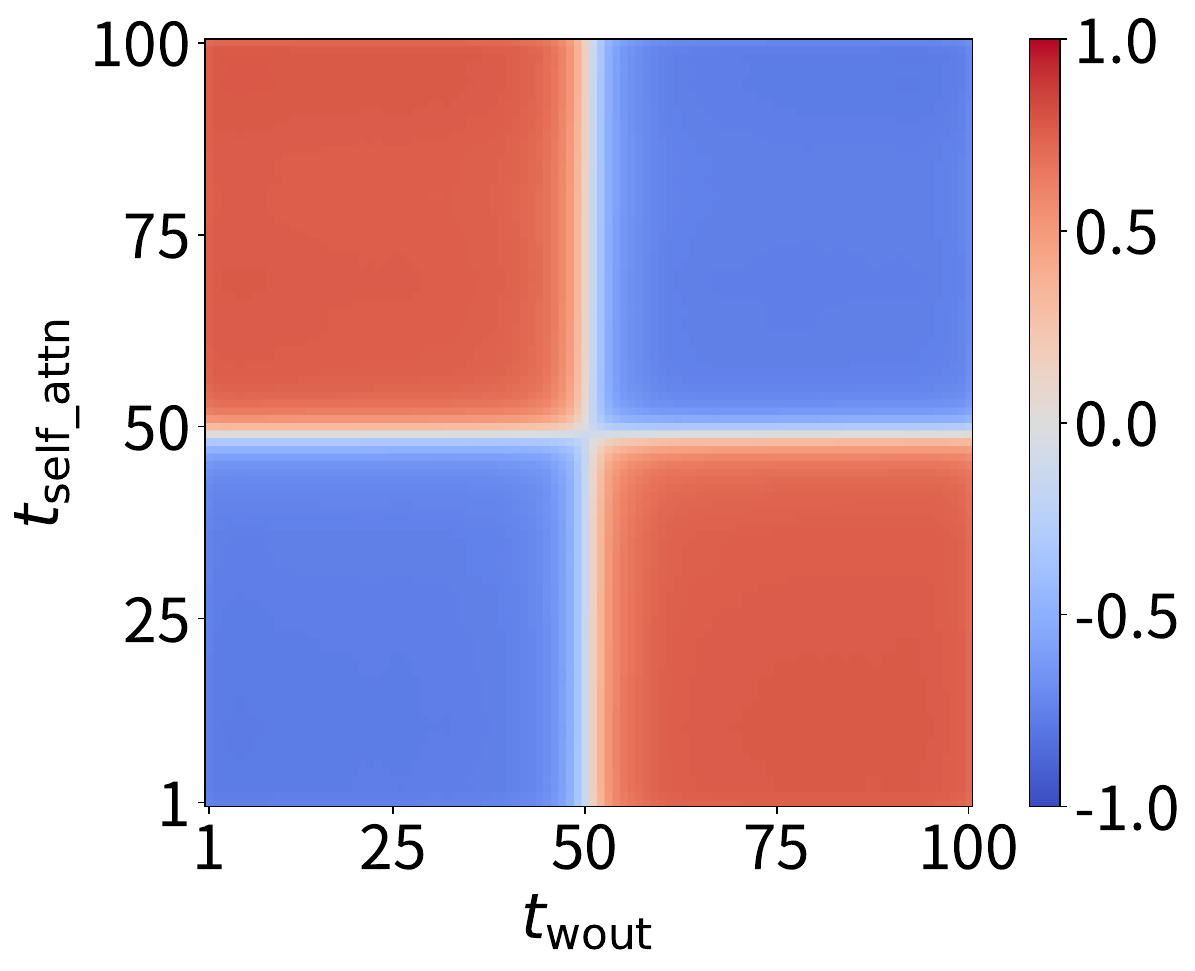}
    \end{subfigure}
     \caption{\textbf{Average cosine similarity of self-attention and post-$W_\text{out}$ vectors.} Layer-29 vectors from \llada reveal a clear geometric organization over $t$: representations for distant indices become nearly anti-aligned, and a transition region emerges around $t=50$, where vectors are approximately perpendicular.}
     \label{fig:llada_mlp_attn_heatmap}
\end{figure}

To query the models' internal representation of the mean vectors we employ the same approach as proposed in \Cref{mean_activation}, but now capture activations inside the models' intermediate layer computations.
With these mean vectors, we measure the cosine similarity of all same-dimensional representations inside the models' layers (as the MLP has a different hidden dimension compared to the rest of the model). We present the results for \llada at \Cref{fig:llada_intermediates_alignment} (and \Cref{sec:dreamAnalysisAppendix} for \dream).
As the results suggest, most operations before and after the MLP remain highly correlated; nonetheless, inside the MLP, the mean vectors produce almost completely independent representations of $\tau$, as for both models they are orthogonal in expectation (with the exception of the post up-projection matrix and post activation function, which remain highly correlated).
We note a particular pattern in both models between the post-MLP and post-attention similarities, as both are highly anti-correlated, sharing a common direction but pointing in opposite ways. This phenomenon can be better appreciated in \Cref{fig:llada_mlp_attn_heatmap}. 
Interestingly, the two stages agree on the distance from the centre of a $\tau$ value but not on its sign. We hypothesise that this seeming contradiction in the mean $\tau$ representation may help the model correct the signal as shown in \Cref{sec:reconstruction}.
\begin{takeawaybox}{Takeaway for RQ3}{takeawaypurplebg}{takeawaypurple}
\emph{The models represent the mean vectors subspace in an ordered and manifold-like manner. Most layers in \llada share common semantics for this space, while \dream is organised into representation blocks. Finally, self-attention and the MLP have opposite representations.}
\end{takeawaybox}

\section{Related Work}
\textbf{LLM Interpretability}. Prior work studies activation subspaces and representation geometry to explain LLM behaviour. Closest to us, \citet{gurnee2025counting} use mean activation vectors to analyse and steer a character counting task. We similarly use mean vectors, but study a distributed denoising-time signal and its propagation across layers.
\citet{engels2025not}, \citet{modell2025origins}, and \citet{karkada2026symmetry} analyse concept and token-level representations in language models, describing low-dimensional manifold geometries that, in the case of ordered concepts such as historical years, resemble the one we find for $\tau$. 
In contrast to their work, we study a sequence-level statistic that the model reconstructs as "meta-level" information to track denoising progress, rather than a human-like concept (such as years or months) whose geometry has been tied to statistical co-occurrences.

\textbf{DLM Denoising Dynamics and Interpretability}. Recent DLM work has examined both steering and internal behaviour. \citet{zhou2026steering} find that concepts are steerable only at particular denoising stages; in contrast, our $\tau$-related signal persists throughout diffusion and is repeatedly recomputed. \citet{rossi2026diffusion} study DLM length-awareness through downstream performance, whereas we characterise an internal representation tied to denoising progress.

\section{Conclusion and Discussion}
We characterise a denoising-progress-related signal in the residual streams of DLMs. The signal is probe-decodable across layers and token types, causally steerable through mean vector directions, repeatedly corrected across depth, and represented in a structured low-dimensional subspace. Steering this representation produces predictable changes in entropy, confidence, and KL divergence, indicating that the recovered directions are not merely descriptive but functionally relevant to the model’s computation.

A key implication of our findings is that probes or auxiliary models operating on DLMs hidden states need not be explicitly conditioned on a timestep embedding to access denoising-progress information. As the probes show, this information is already intrinsically encoded in the residual stream and can therefore be recovered directly from the activations on which such methods operate.

A natural next question is how this signal is computed. Our results show that DLMs internally track information closely related to the fraction of unmasked tokens, but they do not identify the circuit that constructs or updates this representation. Future work could therefore investigate whether the signal is derived from explicit mask-ratio estimation, from distributed sequence-level statistics, or from interactions between attention and MLP components. Understanding this computation would better clarify how DLMs operate in this subspace and enable better architecture designs for these models.

\section{Limitations}
Although our work aims to characterise and explain the representation of $\tau$ inside DLMs, several open questions remain. 
First, our analysis is restricted to \llada and \dream, two DLMs trained with the same cross-entropy loss over \mask tokens; it therefore remains unexplored how the $\tau$ phenomenon emerges in other DLMs that similarly lack explicit timestep conditioning, such as block-diffusion models \citep{cheng2025sdar, arriola2025block, fastdlm2}. 
Moreover, it is unclear whether the identified $\tau$ signal can be exploited at inference time to enable more efficient decoding---or even remasking---strategies, which we leave as a promising direction for performance improvements. 
Another promising direction is studying small DLMs trained from scratch; as it could help determine which properties of the \(\tau\) representation emerge early in training or at small scale, and which depend on model depth or capacity.
Finally, the token-level effects of steering remain unexplored: our analyses focus on sequence-level statistics rather than examining which tokens change, and how, as the signal is steered.

\section{Acknowledgements}
This work was supported by a grant from Coefficient Giving, administered by the Berkeley Existential Risk Initiative (BERI), and by Sapienza grant RG123188B3EF6A80 (CENTS). We acknowledge ISCRA for awarding this project access to the LEONARDO supercomputer, owned by the EuroHPC Joint Undertaking and hosted by CINECA (Italy). We thank Fastweb S.p.A. for providing the computational resources used in this paper. We also thank Donatella Genovese and Francesco Piccolo for helpful discussions and valuable feedback.

We used LLMs only for minor grammar, wording, and readability checks. We reviewed and edited the final manuscript and take full responsibility for its content.

\section{Ethical considerations}
This work analyses the internal representations of publicly released diffusion language models (\llada and \dream) and does not involve human subjects, new data collection, or the release of new models. We foresee no ethical concerns arising from this work.

\bibliography{custom}

\clearpage          
\onecolumn
\appendix


\section{Denoising Progress and the Training Noising Level}
\label{sec:timetauAppendix}

Here we provide more details about DLMs' training objectives and clarify the relationship between the training noising level \(s\) and the empirical denoising-progress variable \(\tau_t\) used throughout the paper. For more details about these models' formulations we refer the reader to \citet{llada} and \citet{dream}.
In absorbing-mask diffusion language models, the fraction of response tokens that remain masked is an empirical analogue of the training noising level, while the fraction of response tokens that have been unmasked measures denoising progress.

\paragraph{Absorbing-mask corruption.}
Let \(x_0^j\) denote the clean token at position \(j\), represented as a one-hot vector. We write \mask{} both for the absorbing mask token and, when used inside a categorical parameter, for its corresponding one-hot vector. 
At noising level \(s \in [0,1]\), each token is independently replaced by \mask{} with probability \(s\). For a single position \(j\),

\[
q^j_{s \mid 0}(z \mid x_0^j)
=
\begin{cases}
1-s, & z = x_0^j, \\
s, & z = \mask, \\
0, & \text{otherwise}.
\end{cases}
\]

Equivalently,
\[
q^j_{s \mid 0}(\cdot \mid x_0^j)
=
\operatorname{Cat}
\bigl(\cdot ; (1-s)x_0^j+s\mask\bigr).
\]

Assuming independent corruption across positions, the full corrupted
sequence distribution factorises as

\[
q_{s \mid 0}(\cdot \mid x_0)
=
\prod_{j}
q^j_{s \mid 0}(\cdot \mid x_0^j).
\]

Thus, \(s\) is the probability that a token is masked, and \(1-s\) is the
probability that it remains visible.

\paragraph{Training objectives.}
For a corrupted sequence \(x_s\), define the masked-position set

\[
\mathcal{M}(x_s) := \{j : x_s^j = \mask\}.
\]

The pretraining objective can then be written compactly as

\begin{equation}
\label{eq:pretraining_objective}
\mathcal{L}_{\mathrm{pre}}
=
-
\mathbb{E}_{x_0,s,x_s}
\left[
\frac{1}{s}
\sum_{j \in \mathcal{M}(x_s)}
\log p_{\theta}(x_0^j \mid x_s)
\right].
\end{equation}

During supervised fine-tuning, the prompt is kept fixed and only response tokens are corrupted. Let \(p_0\) denote the prompt, \(r_0\) the clean response, and \(r_s\) the corrupted response. Define

\[
\mathcal{M}(r_s) := \{j : r_s^j = \mask\}.
\]

The SFT objective is

\begin{equation}
\label{eq:sft_objective}
\mathcal{L}_{\mathrm{SFT}}
=
-
\mathbb{E}_{p_0,r_0,s,r_s}
\left[
\frac{1}{s}
\sum_{j \in \mathcal{M}(r_s)}
\log p_{\theta}(r_0^j \mid p_0,r_s)
\right].
\end{equation}

Both objectives therefore train the model on partially masked sequences,
where the expected fraction of masked tokens is controlled by \(s\).

\paragraph{Inference-time denoising progress.}
At inference time, the model does not sample a corrupted sequence directly
from \(q_{s \mid 0}\). Instead, it starts from a fully masked response and
progressively unmasks tokens over discrete denoising steps
\(t \in \{0, \ldots, T\}\).

Let \(R\) be the set of response positions, with \(L := |R|\). For a
realised inference state \(x_t\), define the response-local masked and
unmasked sets as

\begin{equation}
\label{eq:masked_unmasked_sets}
\begin{aligned}
\mathcal{M}_R(x_t) &:= \{j \in R : x_t^j = \mask\}, \\
\mathcal{U}_R(x_t) &:= \{j \in R : x_t^j \neq \mask\}.
\end{aligned}
\end{equation}

These sets partition the response positions, so

\[
|\mathcal{M}_R(x_t)| + |\mathcal{U}_R(x_t)| = L.
\]

We define the empirical denoising progress at step \(t\) as

\begin{equation}
\label{eq:tau_definition}
\tau_t
:=
\frac{|\mathcal{U}_R(x_t)|}{L}.
\end{equation}

Its complement is the remaining response-local mask ratio:

\begin{equation}
\label{eq:local_mask_ratio}
1-\tau_t
=
\frac{|\mathcal{M}_R(x_t)|}{L}.
\end{equation}

Thus, \(\tau_t=0\) corresponds to a fully masked response, while \(\tau_t=1\) corresponds to a fully unmasked response.

\paragraph{Connection to the training noising level.}
Under the SFT corruption process, each response token is masked independently with probability \(s\). Therefore, for
\(r_s \sim q_{s \mid 0}(\cdot \mid r_0)\),

\begin{equation}
\label{eq:expected_mask_ratio}
\mathbb{E}_{r_s}
\left[
\frac{|\mathcal{M}(r_s)|}{L}
\right]
=
s.
\end{equation}

Similarly, if \(\mathcal{U}(r_s) := \{j : r_s^j \neq \mask\}\), then

\begin{equation}
\label{eq:expected_unmask_ratio}
\mathbb{E}_{r_s}
\left[
\frac{|\mathcal{U}(r_s)|}{L}
\right]
=
1-s.
\end{equation}

Consequently, the realised inference-time mask ratio \((1-\tau_t)\) plays the role of an effective noising level, while \(\tau_t\) measures effective denoising progress.

\paragraph{Interpretation.}
The equivalence above should be understood at the level of mask ratios, not as an equality between the training and inference distributions. During training, \(s\) indexes an independently sampled corruption process. During inference, the sequence of states is generated by the model and its unmasking policy. Nevertheless, both procedures produce partially masked responses, and each such response has a well-defined response-local mask ratio.

We therefore use \(\tau_t\) as an empirical measure of denoising progress throughout the paper. Its complement, \(1-\tau_t\), is the realised fraction of response tokens that remain masked and is directly analogous to the training noising level \(s\). Recovering \(\tau_t\) from the residual stream can thus be interpreted as recovering the model's internal representation of its position along the denoising trajectory.

\section{Probe Architectural and Training Details} \label{sec:probetrainAppendix}
We use a 5-layer residual neural network with LayerNorm+GELU blocks, shown in \Cref{fig:probe_arch}, and bound the outputs to the interval $(0, 1)$ by applying a sigmoid function to the MLP logits, since the target $\tau$ is itself bounded. We train the probes for 20 epochs using 300 training examples and 100 validation examples. For each example and at each epoch, we dynamically vary the generation length and number of steps to improve the probe’s coverage. We implement batch gradient descent by updating the parameters after each full example, i.e., by batching the gradients over the entire denoising stage. We employ the AdamW \cite{loshchilov2017decoupled} weight optimiser, setting the learning rate to $\alpha = 10^{-3}$ and the weight decay coefficient to $6 \times 10^{-6}$. All experiments introduced earlier were conducted on NVIDIA H100 and A100 GPUs.

\begin{figure*}[t]
\centering
\begin{minipage}{0.62\textwidth}
\begin{tcblisting}{
    listing only,
    listing style=torch,
    enhanced, arc=2mm,
    colback=codebg, colframe=codeframe, boxrule=0.6pt,
    left=2mm, right=2mm, top=1mm, bottom=1mm,
}
def block(d_in, d_out):
    return nn.Sequential(
        nn.Linear(d_in, d_out),
        nn.LayerNorm(d_out),
        nn.GELU(),
    )

class MLPProbe(nn.Module):
    def __init__(self, dim, bounded=True):
        super().__init__()
        w = min(dim, 1024)
        self.inp = block(dim, w)
        self.middle_in = block(w, w)
        self.expansion = nn.Sequential(
            block(w, 2 * w),
            block(2 * w, w),
        )
        self.middle_out = block(w, w)
        self.out = nn.Linear(w, 1)
        self.bounded = bounded

    def forward(self, x):
        h = self.middle_in(self.inp(x))
        e = self.expansion(h)
        h = self.middle_out(e) + h   # residual
        out = self.out(h)
        if self.bounded:
            return out.sigmoid()
        return out
\end{tcblisting}
\end{minipage}
\caption{\textbf{MLP probe architecture.} Each \texttt{block} is a $\mathrm{Linear}\!\to\!\mathrm{LayerNorm}\!\to\!\mathrm{GELU}$ unit; the probe stacks five such blocks around a single residual connection, and a sigmoid head bounds the output to $(0,1)$ to match the range of $\tau$. The hidden width $w$ is capped at $1024$.}
\label{fig:probe_arch}
\end{figure*}

\section{Additional Experiments on Probing and Steering \texorpdfstring{$\tau$}{tau}}
\label{sec:probeSteeringAppendix}

In this section, we present additional results concerning the probes' capabilities to correctly identify the value of $\tau$, followed by further analyses of the mean activation vectors' ability to steer the model. The first part extends the probing analysis to \dream and to linear probes; the second broadens the steering analysis to additional layers, generation lengths, and downstream tasks.

\begin{figure*}[t]
    \includegraphics[width=\textwidth]{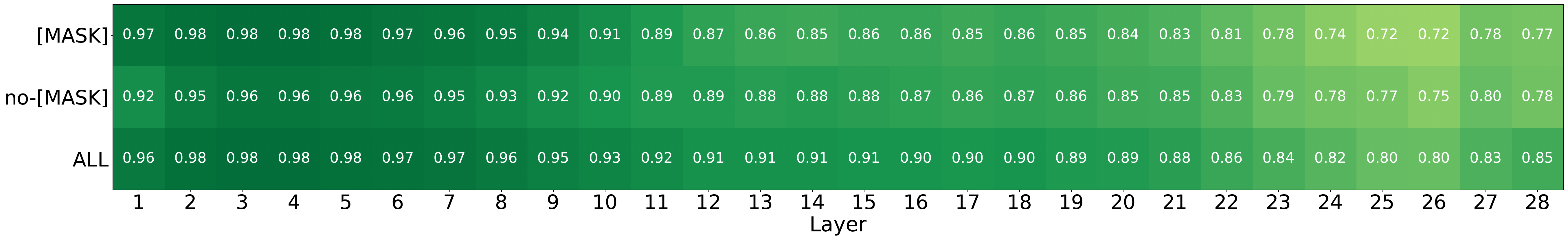}
    \caption{\textbf{$\mathbf{R}^2$ coefficients of MLPs in \dream}. The \rsq coefficient slightly degrades as we probe deeper into the model, which remains consistent with the observations of \Cref{fig:llada_mlp_sweep}.}
    \label{fig:dream_mlp_sweep}
\end{figure*}

\paragraph{MLP probes on \dream.}
\Cref{fig:dream_mlp_sweep} shows the \rsq obtained using the MLP probe over \dream. Similarly to what was observed with \llada, the probes' accuracy is high across the whole model but drops in the deeper layers.

\begin{figure*}[t]
    \begin{subfigure}{\columnwidth}
        \centering
        \includegraphics[width=\linewidth]{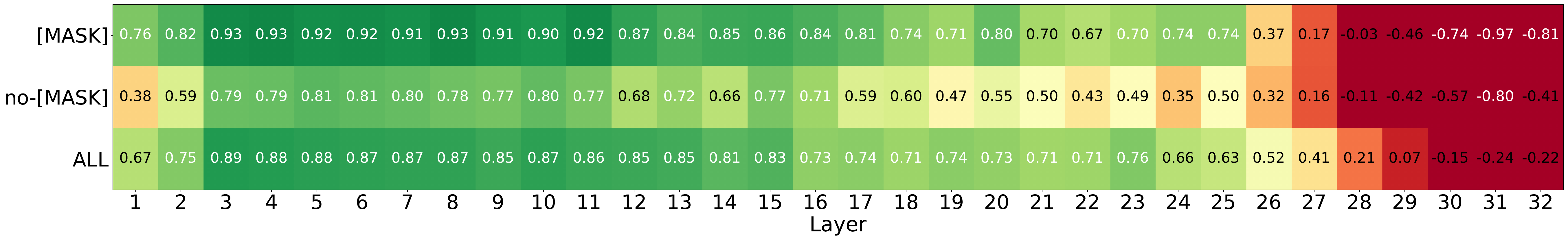}  
    \end{subfigure}
    \begin{subfigure}{\columnwidth}
        \centering
        \includegraphics[width=\linewidth]{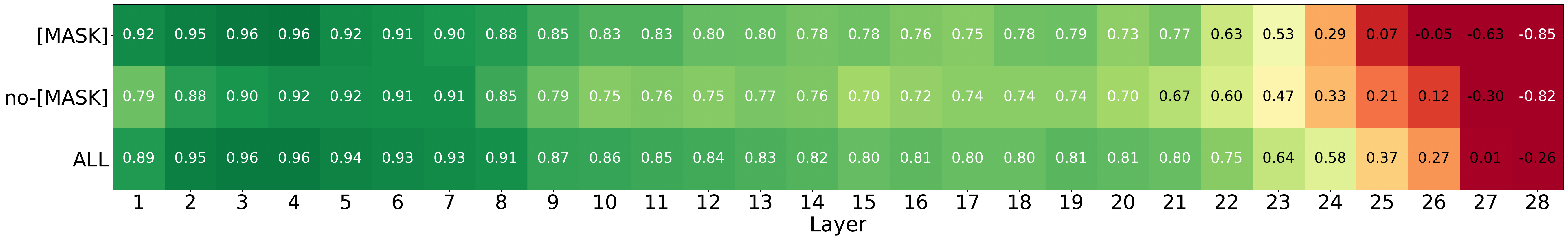}  
    \end{subfigure}
    \caption{\textbf{Linear probe performance}. The linear probes degrade markedly as we probe deeper into the model---more so than the MLP probes shown in \Cref{fig:llada_mlp_sweep} and \Cref{fig:dream_mlp_sweep}---and fall to the mean-$\tau$ baseline in the final layers for both \llada (top) and \dream (bottom). This indicates that, at the token level, $\tau$ information becomes increasingly hard to decode linearly with depth, while an MLP still recovers much of it until the last few layers.}
    \label{fig:linear_sweep}
\end{figure*}

\paragraph{Linear probes.}
We also train linear probes with the same sigmoid output and regression objective as the MLPs (\Cref{fig:probe_arch}), reporting the results in \Cref{fig:linear_sweep}.
Both models' linear probes track $\tau$ well through the early and middle layers, but their accuracy then falls off more steeply than the MLP probes as we move deeper, and in the final layers it collapses essentially to the baseline of predicting the mean $\tau$.
Across the broad mid-to-deep range the MLP probes retain markedly higher \rsq (\Cref{fig:llada_mlp_sweep}, \Cref{fig:dream_mlp_sweep}), so for those layers we read the gap not as the $\tau$ signal disappearing at the token level, but as it becoming less linearly accessible: the information is still present, yet increasingly encoded in a form a single linear readout cannot capture. Only at the very last layers, where the MLP performance also degrades, does the token-level signal become genuinely hard to recover by either probe.
The choice of training tokens also matters more here than for the MLPs: for \llada, restricting training to non-\mask tokens opens a noticeable gap relative to \mask or all tokens---consistent in direction with the smaller \mask advantage seen for the MLP probes, but more pronounced---whereas \dream is mostly insensitive to the token subset.

We stress that this does not contradict the low-dimensionality of $\tau$ established earlier: that structure is a property of the mean vectors, which average over many tokens and examples and thereby expose sequence-level regularities to which an individual token's activation---and hence a token-level linear probe---has no comparable access.

\begin{figure}[!h]
    \centering
    \begin{subfigure}{0.6\textwidth}
        \includegraphics[width=\textwidth]{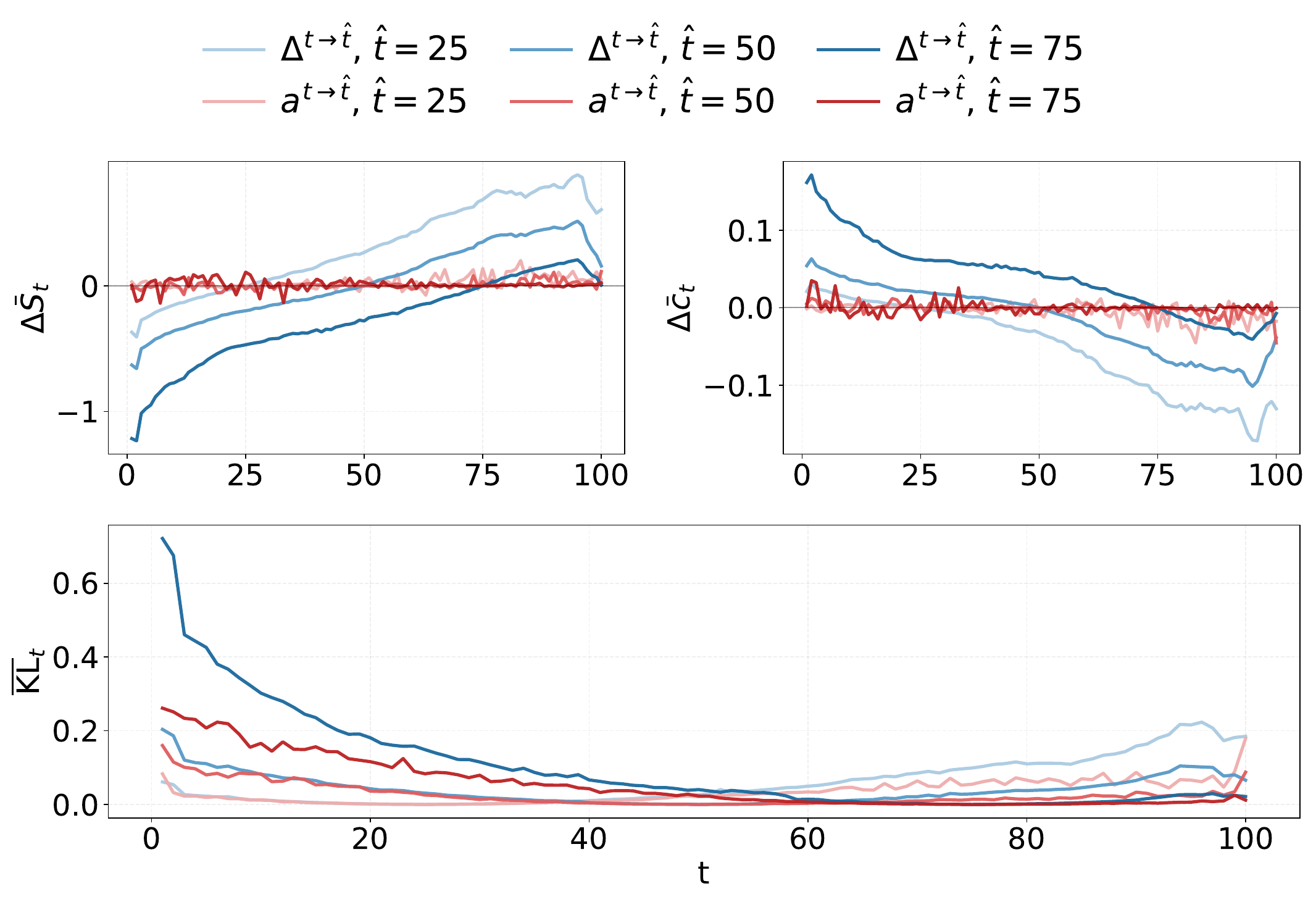}
    \end{subfigure}
    \caption{\textbf{Mean-steering downstream effects on layer 25 of \dream}. We steered the activations in layer 25 using the mean-ratio vectors (blue) targeting different $\tau$ values, and measured the variation in entropy, confidence and the KL divergence. We compared it against random perturbations (red). Steering with the mean vectors has an effect that is consistent with what would be expected.}
    \label{fig:mean_steering_dream_l25}
\end{figure}

\paragraph{Mean-vector steering on \dream.}
Having shown that $\tau$ is decodable across both models, we now turn from reading the signal to perturbing it, extending the steering analysis beyond the layers reported in the main text.
We begin with \dream: \Cref{fig:mean_steering_dream_l25} shows the downstream effects of steering at layer~25.
We observe a similar trend as in \llada, where entropy decreases and confidence increases when we steer towards later denoising steps.

\begin{figure*}[t]
    \centering

    \begin{subfigure}{0.45\textwidth}
        \caption{Layer $=3$}
        \includegraphics[width=\textwidth]{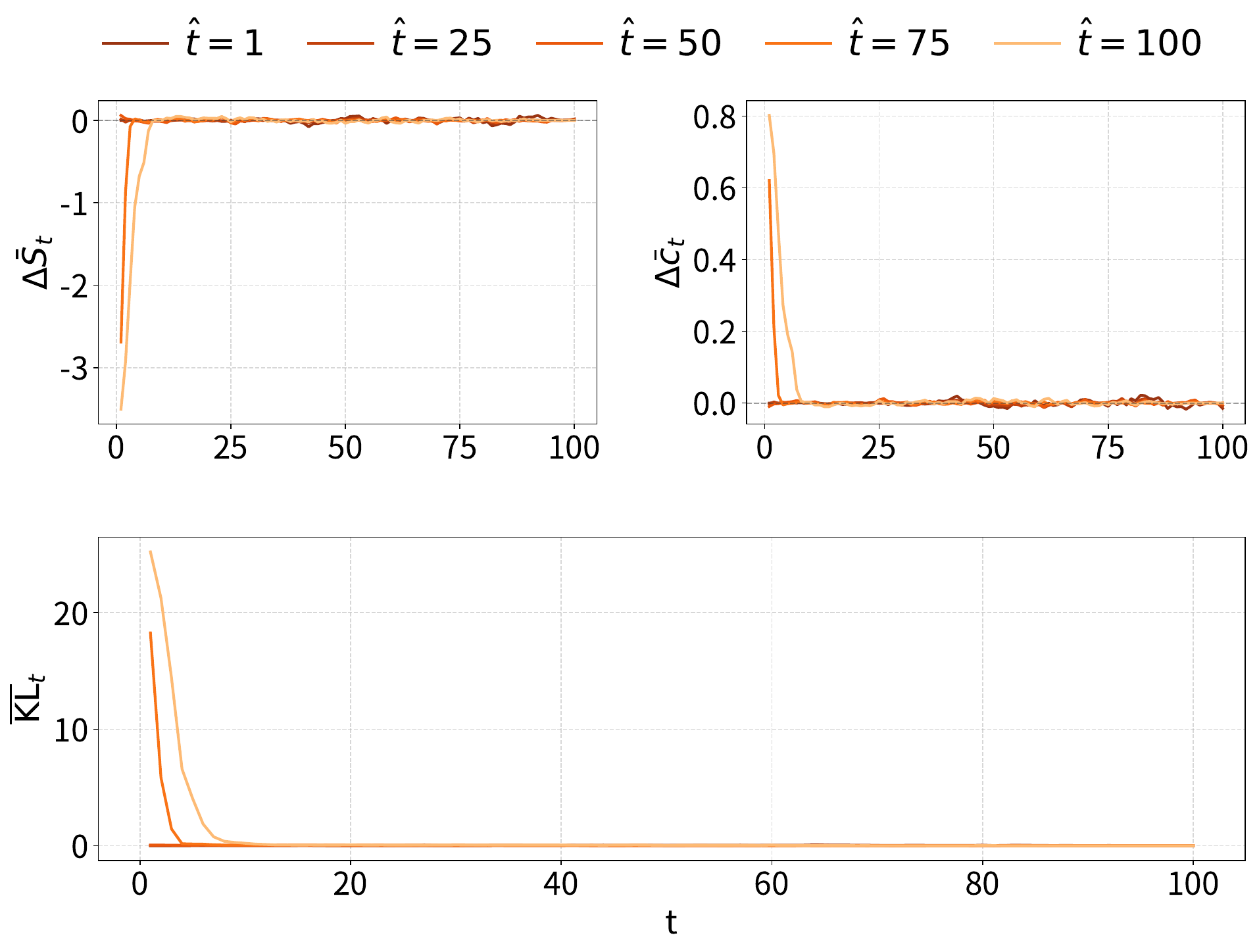}
    \end{subfigure}
    \begin{subfigure}{0.45\textwidth}
        \caption{Layer $=6$}
        \includegraphics[width=\textwidth]{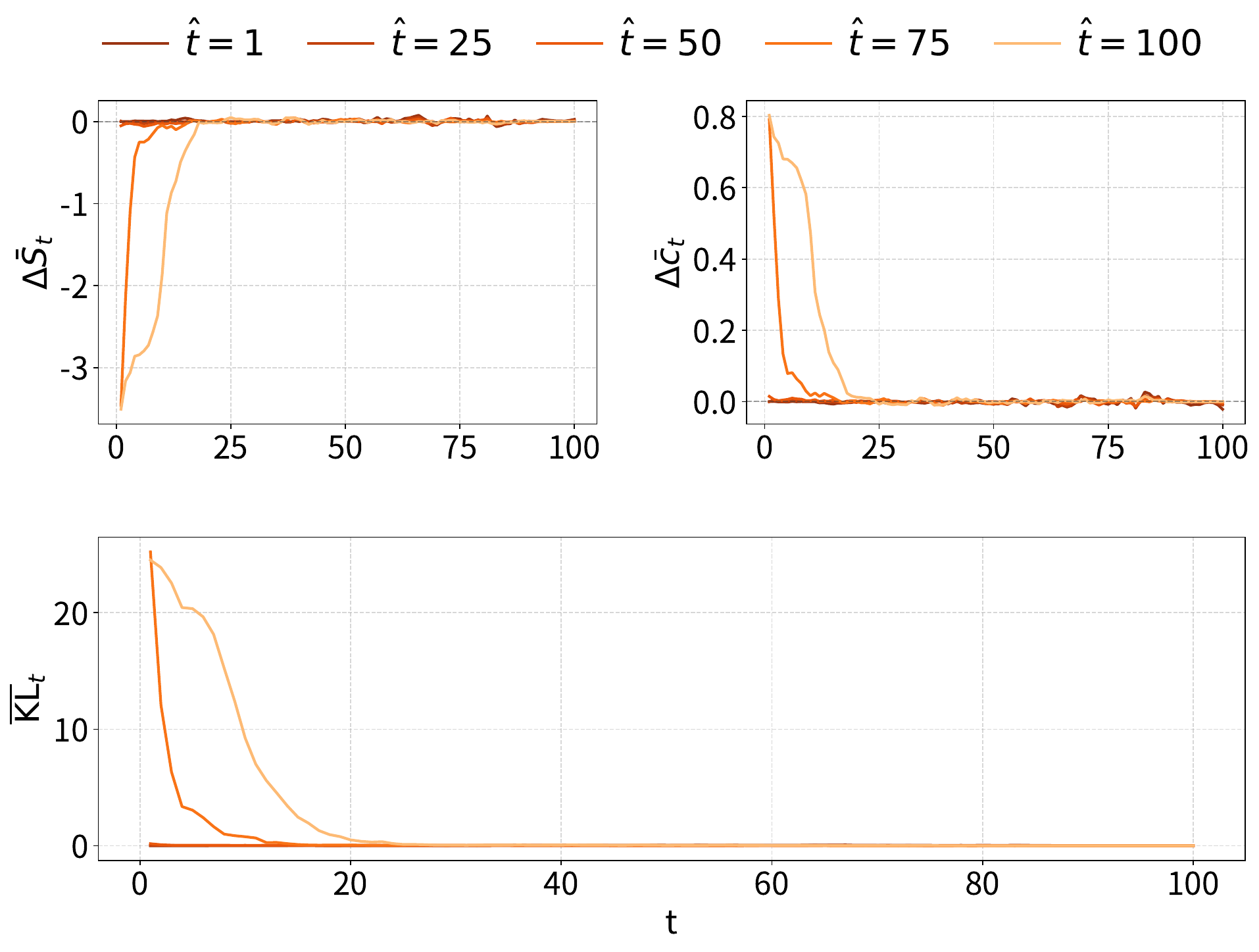}
    \end{subfigure}

    \vspace{0.3em}
    
    \begin{subfigure}{0.45\textwidth}
        \caption{Layer $=12$}
        \includegraphics[width=\textwidth]{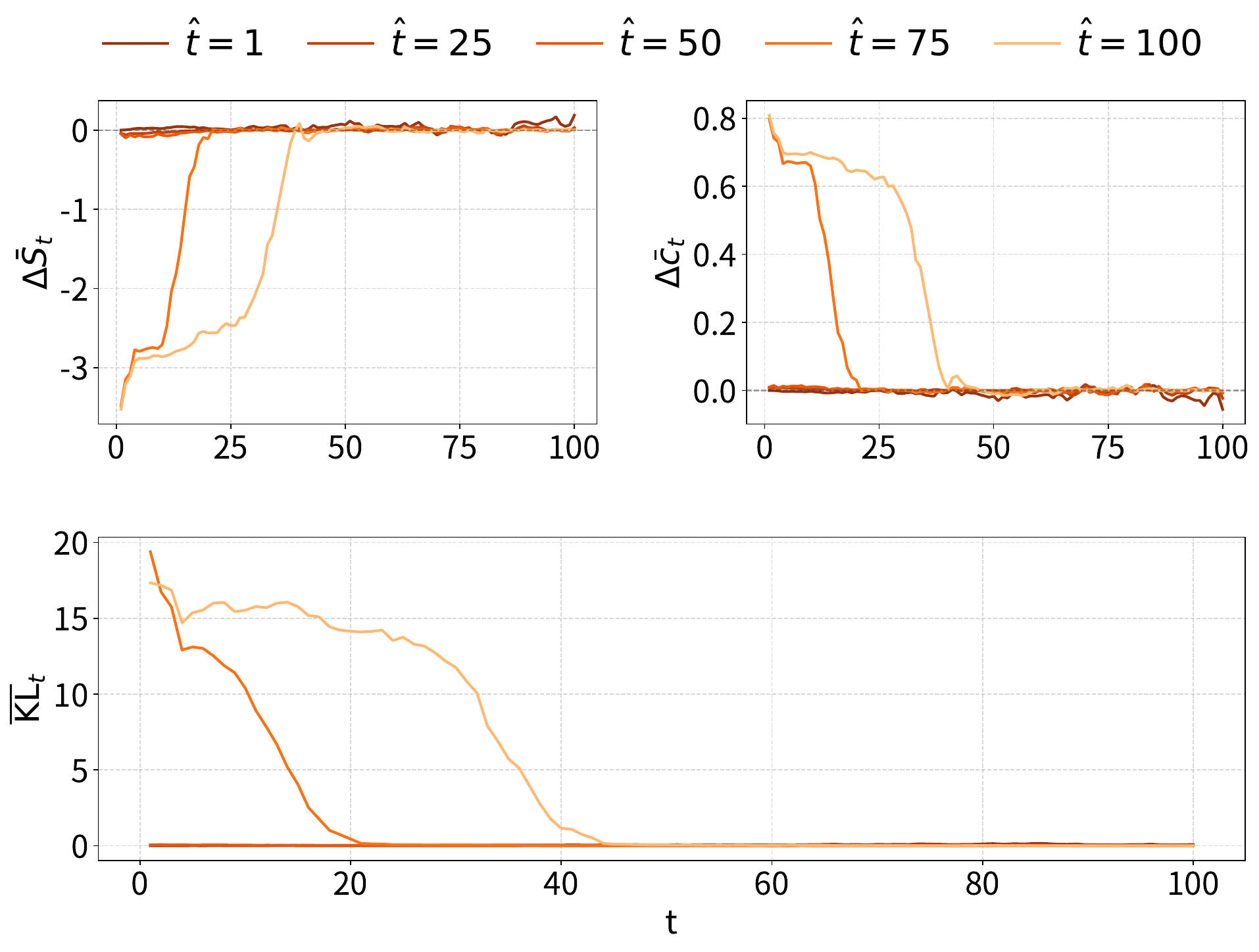}
    \end{subfigure}
    \begin{subfigure}{0.45\textwidth}
        \caption{Layer $=15$}
        \includegraphics[width=\textwidth]{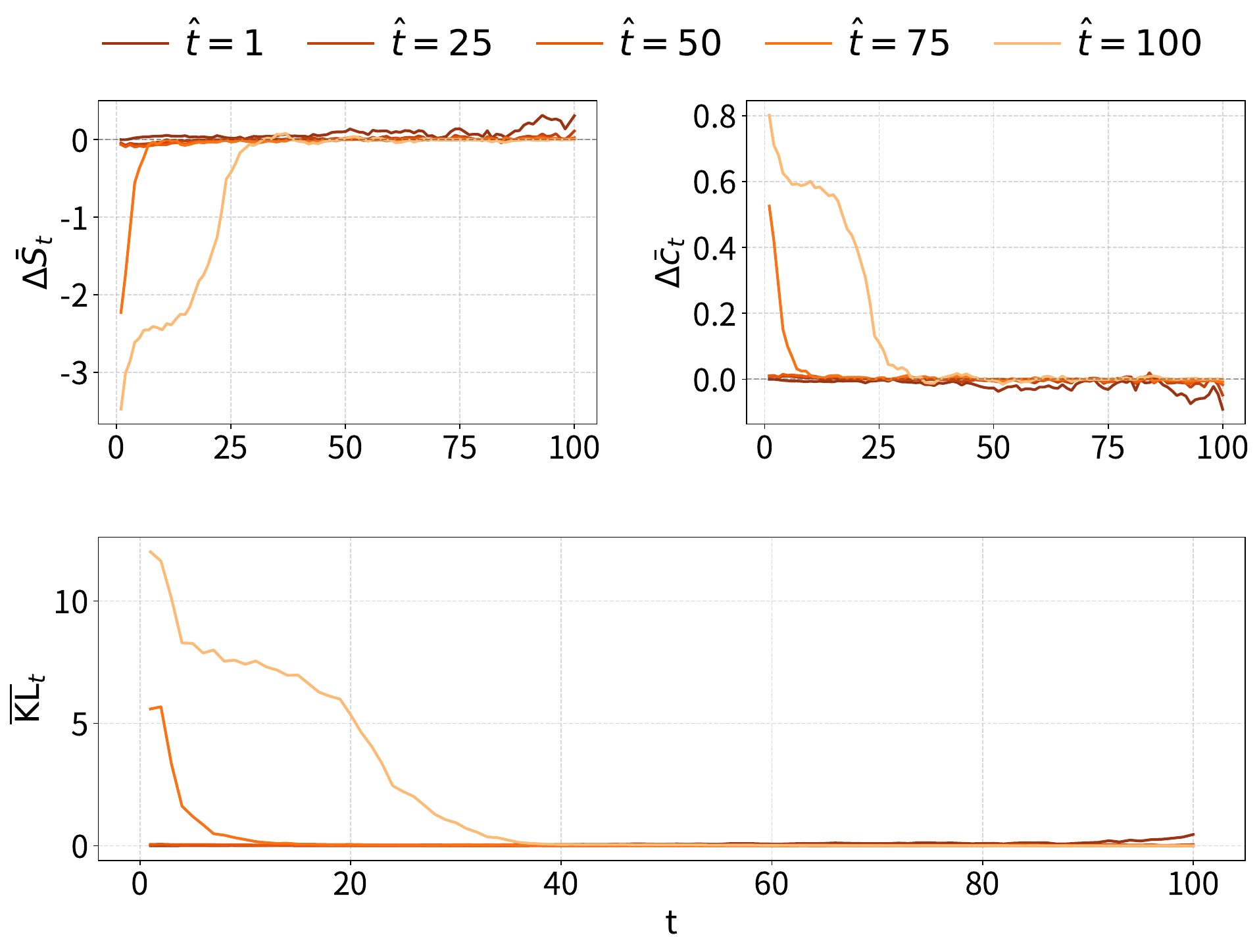}
    \end{subfigure}

    \vspace{0.3em}

    \begin{subfigure}{0.45\textwidth}
        \caption{Layer $=21$}
        \includegraphics[width=\textwidth]{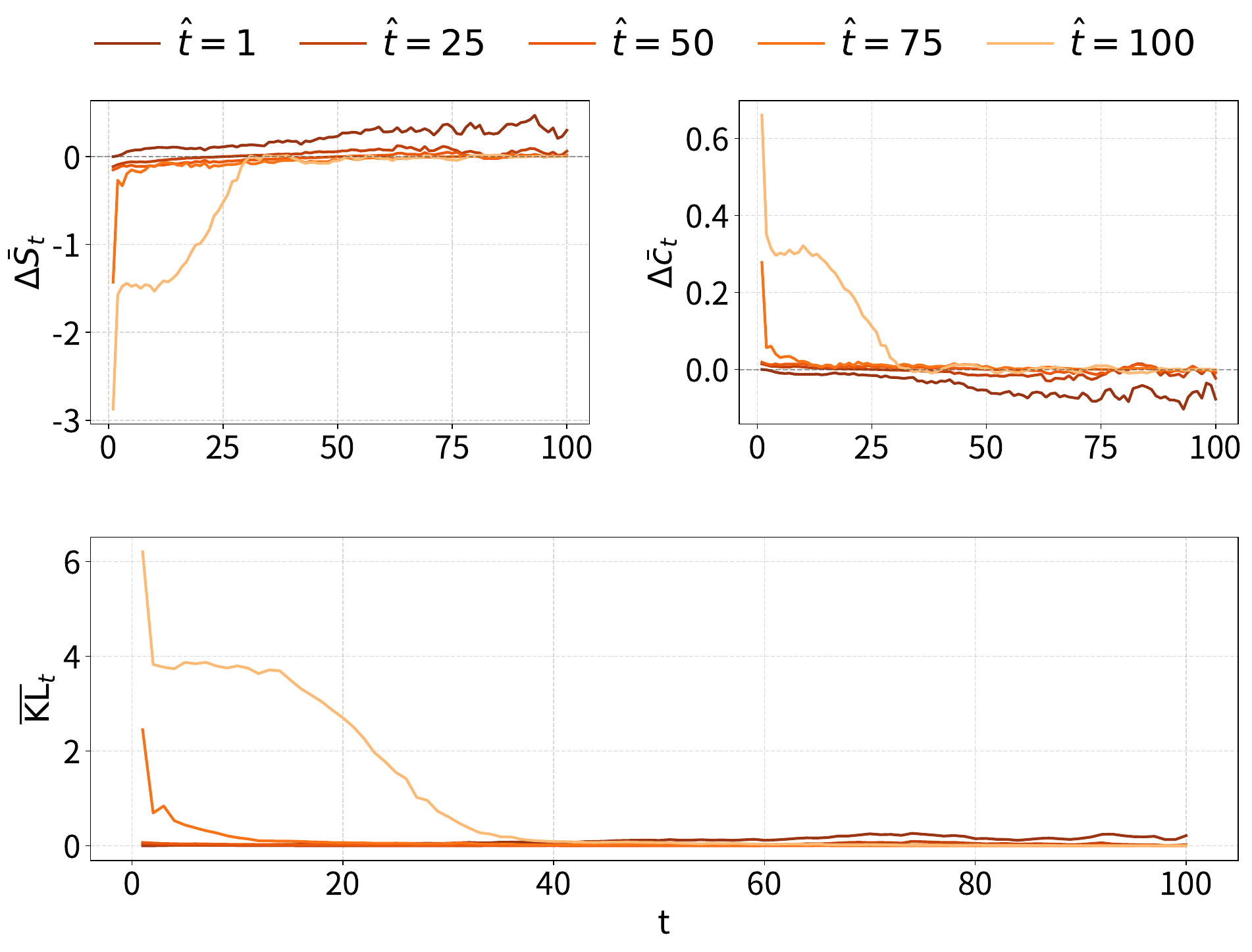}
    \end{subfigure}
    \begin{subfigure}{0.45\textwidth}
        \caption{Layer $=27$}
        \includegraphics[width=\textwidth]{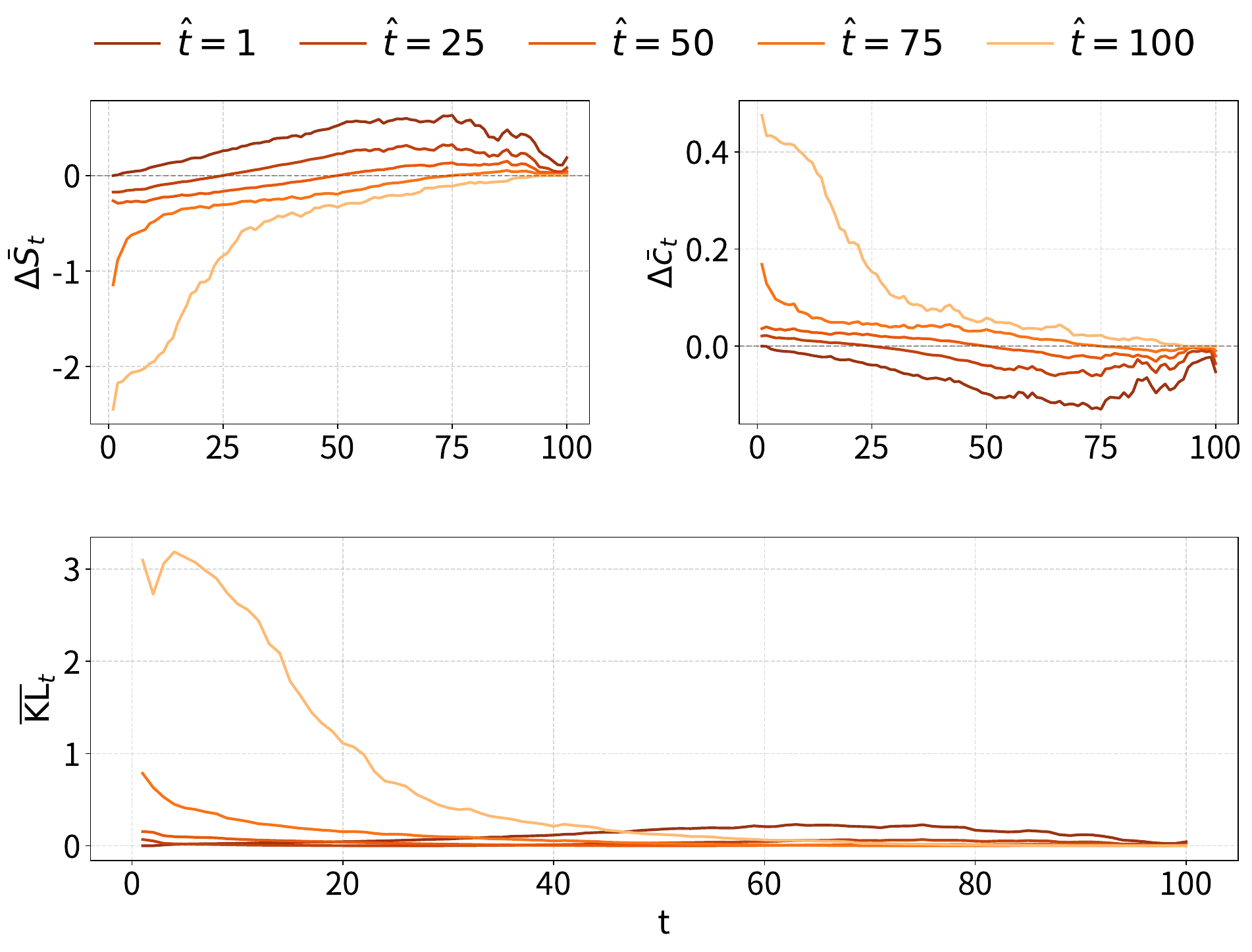}
    \end{subfigure}

    \caption{\textbf{Layer-wise mean-steering effects in LLaDA.} We apply mean vector steering at different intervention layers and target denoising-progress bins \(\hat{t} \in \{1,25,50,75,100\}\). Each panel reports the downstream effect on entropy drift, confidence drift and KL divergence across denoising steps.}
    \label{fig:mean_shift_ablation_llada}
\end{figure*}

\begin{figure*}[t]
    \centering

    \begin{subfigure}{0.45\textwidth}
        \caption{Layer $=3$}
        \includegraphics[width=\textwidth]{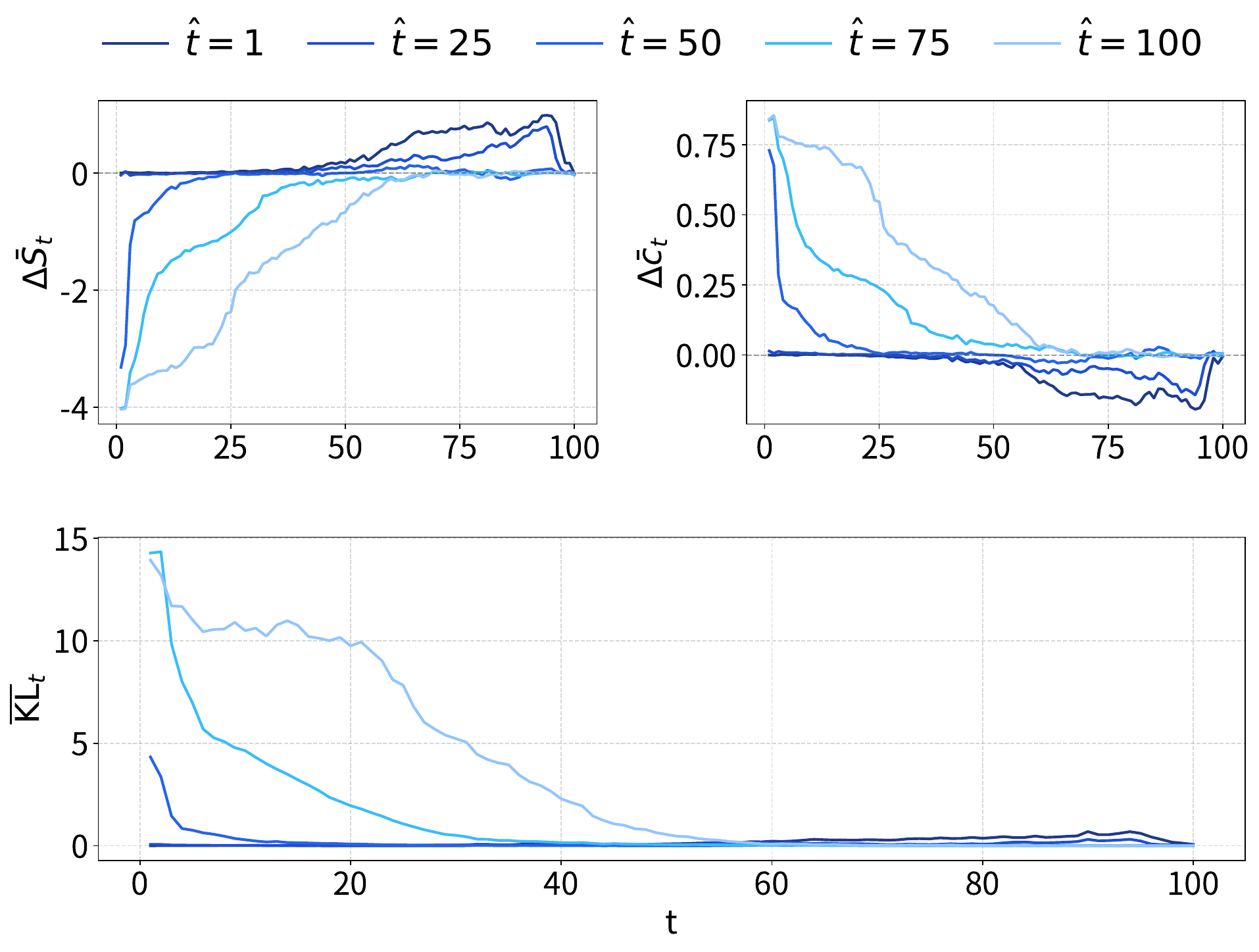}
    \end{subfigure}
    \begin{subfigure}{0.45\textwidth}
        \caption{Layer $=6$}
        \includegraphics[width=\textwidth]{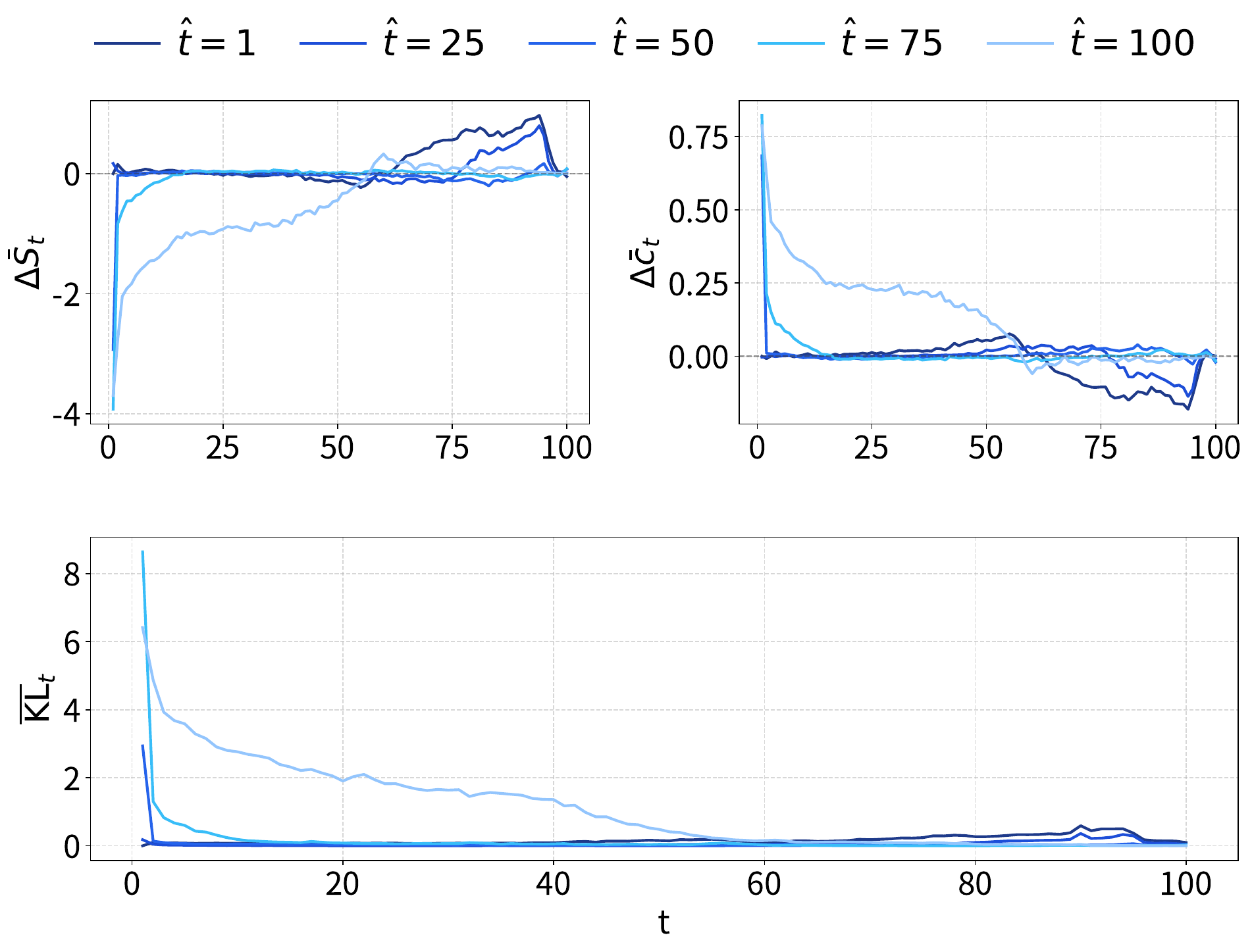}
    \end{subfigure}

    \vspace{0.3em}

    \begin{subfigure}{0.45\textwidth}
        \caption{Layer $=12$}
        \includegraphics[width=\textwidth]{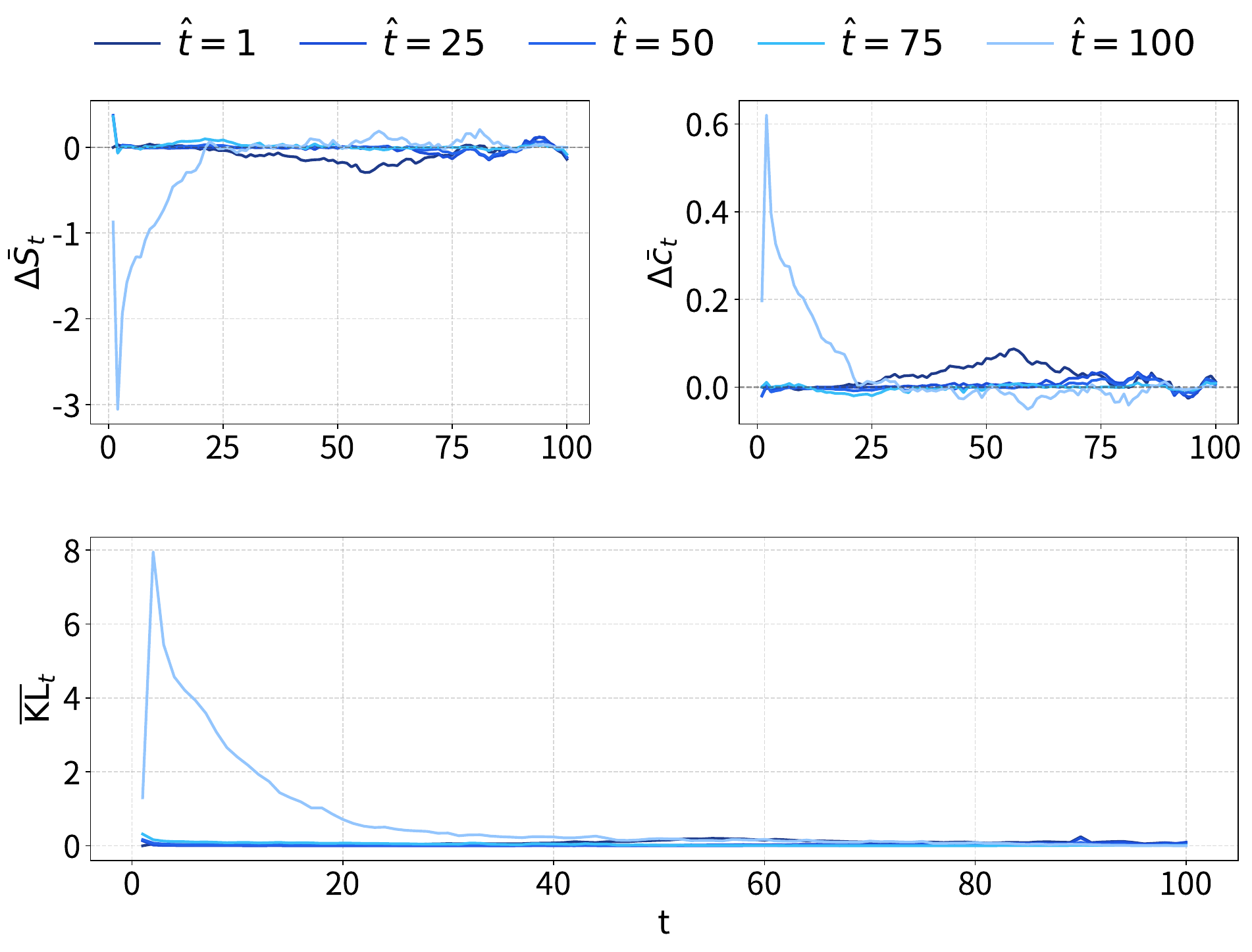}
    \end{subfigure}
    \begin{subfigure}{0.45\textwidth}
        \caption{Layer $=15$}
        \includegraphics[width=\textwidth]{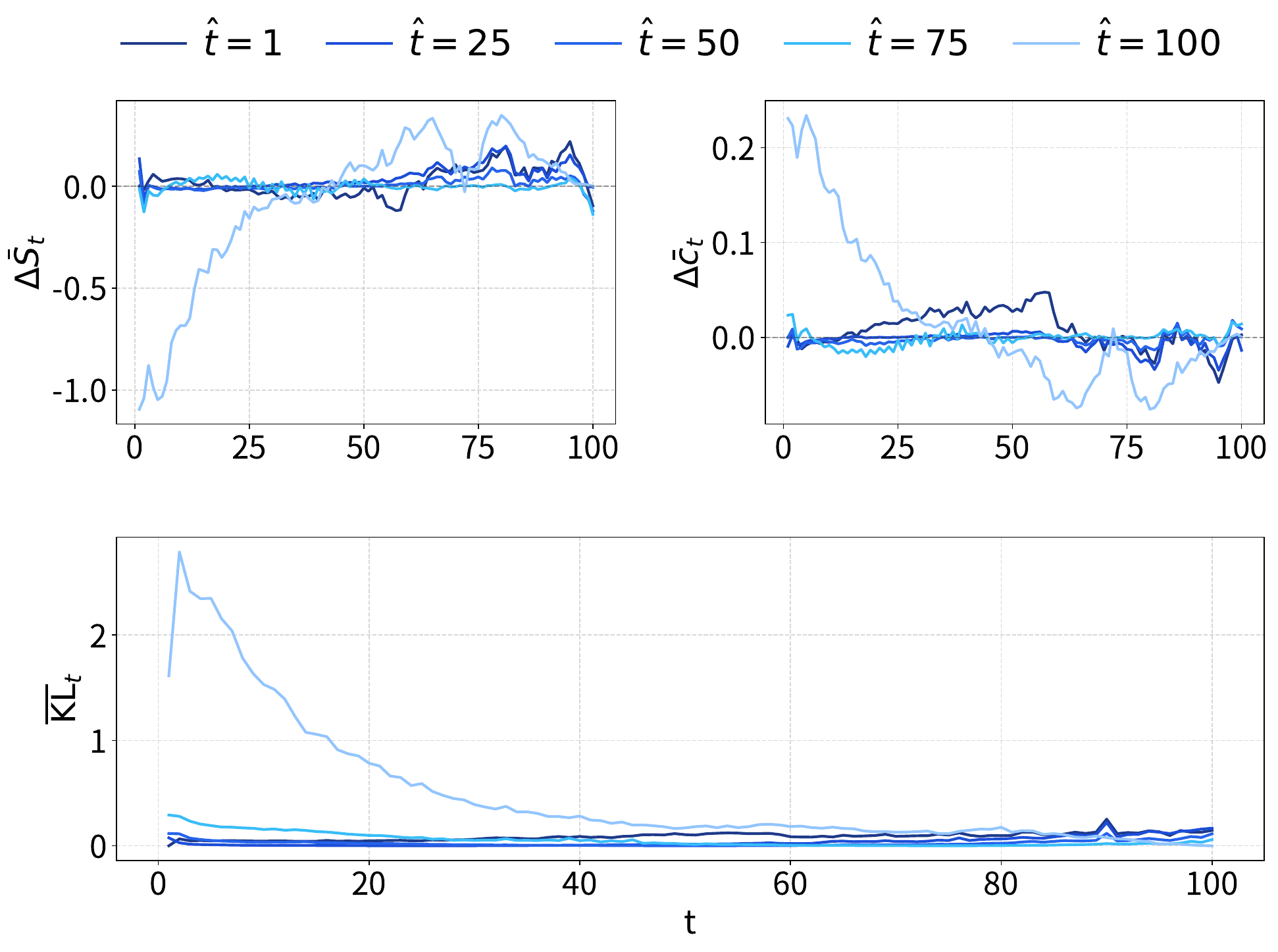}
    \end{subfigure}

    \vspace{0.3em}
    
    \begin{subfigure}{0.45\textwidth}
        \caption{Layer $=18$}
        \includegraphics[width=\textwidth]{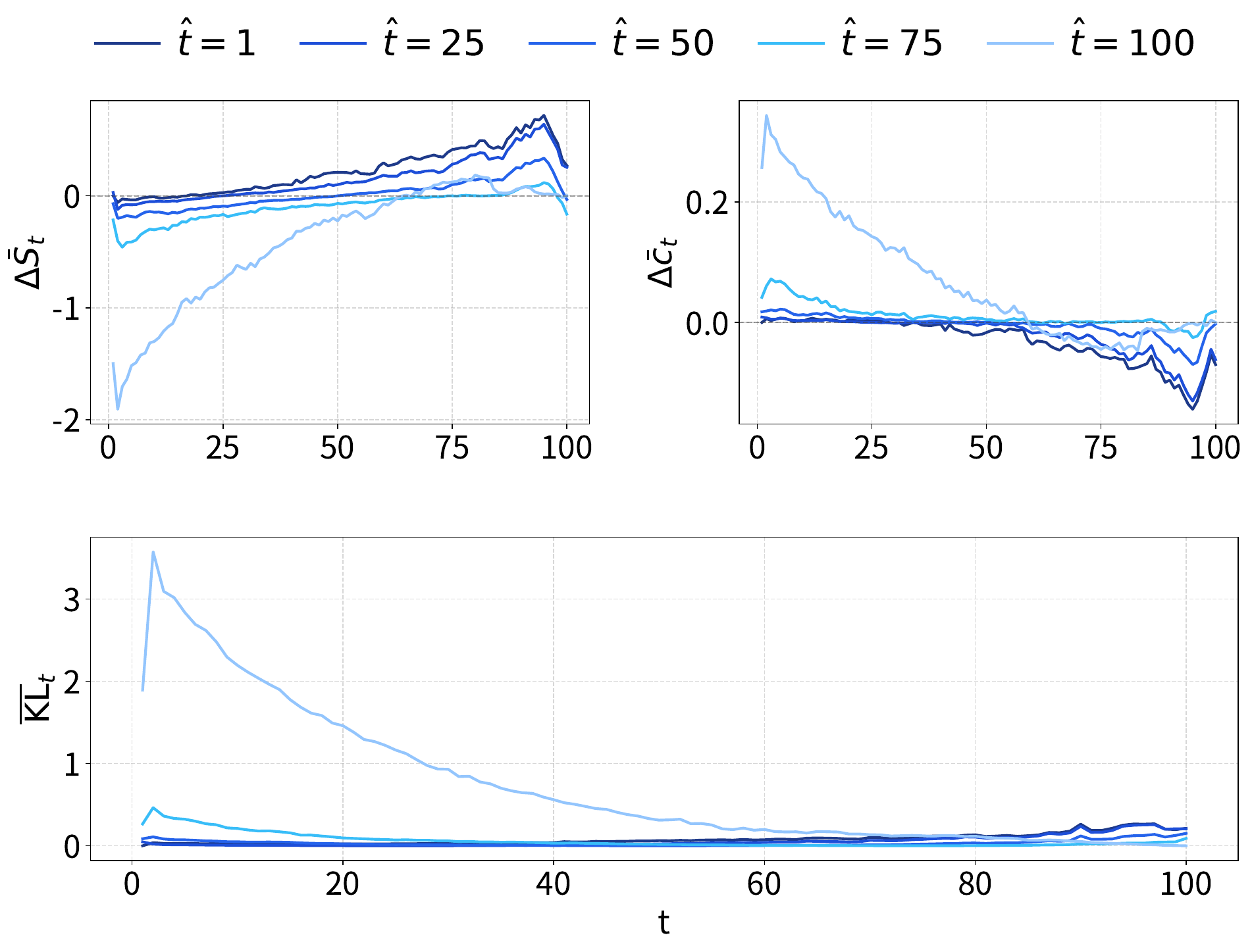}
    \end{subfigure}
    \begin{subfigure}{0.45\textwidth}
        \caption{Layer $=21$}
        \includegraphics[width=\textwidth]{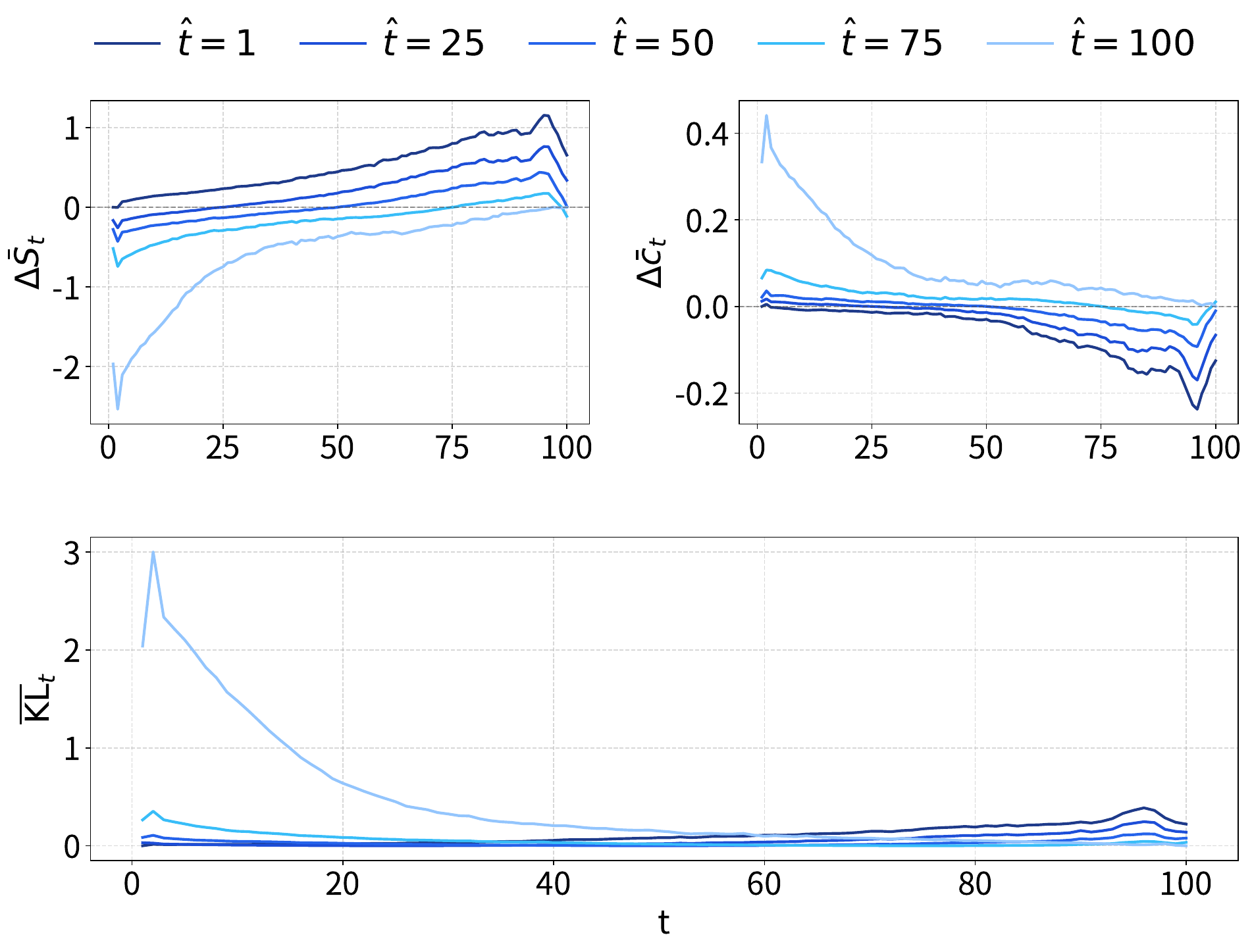}
    \end{subfigure}

    \caption{\textbf{Layer-wise mean-steering effects in Dream.} We apply mean vector steering at different intervention layers and target denoising-progress bins \(\hat{t} \in \{1,25,50,75,100\}\). Each panel reports the downstream effect on entropy drift, confidence drift and KL divergence across denoising steps.}
    \label{fig:mean_shift_ablation_dream}
\end{figure*}

\paragraph{Steering across intervention layers.}
\Cref{fig:mean_shift_ablation_llada} and \Cref{fig:mean_shift_ablation_dream} show the downstream results for \llada and \dream across a range of intervention layers and target bins $\hat{t}$.

\begin{figure*}
    \centering
    \begin{subfigure}{\textwidth}
        \centering
        \includegraphics[width=\textwidth]{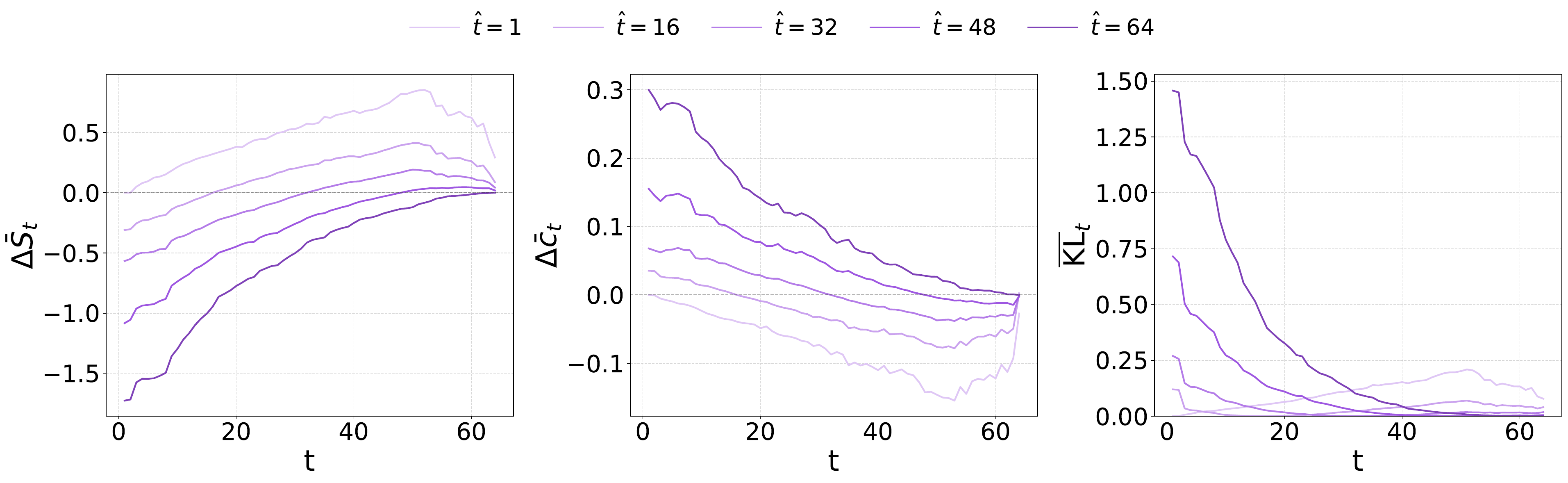}
        \caption{$L=64$,\quad $\hat{t}\in\{1,16,32,48,64\}$.}
        \label{fig:mean_steering_g64}
    \end{subfigure}
    \medskip
    \begin{subfigure}{\textwidth}
        \centering
        \includegraphics[width=\textwidth]{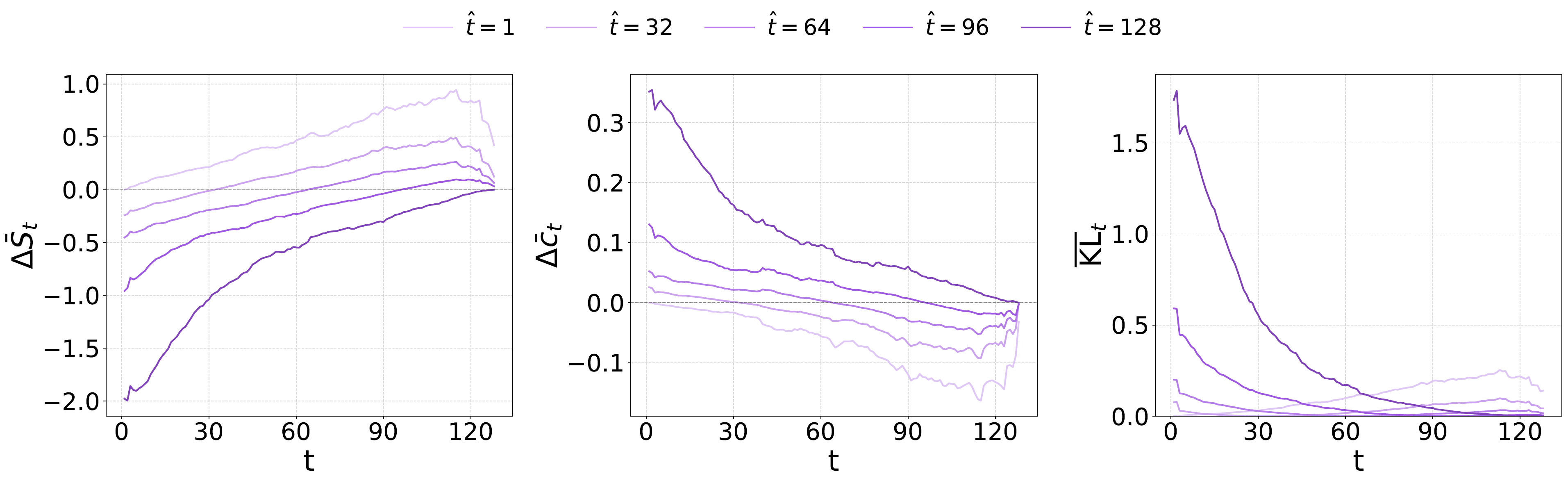}
        \caption{$L=128$,\quad $\hat{t}\in\{1,32,64,96,128\}$.}
        \label{fig:mean_steering_g128}
    \end{subfigure}
\caption{\textbf{Mean-steering downstream effects across generation lengths for \llada}. Ablation over generation length $L \in \{64, 128\}$, with the number of denoising steps matched to $L$. Each row shows, left to right, $\Delta\bar{S}_{t}$, $\Delta\bar{c}_{t}$ and $\overline{\mathrm{KL}}_{t}$ versus the denoising step $t$, for five steering targets $\hat{t}$. Vertical axes use a symmetric-log scale to resolve the convergence near 0.}
    \label{fig:mean_steering_genlen}
\end{figure*}

\paragraph{Robustness to generation length.}
In \Cref{fig:mean_steering_genlen} we ablate the effect of generation length, repeating the mean-steering analysis at $L=64$ and $L=128$ (with the number of denoising steps matched to $L$).
The downstream effects follow the same trajectory as in the main results across all steering targets $\hat{t}$, indicating that the previously-observed phenomena are not artifacts of a particular generation length.

\paragraph{Downstream task performance.}
In \Cref{tab:meanratio-sweep} we show the effect of mean vector steering on downstream task performance.
We observe that steering does not lead to a collapse in performance across the range of $\hat{t}$ values, with \llada remaining within a few points of its base score on all three benchmarks. \dream is more sensitive, particularly on GSM8K, where performance degrades substantially at $\hat{t}=100$.
A qualitative inspection of \dream's incorrect GSM8K generations reveals that this degradation is largely driven by the premature emission of end-of-sequence (EOS) tokens. Steering towards $\hat{t}=100$ leads the model to unmask EOS tokens both early in the denoising trajectory and at early positions within the response window. Since these models are trained to keep predicting EOS once the first one has been emitted, a single EOS placed near the start of the response causes the generation to collapse into a degenerate, EOS-filled sequence.
This failure mode is consistent with the decoding biases reported by \citet{huang2025pc}, who show that uncertainty-based samplers in DLMs over-select trivial, high-confidence tokens---including EOS, newlines, and punctuation---during the first denoising steps, and tend to commit the two ends of the sequence before its centre (a ``U-shaped'' decoding trajectory). 
Our intervention can be read as amplifying this bias from the representation side: by driving the model's internal estimate of denoising progress to its maximum, we further inflate the probability of completion tokens such as EOS, which in turn triggers the collapse described above. 
Taken together, and while we leave a quantitative analysis to future work, this suggests that the $\tau$ signal is not a content-agnostic progress counter, but is instead entangled with the token distribution itself---carrying latent information about which tokens become more or less probable as denoising advances, with sequence-ending tokens such as EOS being a clear example.

\begin{table}[t]
\centering
\caption{Effect of mean-steering on \llada (layer 29) and \dream (layer 25) on downstream performance.}
\label{tab:meanratio-sweep}
\setlength{\tabcolsep}{5pt} 
\renewcommand{\arraystretch}{1.12} 
\normalsize 
\begin{tabular}{llcccccc}
\toprule
 & & & \multicolumn{5}{c}{$\hat{t}$} \\
\cmidrule(lr){4-8}
Model & Dataset & base & $1$ & $25$ & $50$ & $75$ & $100$ \\
\midrule
\multirow{3}{*}{LLaDA-1.5} & GSM8K & 84.3 & 81.6 & 81.0 & 83.1 & 84.1 & 81.0 \\
 & HumanEval & 44.5 & 41.6 & 47.4 & 46.0 & 45.3 & 46.7 \\
 & StrategyQA & 69.5 & 63.1 & 69.5 & 66.6 & 65.3 & 69.2 \\
\midrule
\multirow{3}{*}{Dream-7B} & GSM8K & 68.8 & 51.6 & 61.5 & 59.5 & 56.9 & 23.1 \\
 & HumanEval & 35.0 & 28.0 & 32.9 & 35.0 & 29.4 & 23.8 \\
 & StrategyQA & 68.3 & 67.2 & 66.3 & 67.2 & 68.5 & 68.3 \\
\bottomrule
\end{tabular}
\end{table}

\section{Low-dimensional Steering}
\label{sec:appendixsubspacesteering}
\begin{figure*}[t]
    \centering

    \begin{subfigure}[t]{0.45\textwidth}
        \centering
        \caption{Layer $=6$, $k=1$}
        \includegraphics[width=\textwidth]{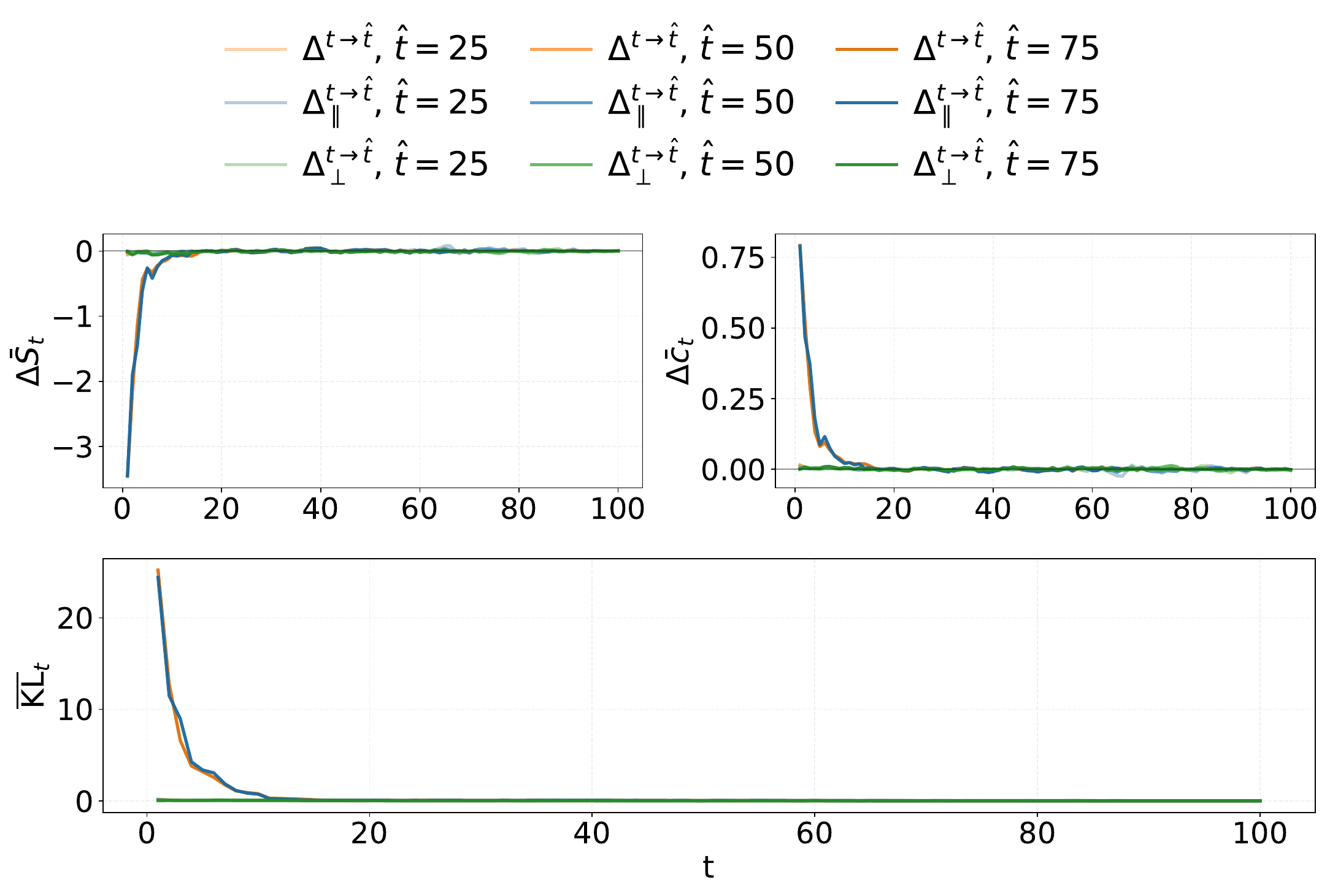}
    \end{subfigure}
    \begin{subfigure}[t]{0.45\textwidth}
        \centering
        \caption{Layer $=6$, $k=10$}
        \includegraphics[width=\textwidth]{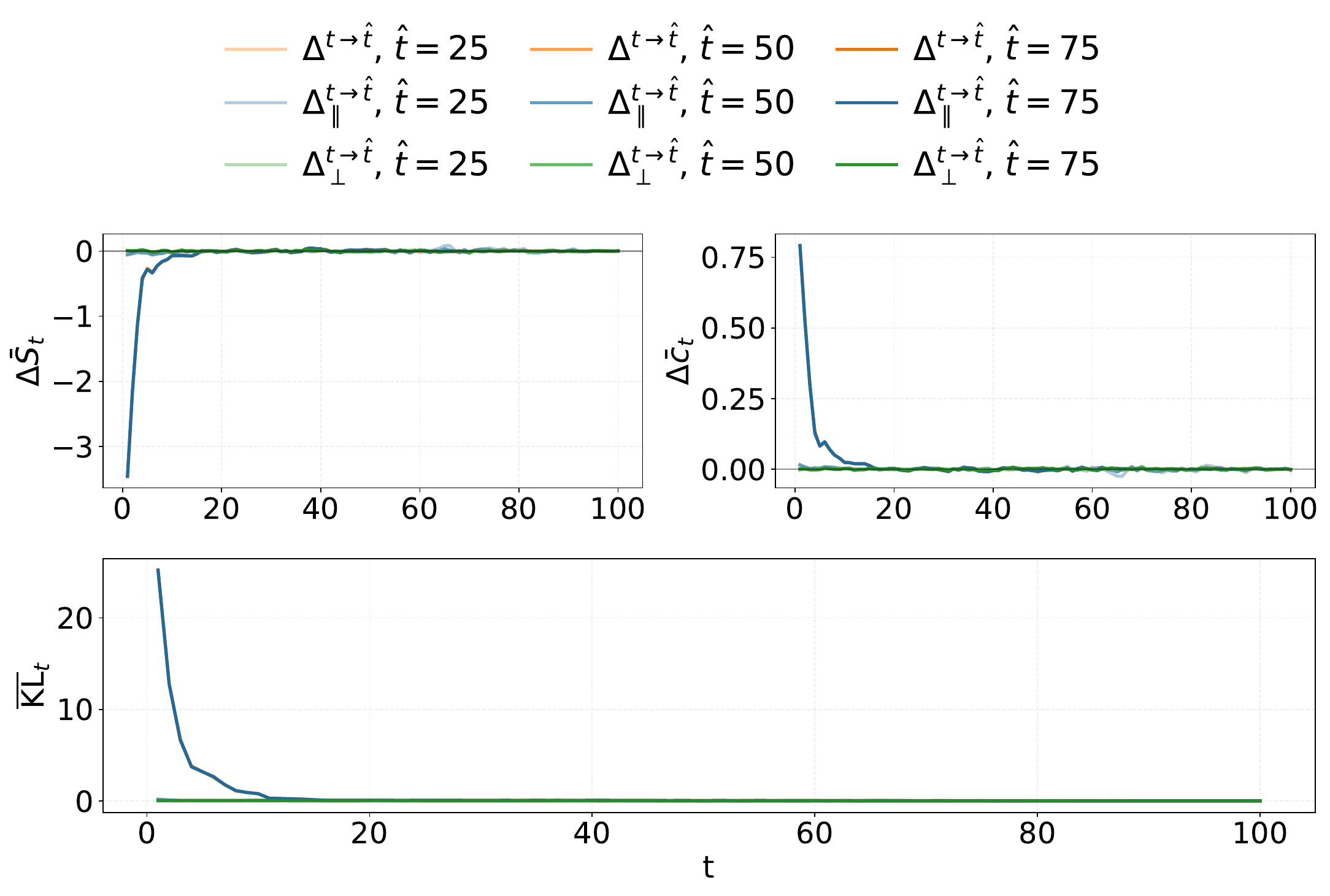}
    \end{subfigure}

    \begin{subfigure}[t]{0.45\textwidth}
        \centering
        \caption{Layer $=15$, $k=1$}
        \includegraphics[width=\textwidth]{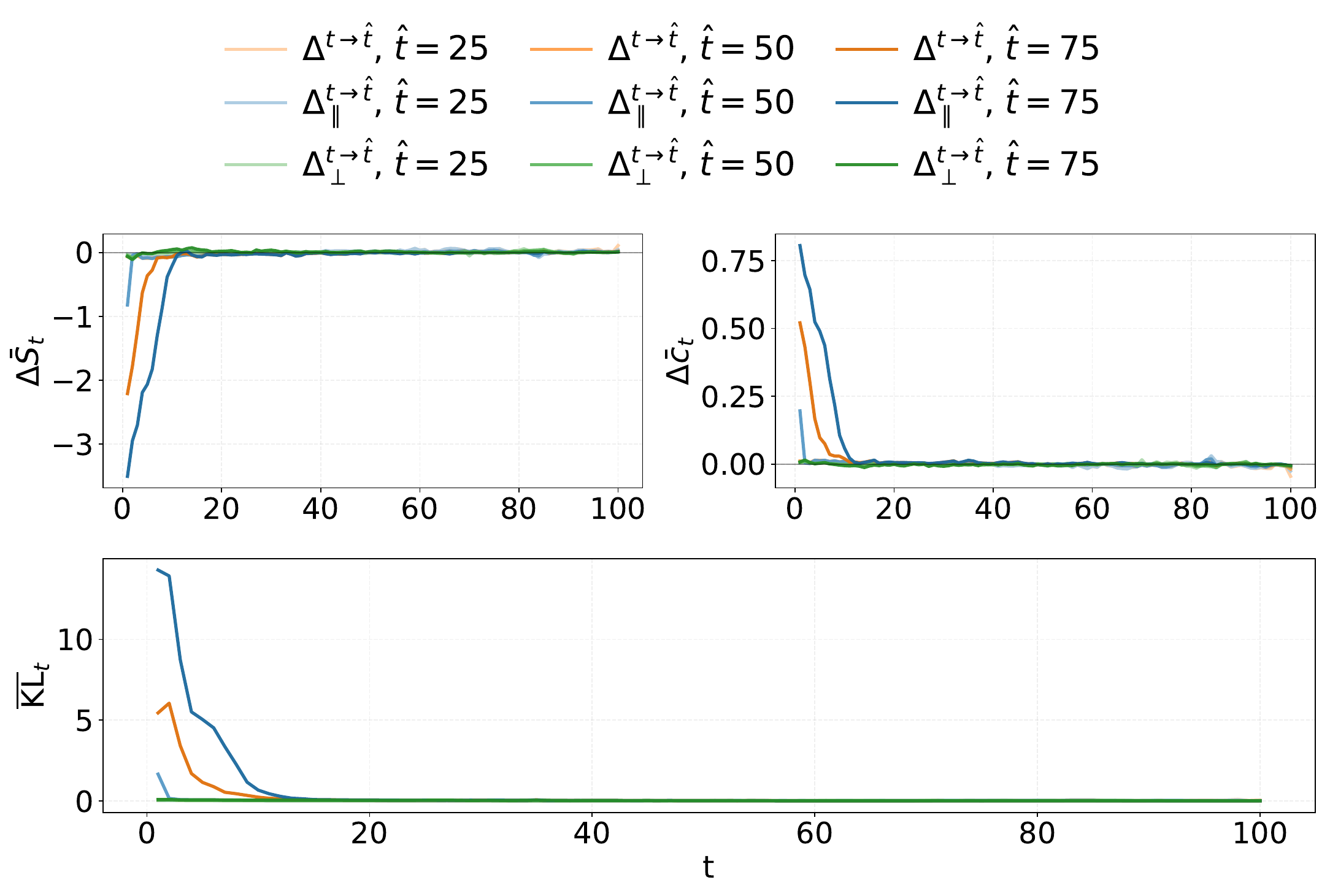}
    \end{subfigure}
    \begin{subfigure}[t]{0.45\textwidth}
        \centering
        \caption{Layer $=15$, $k=2$}
        \includegraphics[width=\textwidth]{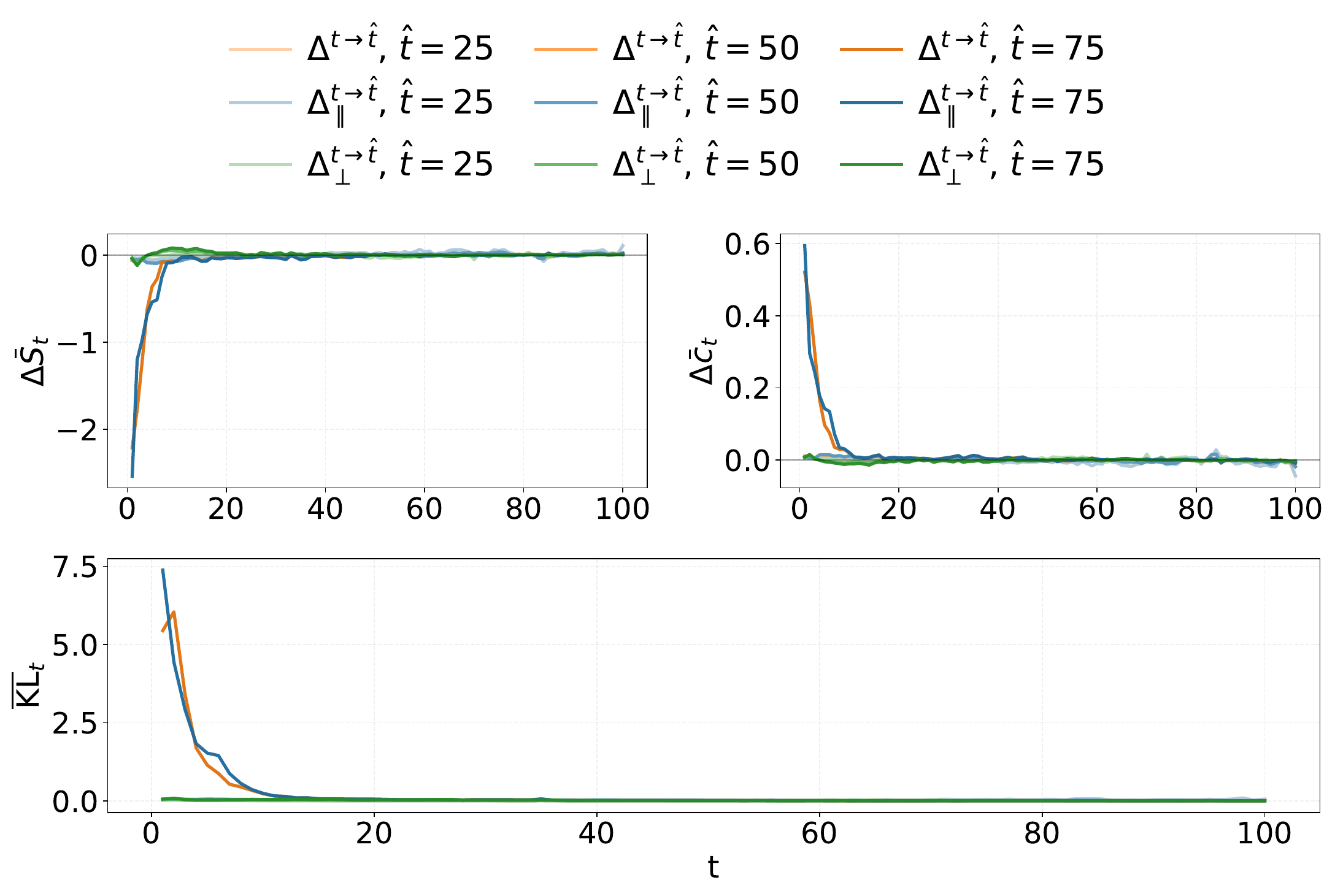}
    \end{subfigure}

    \vspace{0.3em}

    \begin{subfigure}[t]{0.45\textwidth}
        \centering
        \caption{Layer $=15$, $k=10$}
        \includegraphics[width=\textwidth]{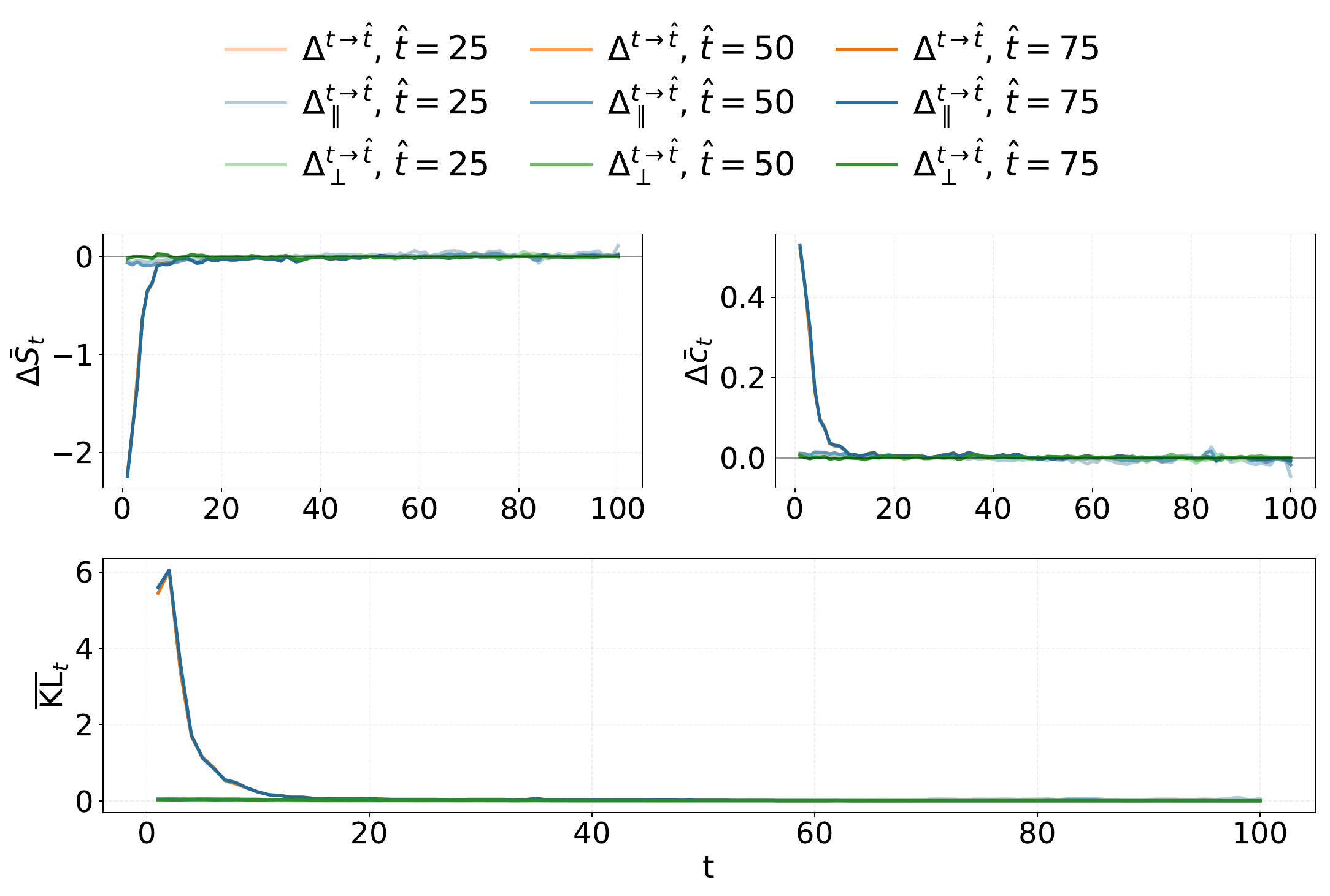}
    \end{subfigure}
    \begin{subfigure}[t]{0.45\textwidth}
        \centering
        \caption{Layer $=29$, $k=1$}
        \includegraphics[width=\textwidth]{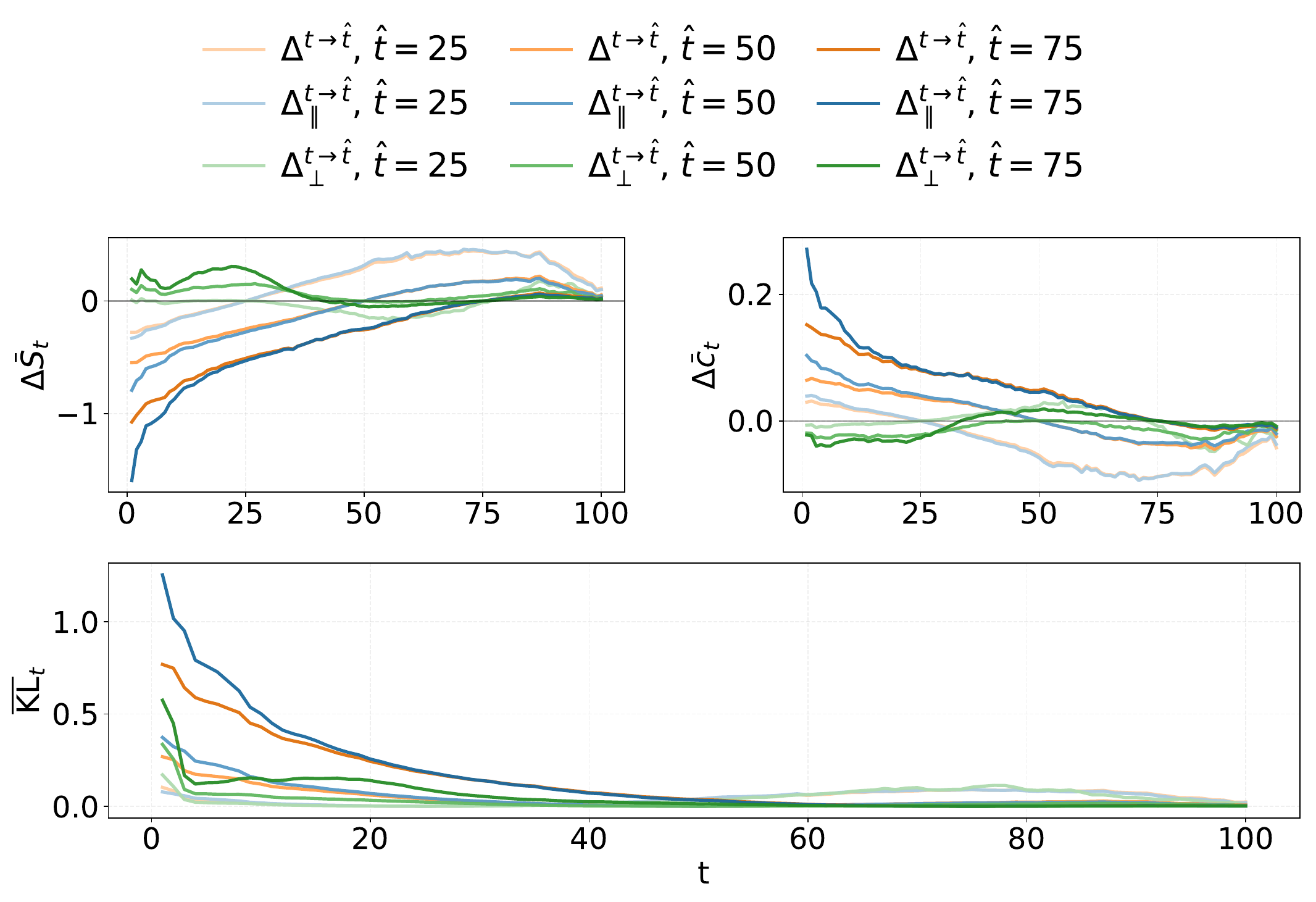}
    \end{subfigure}

    \vspace{0.3em}
    
    \begin{subfigure}[t]{0.45\textwidth}
        \centering
        \caption{Layer $=29$, $k=2$}
        \includegraphics[width=\textwidth]{assets/subspace_steering/llada_subspace_steering_layer29_k2.pdf}
    \end{subfigure}
    \begin{subfigure}[t]{0.45\textwidth}
        \centering
        \caption{Layer $=29$, $k=10$}
        \includegraphics[width=\textwidth]{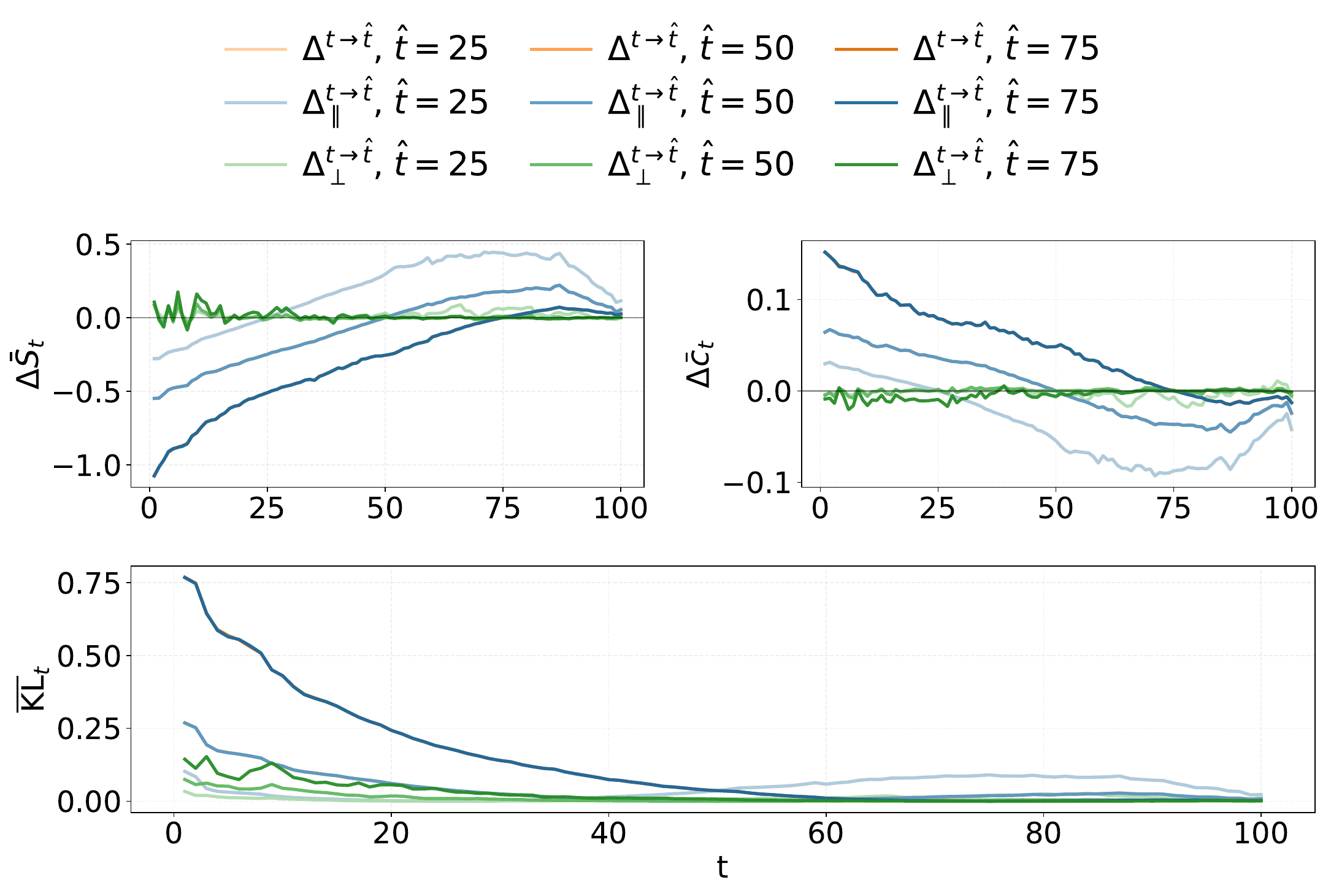}
    \end{subfigure}

    \caption{\textbf{Low-dimensional steering on \llada.} \llada's mean vectors concentrate around a low-dimensional subspace. Steering across the top-1, top-2, and top-10 principal components yields results similar to using the unprojected mean vector.}
    \label{fig:llada_subspace_steering}
\end{figure*}

\begin{figure*}[t]
    \centering

    \begin{subfigure}[t]{0.45\textwidth}
        \centering
        \caption{Layer $=6$, $k=1$}
        \includegraphics[width=\textwidth]{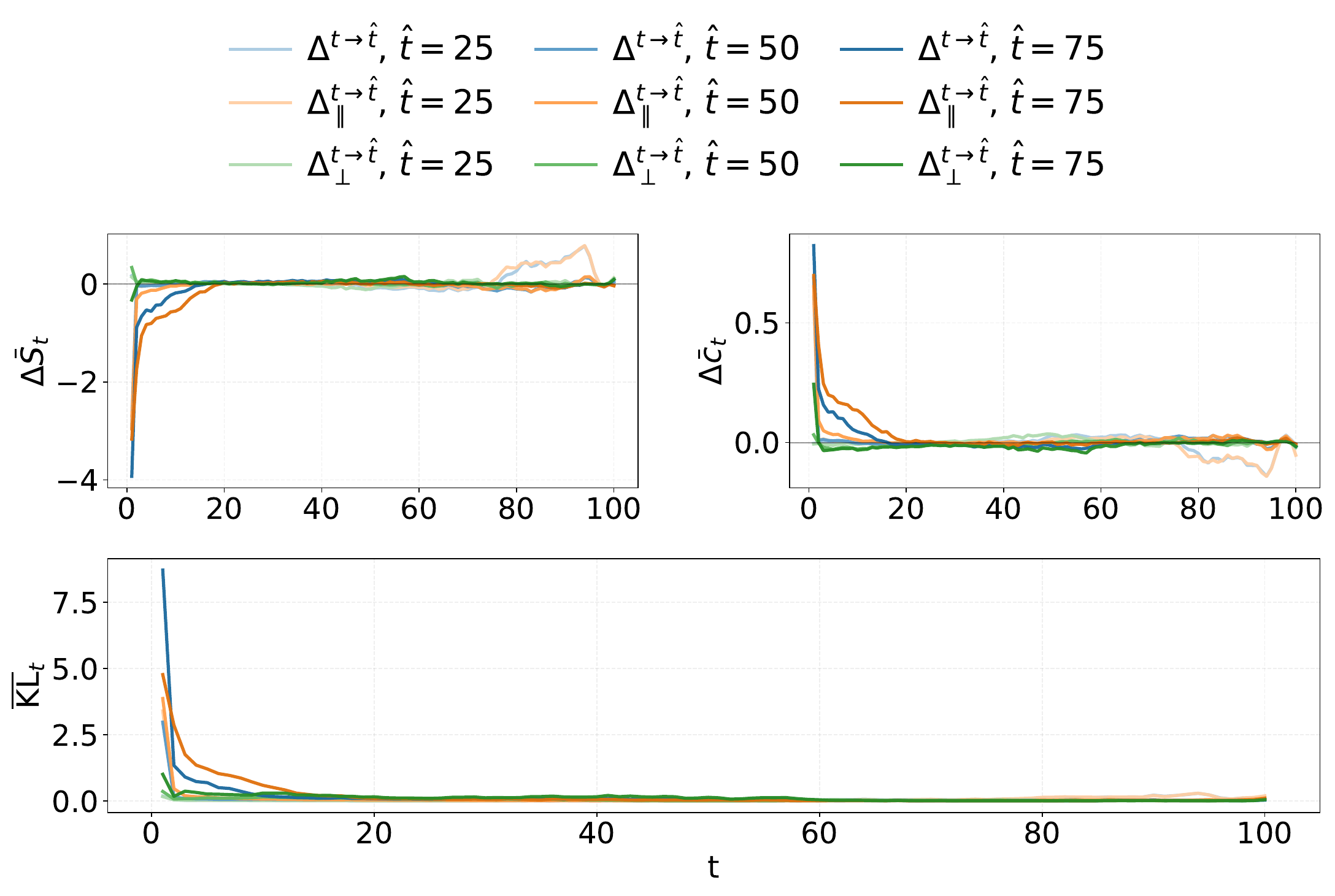}
    \end{subfigure}
    \begin{subfigure}[t]{0.45\textwidth}
        \centering
        \caption{Layer $=6$, $k=10$}
        \includegraphics[width=\textwidth]{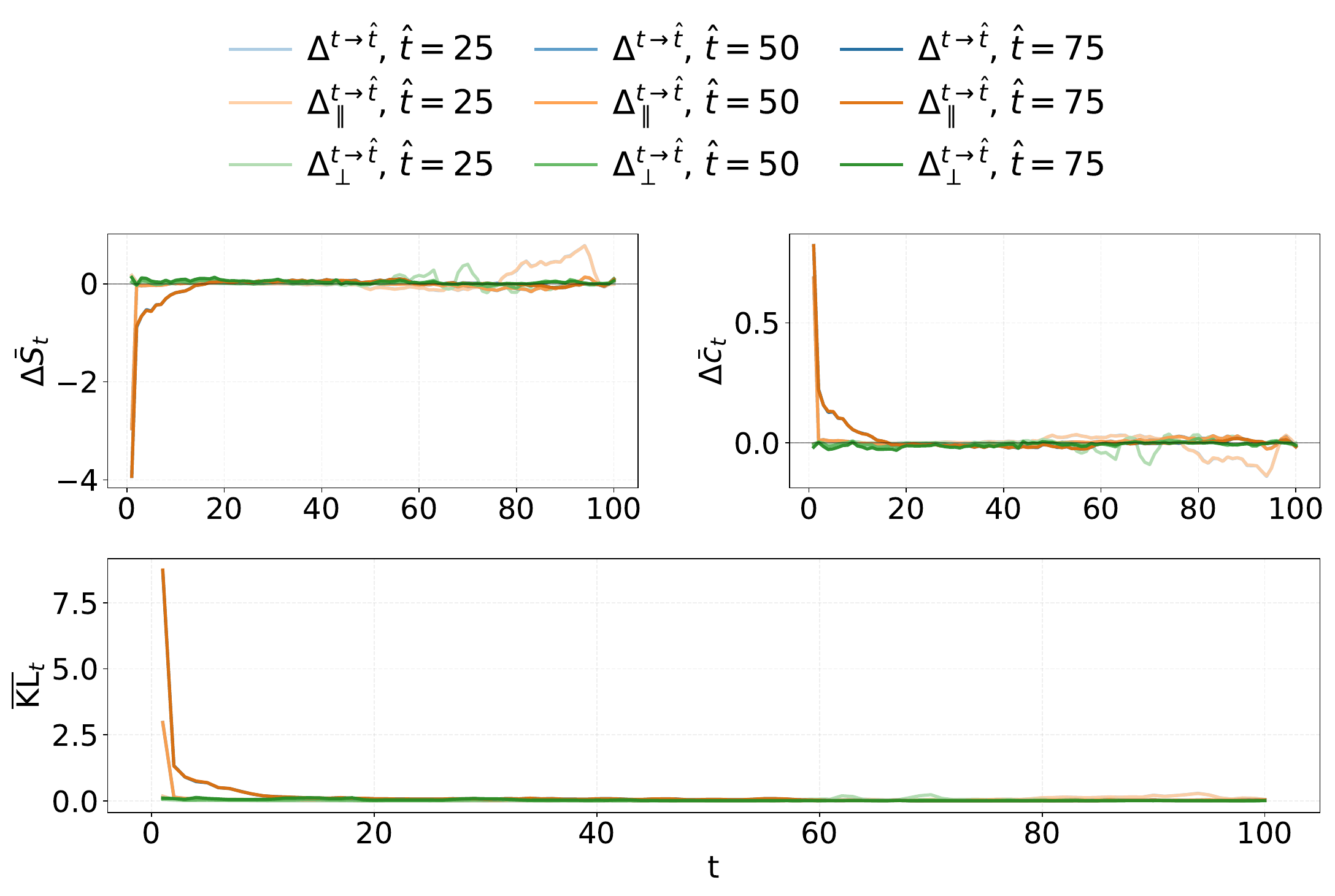}
    \end{subfigure}

    \begin{subfigure}[t]{0.45\textwidth}
        \centering
        \caption{Layer $=15$, $k=1$}
        \includegraphics[width=\textwidth]{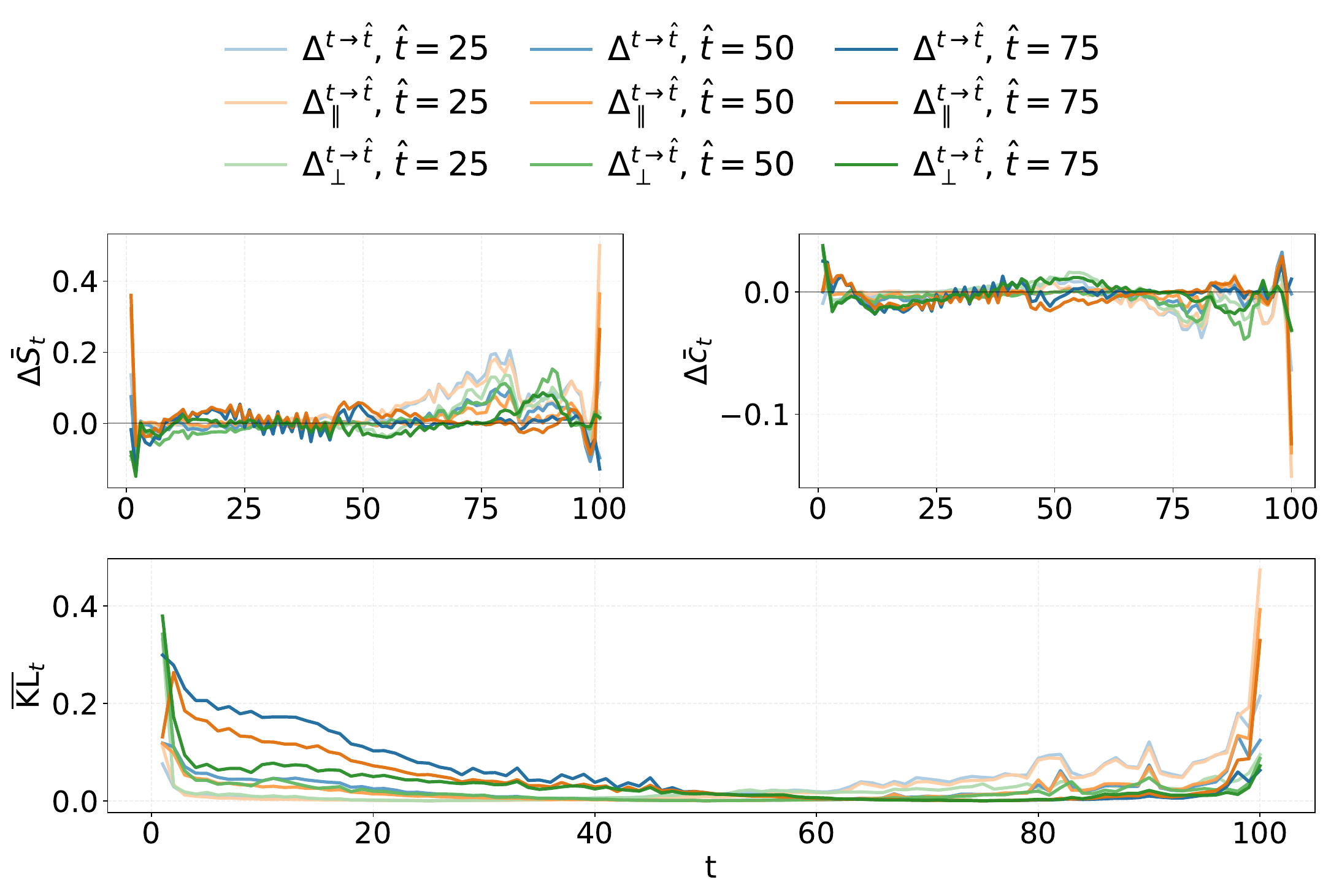}
    \end{subfigure}
    \begin{subfigure}[t]{0.45\textwidth}
        \centering
        \caption{Layer $=15$, $k=2$}
        \includegraphics[width=\textwidth]{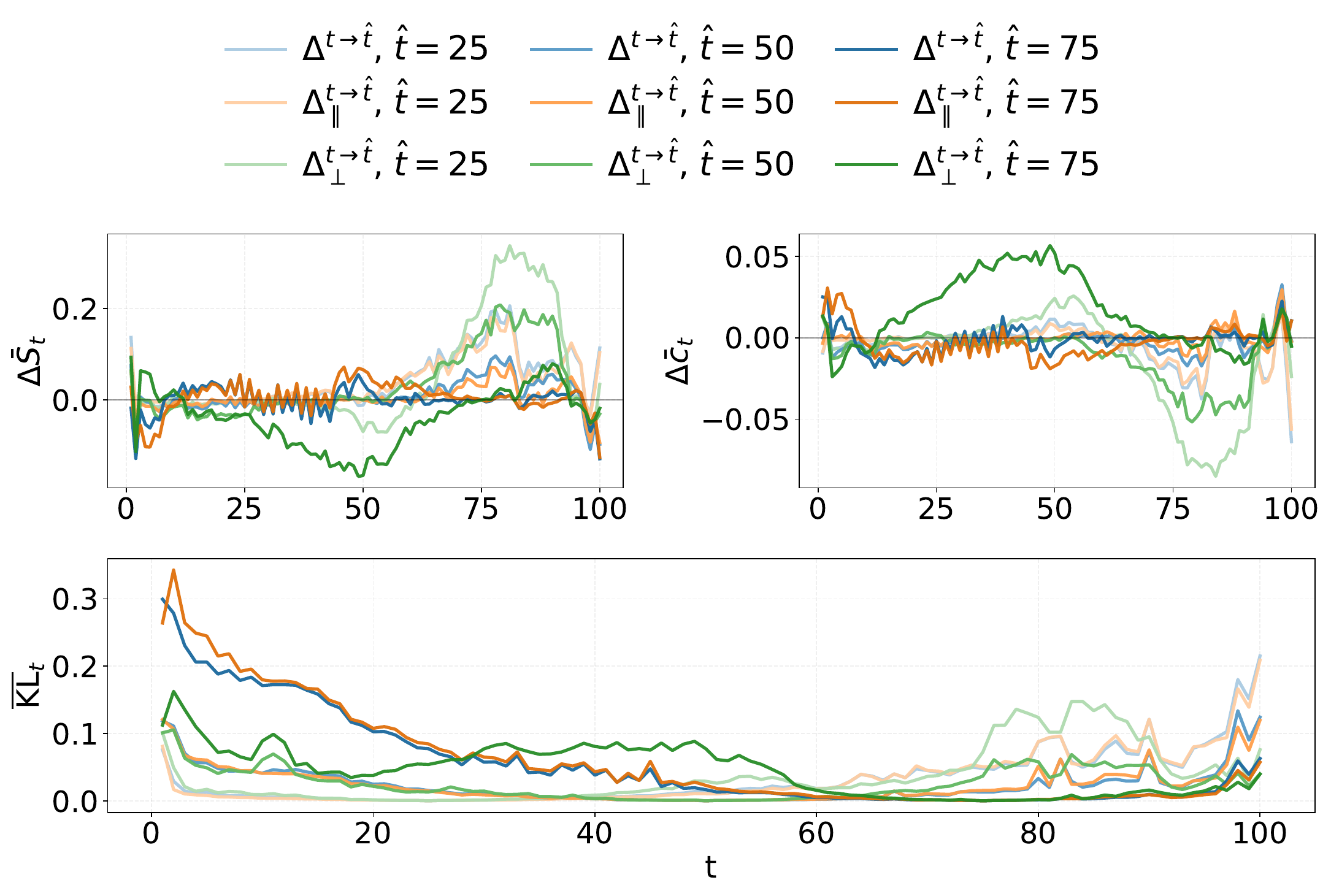}
    \end{subfigure}

    \vspace{0.3em}

    \begin{subfigure}[t]{0.45\textwidth}
        \centering
        \caption{Layer $=25$, $k=1$}
        \includegraphics[width=\textwidth]{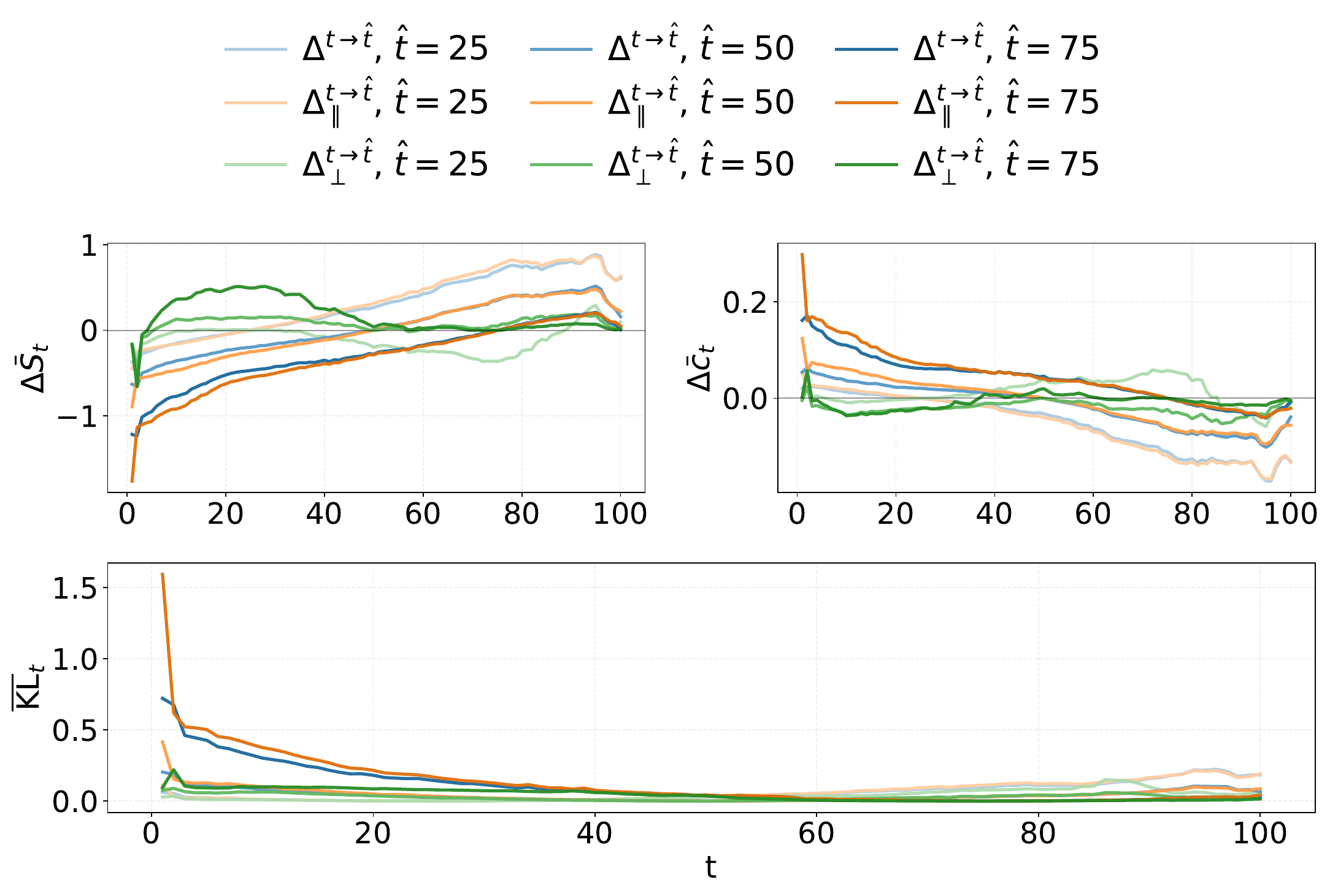}
    \end{subfigure}
    \begin{subfigure}[t]{0.45\textwidth}
        \centering
        \caption{Layer $=25$, $k=2$}
        \includegraphics[width=\textwidth]{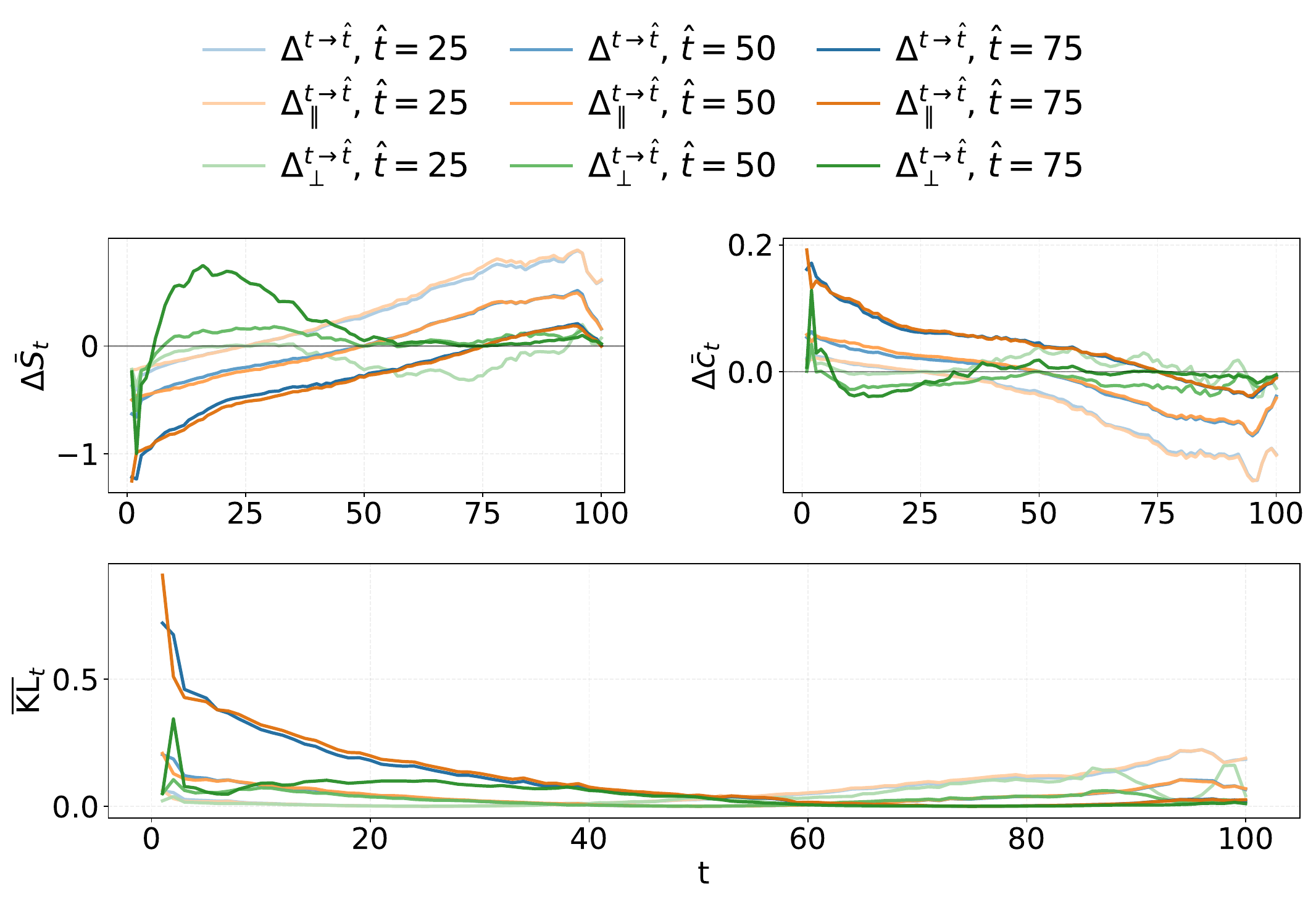}
    \end{subfigure}

    \vspace{0.3em}
    
    \begin{subfigure}[t]{0.45\textwidth}
        \centering
        \caption{Layer $=25$, $k=4$}
        \includegraphics[width=\textwidth]{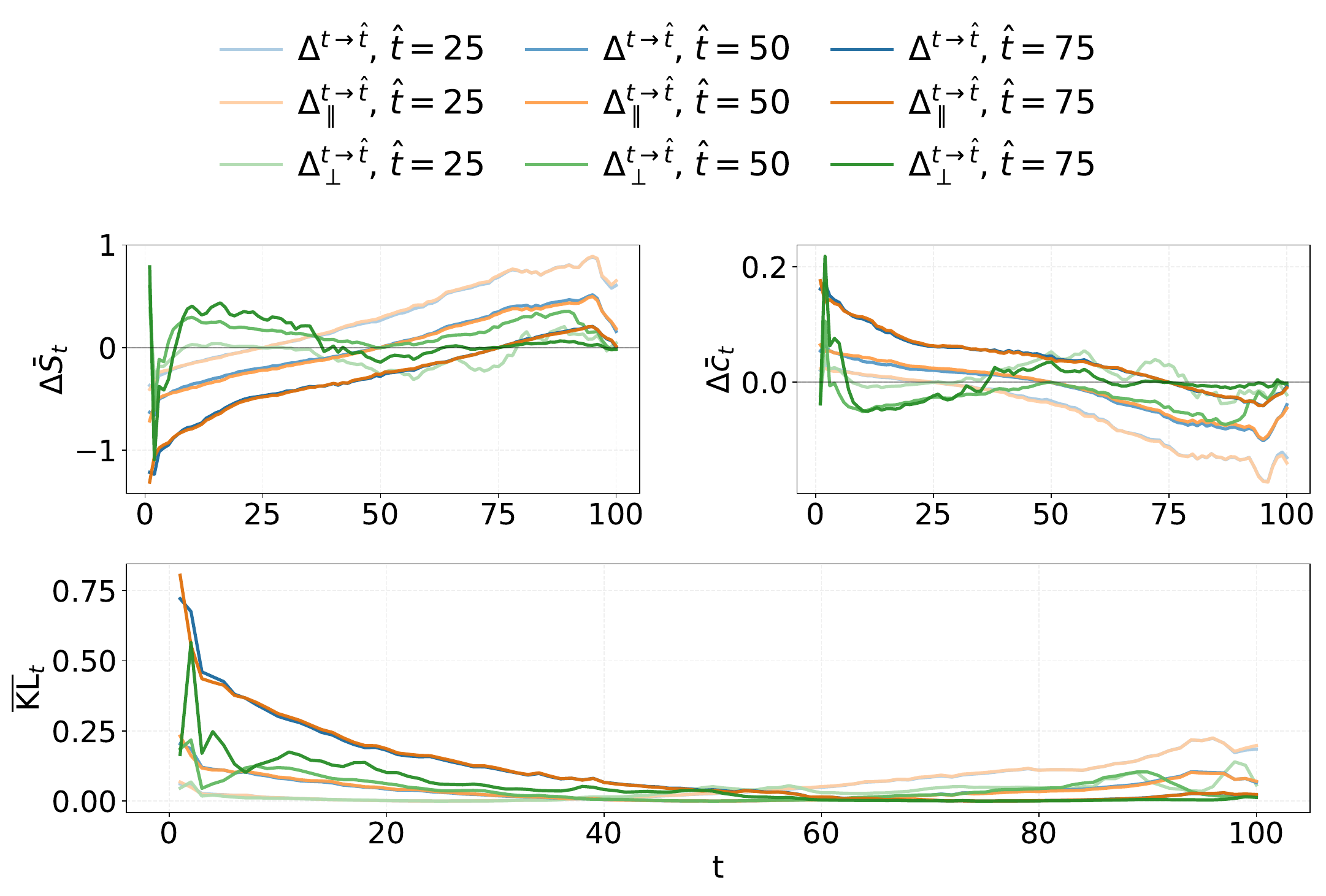}
    \end{subfigure}
    \begin{subfigure}[t]{0.45\textwidth}
        \centering
        \caption{Layer $=25$, $k=10$}
        \includegraphics[width=\textwidth]{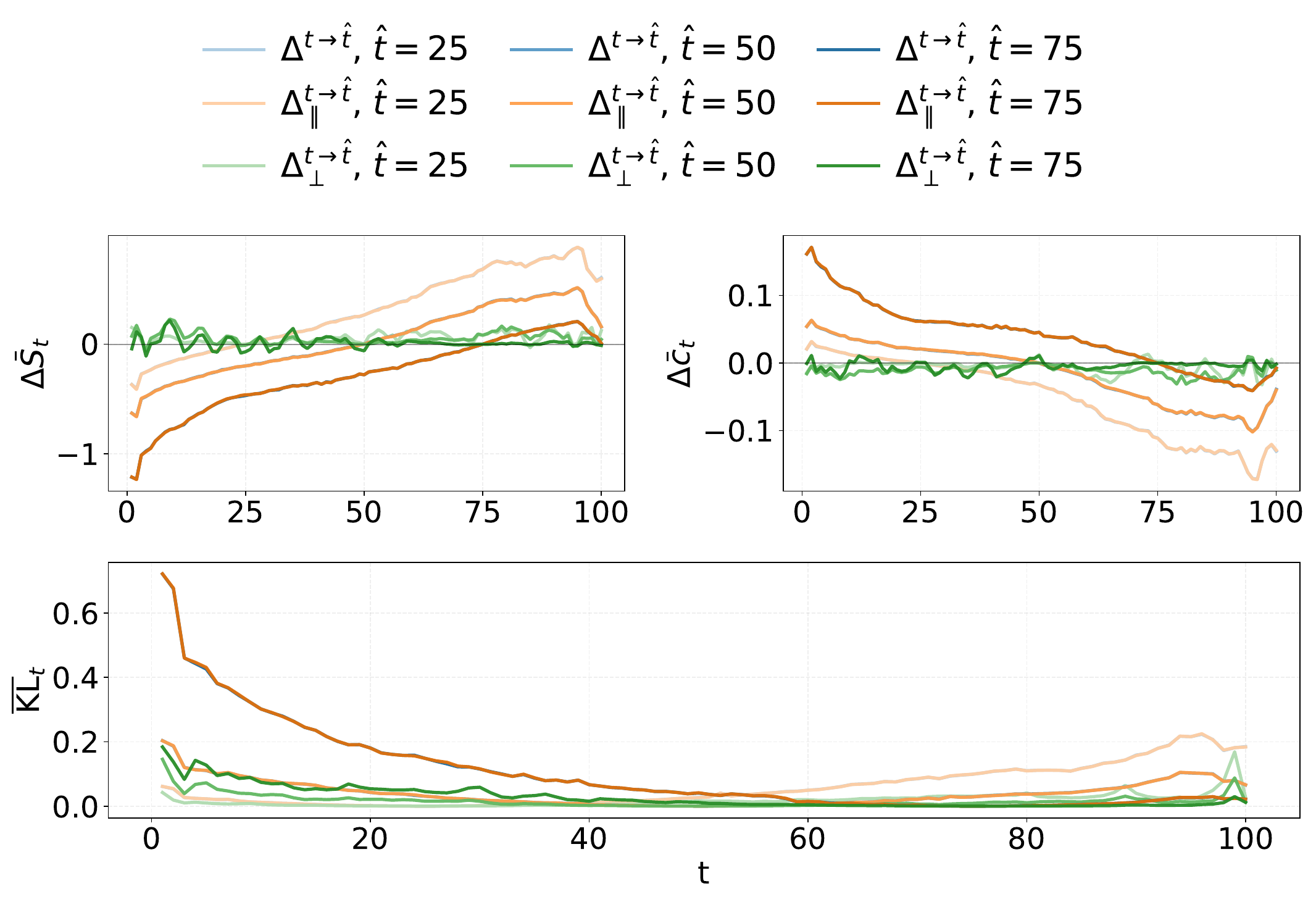}
    \end{subfigure}

    \caption{\textbf{Low-dimensional steering on \dream.} \dream's mean vectors concentrate around a low-dimensional subspace as in \llada. Similar to \Cref{fig:llada_subspace_steering}, steering across the top-1, top-2, top-4, and top-10 principal components yields results similar to using the unprojected mean vector.}
    \label{fig:dream_subspace_steering}
\end{figure*}

In this section we provide further insights on the causal importance of the found low-dimensional subspaces when steering \llada and \dream. To do so, we employ the subspace steering method introduced in \Cref{eq:subspace_mean_distortion}. As \Cref{fig:llada_subspace_steering} and \Cref{fig:dream_subspace_steering} suggest, with just the 1D projection of the mean vectors, we can recover a steering downstream impact similar to the one we attain when performing the steering with the untouched mean vectors (\Cref{eq:steer_mean_switch}).
Furthermore, as we increase the subspaces' dimensions from one to two, the results resemble with high precision the phenomena induced by the untouched mean vectors, implying that the 2D parabolas described at \Cref{fig:canonical_2d}, do encode the majority of the $\tau$-relevant geometry.
Finally, we observe how the effects caused by $\Delta^{t \to \hat{t}}_{l,\perp}$ remain of low-impact and do not follow any clear trend, implying that those dimensions have little relevance to modelling $\tau$.

\section{Depth Correction}
\label{sec:appendixdepthCorrection}
\begin{figure}[!h]
    \centering
    \begin{subfigure}{0.45\textwidth}
        \caption{$\hat{t}=1$}
        \includegraphics[width=\textwidth]{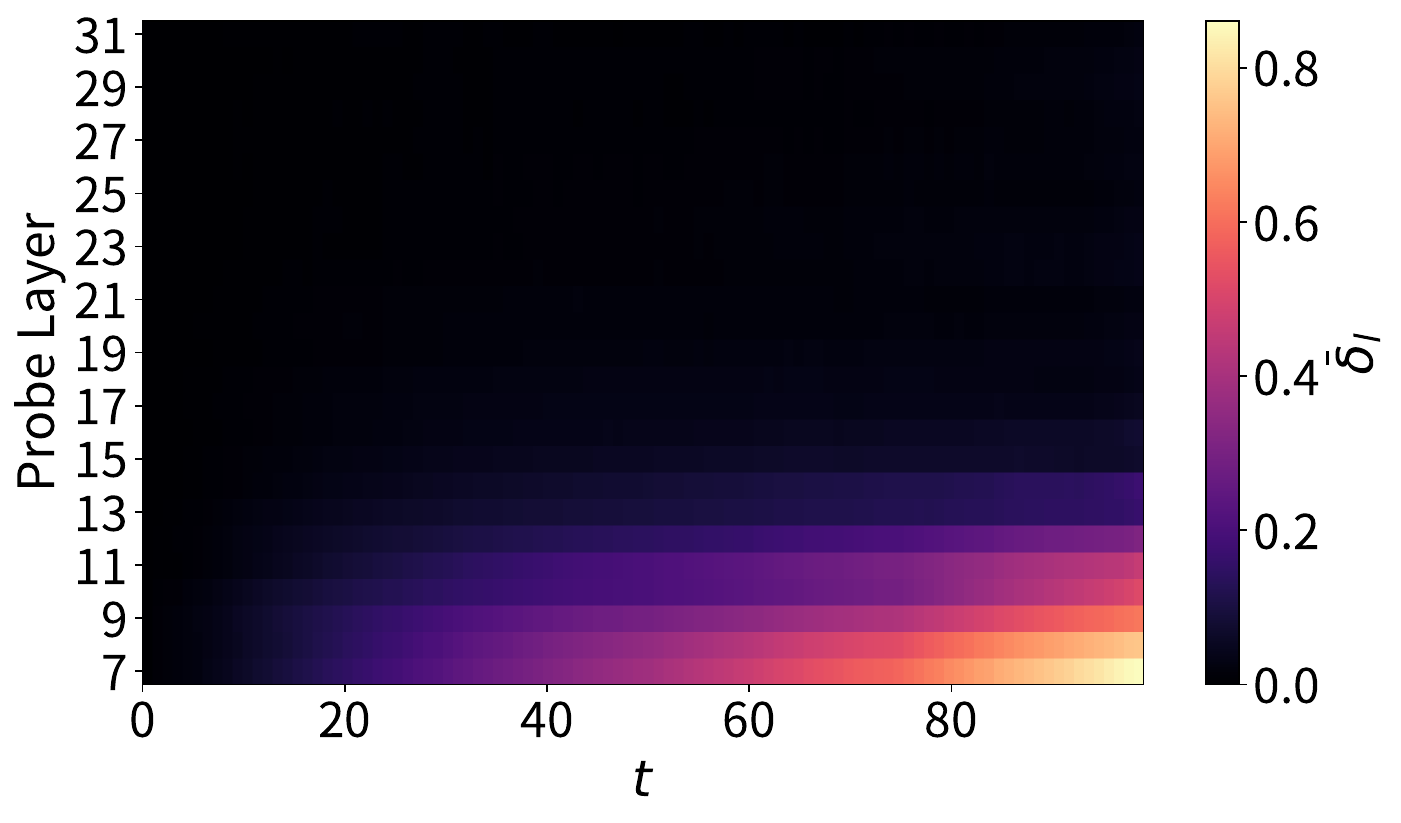}
    \end{subfigure}
    \begin{subfigure}{0.45\textwidth}
        \caption{$\hat{t}=20$}
        \includegraphics[width=\textwidth]{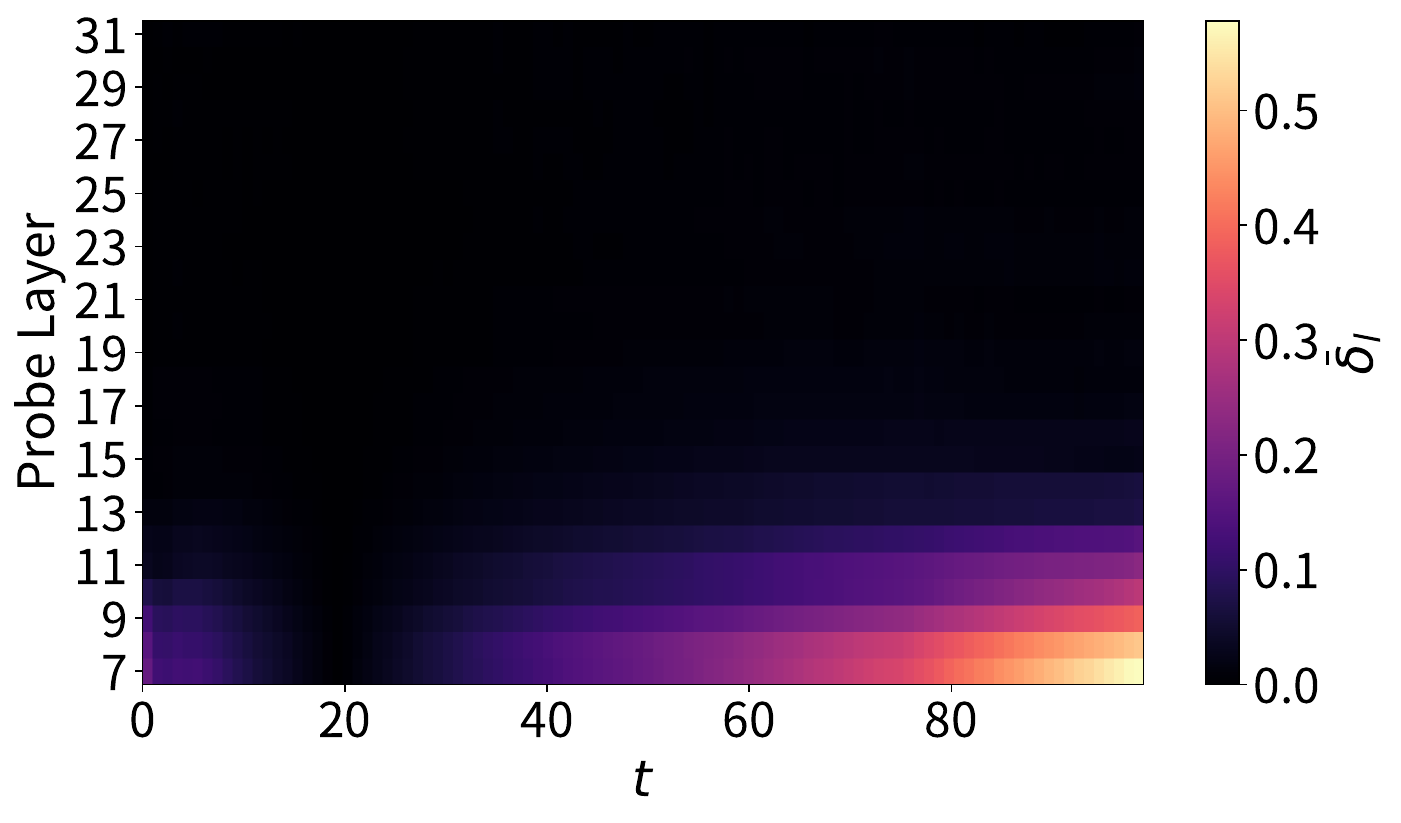}
    \end{subfigure}
    \begin{subfigure}{0.45\textwidth}
        \caption{$\hat{t}=40$}
        \includegraphics[width=\textwidth]{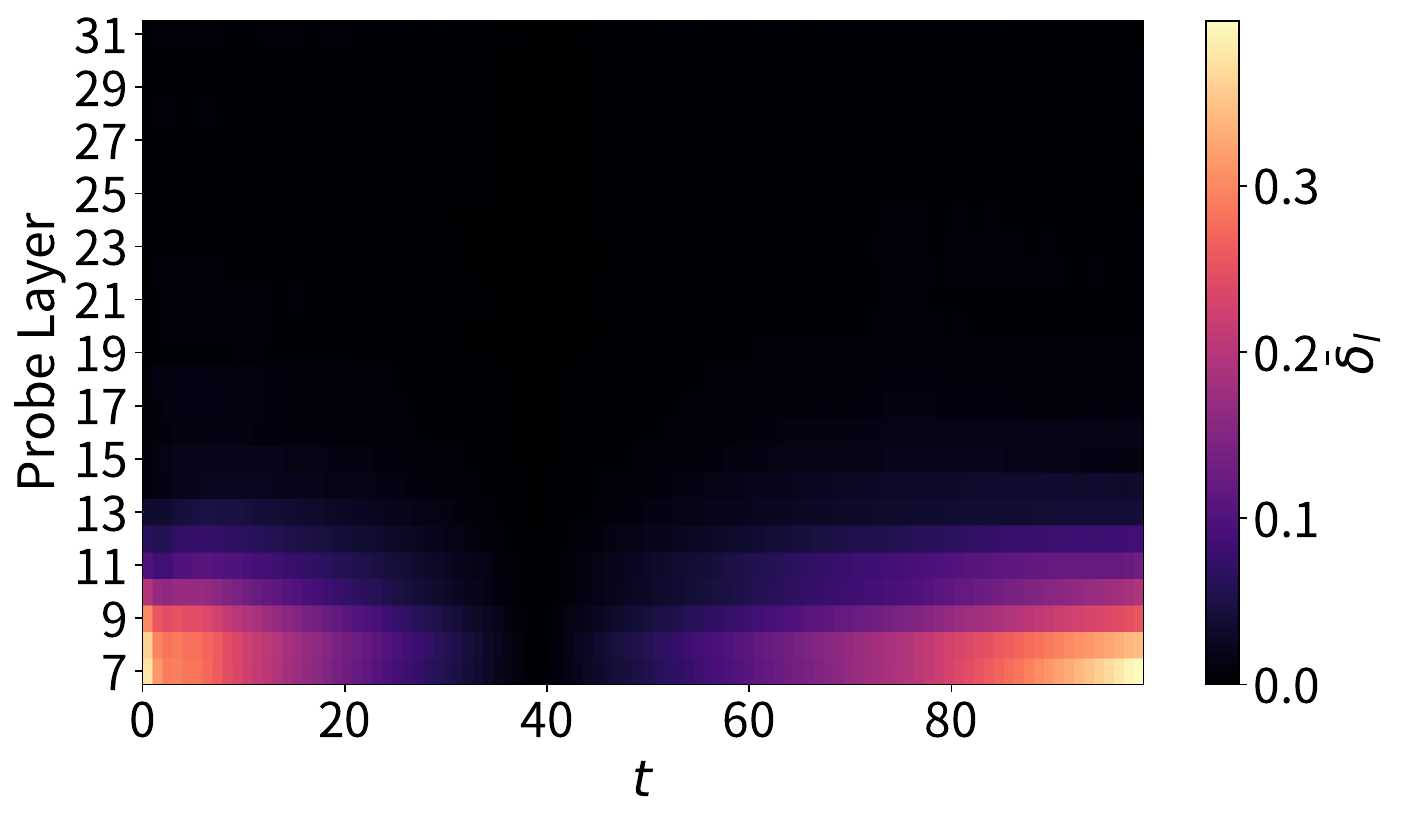}
    \end{subfigure}
    \begin{subfigure}{0.45\textwidth}
        \caption{$\hat{t}=100$}
        \includegraphics[width=\textwidth]{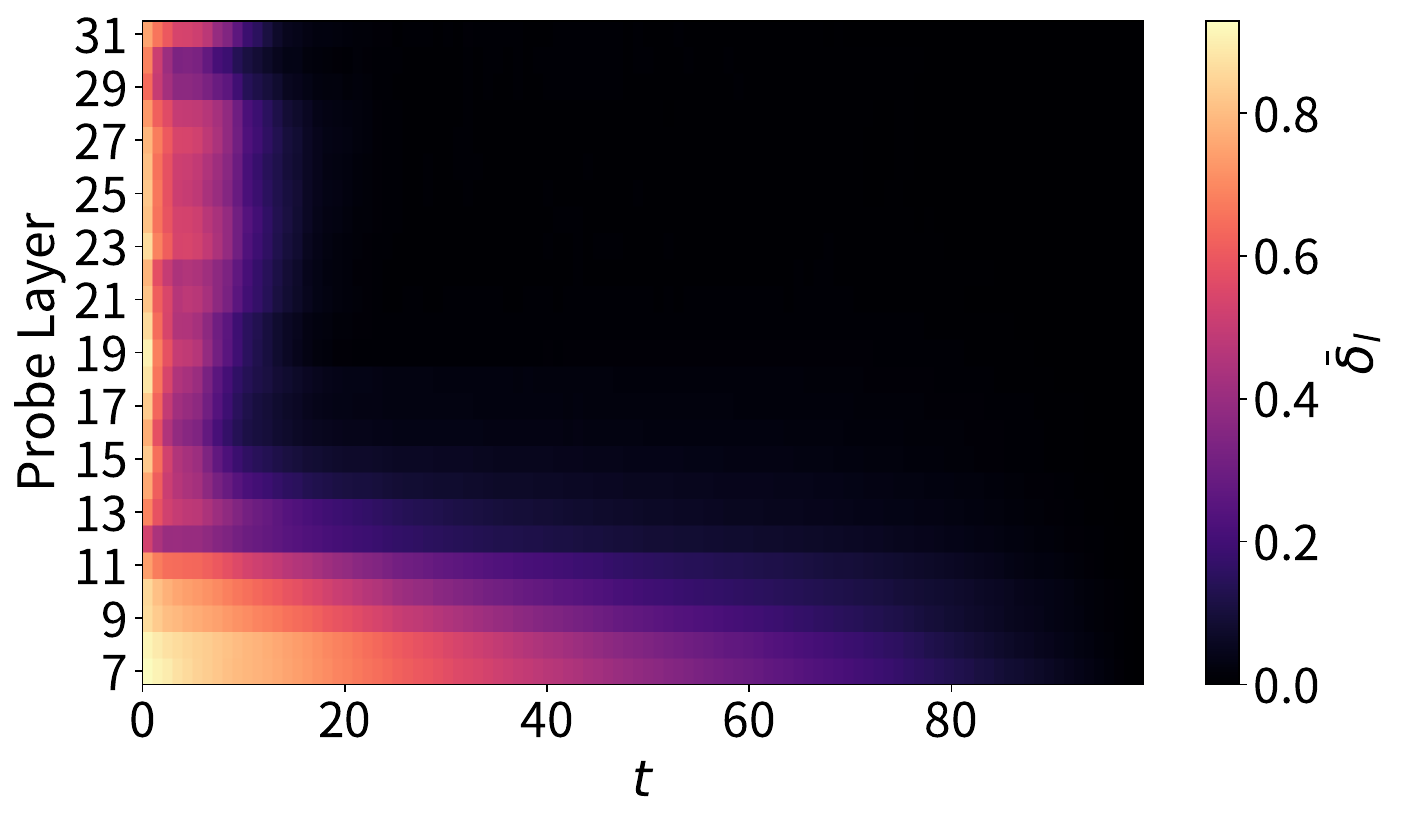}
    \end{subfigure}
    \caption{\textbf{Probe drift across layers and denoising steps when steering at layer 6.} Each column shows the absolute drift $\bar\delta_l$ between clean and steered probe predictions at a given denoising step $t$, resolved across probe layers. We can observe how only for $\hat{t}=100$ the perturbation at some steps penetrates all across the model and achieves a distortion at the last layers.}
    \label{fig:layer_depth_correction}
\end{figure}

Here we provide a finer-grained view of the depth-correction phenomenon discussed in \Cref{sec:reconstruction}.
\Cref{fig:layer_depth_correction} shows the per-step unrolled version of \Cref{fig:depth_correction} for different $\hat{t}$: instead of averaging the probe drift $\bar\delta_l$ over denoising steps, we plot one column per step, so that averaging the columns recovers the \Cref{fig:depth_correction} curve.
Early denoising steps are markedly more sensitive to the injected perturbation: the absolute probe drift remains substantial across nearly the entire depth of the network, propagating up to layer $31$, while at later steps it is suppressed within the first few layers above the injection site.
This is consistent with the sharp peaks observed for shallow-layer steering (see \Cref{sec:probeSteeringAppendix}), and indicates that the model's ability to correct an injected $\hat{t}$ depends strongly on the denoising step at which the intervention occurs.
Moreover, the residual drift decreases monotonically as the steered step becomes more compatible with $\hat{t} = 100$, i.e.\ as $t$ approaches the end of the denoising trajectory.

\section{Dream's Mean Activation Analysis} \label{sec:dreamAnalysisAppendix}
In this section, we present additional results regarding \dream's mean activations, mirroring our \llada analysis from the main text.
Similarly to the behaviour observed in \llada, we find that the representations associated with $\tau$ concentrate in a structured low-dimensional subspace, exhibiting a smooth and ordered trajectory across denoising time.

\begin{figure*}[t]
    \centering

    \begin{subfigure}{0.48\textwidth}
        \centering
        \includegraphics[width=\textwidth]{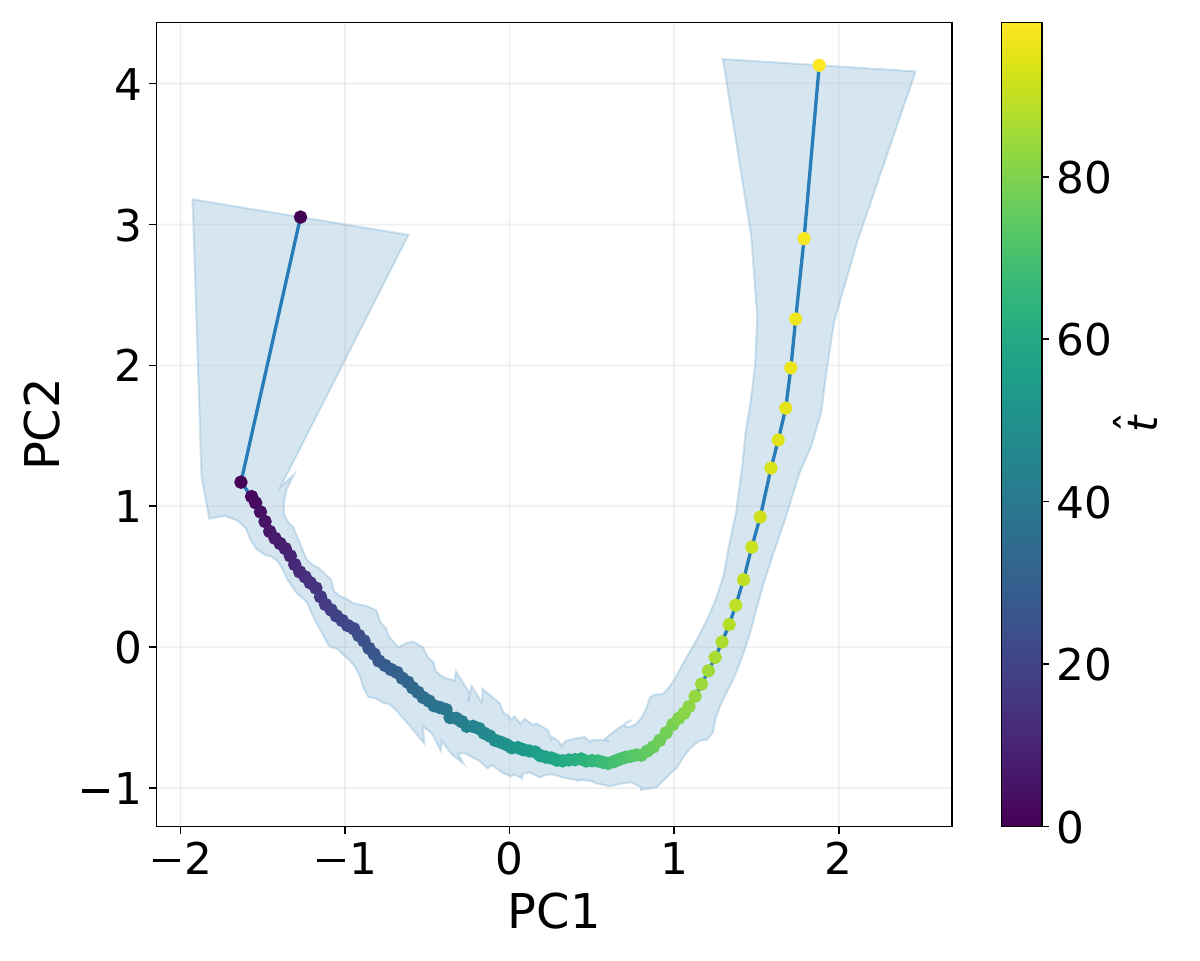}
        \caption{\textbf{The shared 2-dimensional geometry of $\hat{t}$ in \dream.}
        We compute an average trajectory by taking the standardised 2D PC projections of the mean vectors across the model. The parabolic geometry is shared across layers, similarly to \llada. PC1 and PC2 are standardised to unit variance.}
        \label{fig:canonical2dDream}
    \end{subfigure}
    \hfill
    \begin{subfigure}{0.48\textwidth}
        \centering
        \includegraphics[width=\textwidth]{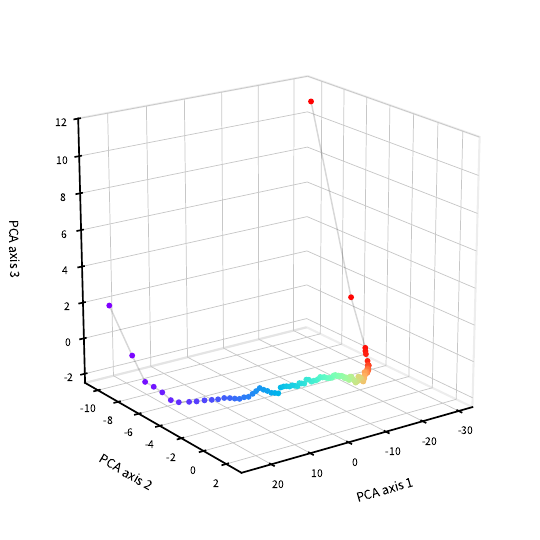}
        \caption{\textbf{3D visualisation of the PCA projection of the mean vectors for layer 25 of \dream.} As in \llada, we again observe a low-dimensional ordered structure. Notably, two points, corresponding to $t=1$ and $t=100$, lie far from the rest of the geometry, suggesting a special representation.}
        \label{fig:example_3d_dream}
    \end{subfigure}

    \caption{\textbf{Low-dimensional geometry of $\tau$'s mean-vectors in \dream.} Left: shared 2D mean-vector trajectory across layers. Right: 3D PCA projection for layer 25.}
    \label{fig:dream_low_dimensional_geometry}
\end{figure*}

\paragraph{Shared 2D geometry.}
\Cref{fig:canonical2dDream} shows the shared 2D geometry obtained by averaging the standardised PCA projections across layers for \dream.
As in \llada, the resulting geometry follows a clear parabola-like trajectory, supporting the idea of a shared geometry of $\tau$ across the whole model. Nonetheless, differently from \llada, \dream's parabola has a bigger spread at the two endpoints $t=1$ and $t=100$, implying that, at the boundaries, the geometrical organisation significantly changes across layers.

\paragraph{3D geometry.}
\Cref{fig:example_3d_dream} presents a 3D PCA projection of the mean vectors from layer 25 of \dream.
Again, the resulting trajectory reveals a low-dimensional, ordered structure.
Notably, the representations associated with the extreme denoising steps, namely the initial and final steps, appear as outliers, suggesting that the model may behave qualitatively differently at these extremes, which helps explain the big boundary deviations of \Cref{fig:canonical2dDream}.

\begin{figure*}[t]
    \centering

    \begin{subfigure}{0.48\textwidth}
        \centering
        \includegraphics[width=\textwidth]{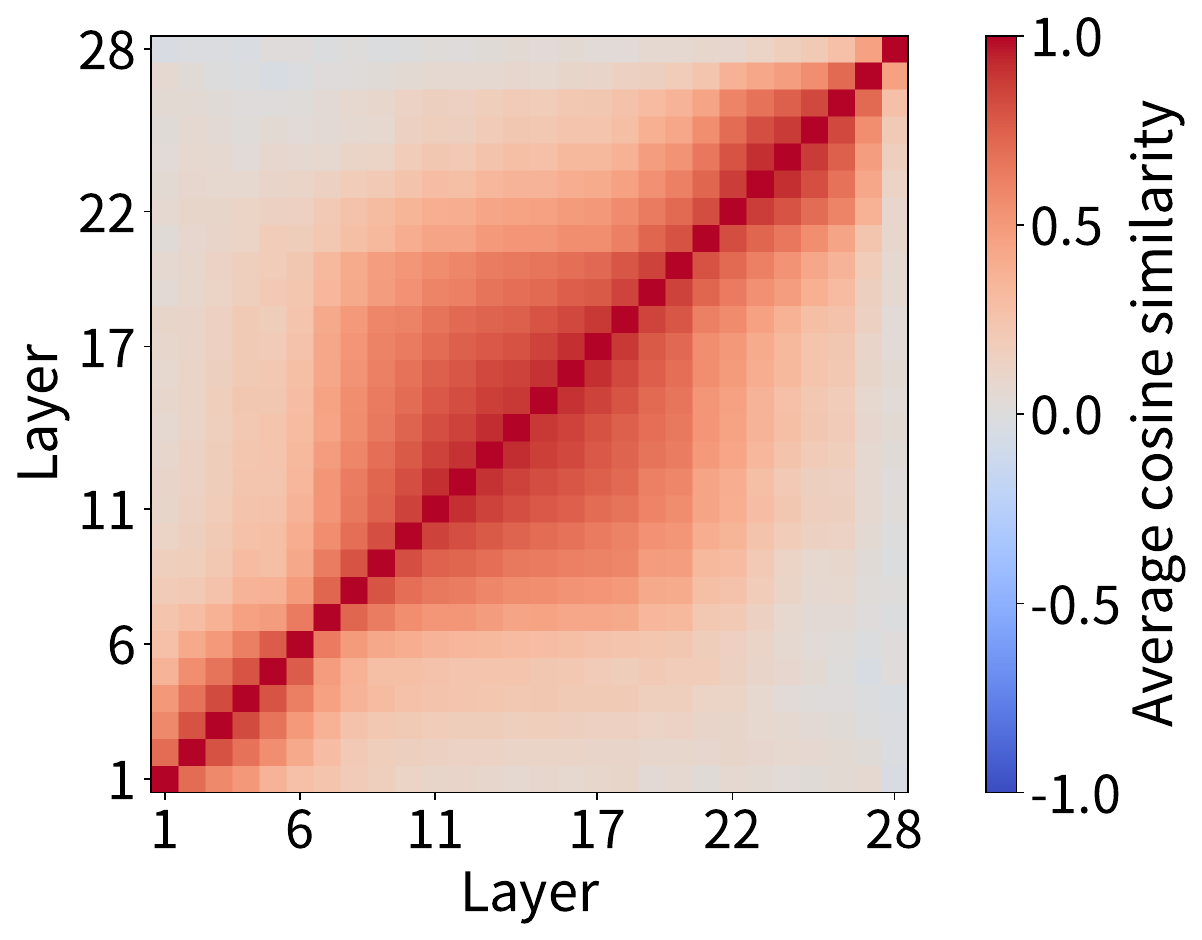}
        \caption{\textbf{Average mean vector cosine similarity for \dream.} \dream presents a more orthogonal relationship than \llada, indicating that layers are more independent in their representation of $\tau$. We also observe three structured high-similarity blocks aligned with the model's computation stages: layers 1--6, 7--20, and 21--27. Note again the last layer's orthogonality.}
        \label{fig:dream_similarity_layers}
    \end{subfigure}
    \hfill
    \begin{subfigure}{0.48\textwidth}
        \centering
        \includegraphics[width=\textwidth]{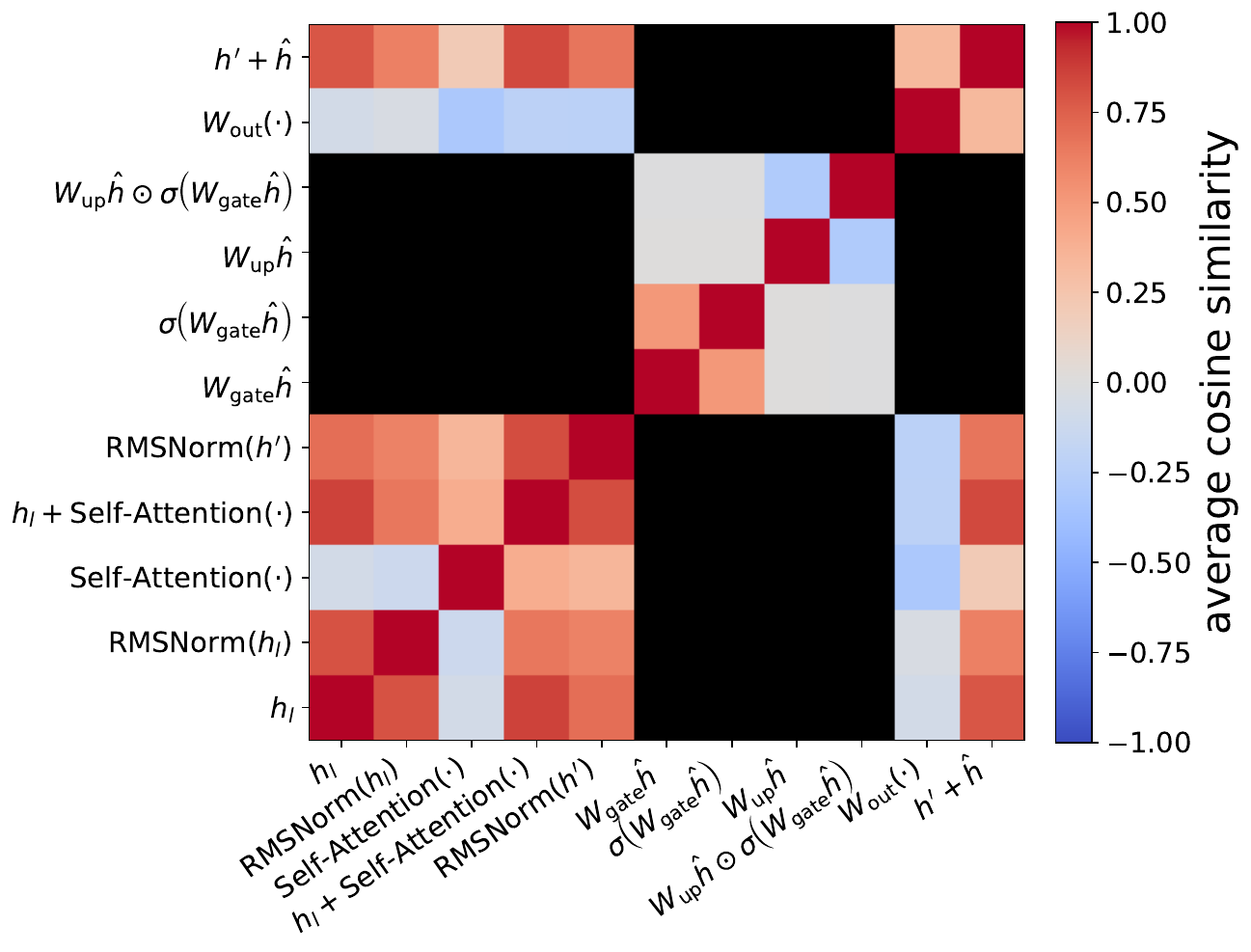}
        \caption{\textbf{Average cosine similarity across layer components in \dream.} The pre-MLP components remain highly correlated, while the MLP representations split into two groups: the gate path, and the up-projection together with the Hadamard product. As in \llada, self-attention and MLP vectors are anticorrelated.}
        \label{fig:dream_intermediates_alignment}
    \end{subfigure}

    \caption{\textbf{Cosine-similarity structure of mean vectors in \dream.} Left: average similarity across layers. Right: average similarity across intermediate layer components.}
    \label{fig:dream_cosine_structure}
\end{figure*}

\paragraph{Cross-layer alignment.}
Mirroring our \llada analysis in \Cref{fig:llada_mean_similarity}, we investigate how the $\tau$ representations align across layers for \dream.
In contrast to \llada, which exhibits strong global alignment across most layers, \Cref{fig:dream_similarity_layers} shows that \dream presents a substantially more heterogeneous organisation. 
Interestingly, the model appears to partition its layers into three main computational stages, with high cosine similarity within each block and weaker alignment across blocks.

\paragraph{Within-layer components.}
We present the cosine similarity across layer components for \dream (as for \llada in \Cref{fig:llada_intermediates_alignment}). In \Cref{fig:dream_intermediates_alignment} we can appreciate how \dream follows an internal geometric organisation of the mean vectors similar to \llada.

\end{document}